\documentclass[authoryear,preprint,review,12pt]{elsarticle}

\usepackage{amssymb}

\usepackage{graphics}
\usepackage{graphicx}
\usepackage{amsmath}

\usepackage{tikz}
\usetikzlibrary{shapes,arrows}
\usetikzlibrary{patterns}
\usetikzlibrary{spy,backgrounds}

\usepackage{pgfplots}   

\usepackage{bm}
\usepackage{bbm}

\usepackage{subcaption}   

\usepackage[labelsep=quad,skip=3pt]{caption}
\usepackage{physics}

\usepackage{pstricks}
\usepackage{pst-fill,pst-grad}
\usepackage{pst-text}
\usepackage{pst-uml}
\usepackage{pst-3d,pst-coil,pst-eps,pst-node}
\usetikzlibrary{shapes}
\usetikzlibrary{arrows}
\usetikzlibrary{positioning}
\usetikzlibrary{decorations.markings}
\usetikzlibrary{decorations.text}
\usetikzlibrary{fadings}

\renewcommand\paragraph{\@startsection{paragraph}{4}{\z@}%
            {-2.5ex\@plus -1ex \@minus -.25ex}%
            {1.25ex \@plus .25ex}%
            {\normalfont\normalsize\bfseries}}
\usepackage{xcolor}
\usepackage[framemethod=TikZ]{mdframed}
\definecolor{ikmgray}{HTML}{5E5E5E}
\definecolor{ikmgreen}{HTML}{C9DA2B}
\newmdenv[frametitle={},
middlelinecolor=ikmgreen,
middlelinewidth=0pt,
backgroundcolor=ikmgray!20,
roundcorner=2pt,
bottomline=false,
leftline=true,
topline=false,
rightline=false,
skipabove=10pt,
skipbelow=10pt,
leftmargin=10pt,
rightmargin=10pt,
innerleftmargin=10pt,
innerrightmargin=10pt,
innertopmargin=10pt,
innerbottommargin=10pt]{Algorithmus}
\usepackage[labelfont=bf, 
format=plain, 
labelsep=endash, 
justification=raggedright 
]{caption}
\DeclareCaptionType{kasten}[Box]
\usepackage{multirow}

\usepackage{arydshln}

\usepackage[acronym]{glossaries}

\makeglossaries
\newacronym{gcd}{GCD}{Greatest Common Divisor} 
\newacronym{lcm}{LCM}{Least Common multiple}
\newacronym{olhd}{OLHD}{Optimal Latin Hypercube Design}
\newacronym{svm}{SVM}{Support Vector Machines}
\newacronym{pr}{PR}{Polynomial Regression}
\newacronym{rbf}{RBFN}{Radial Basis Function Network}
\newacronym{nn}{NN}{Neural Networks}
\newacronym{mae}{MAE}{Mean Absolute Error}
\newacronym{lf}{LF}{Low-Fidelity}
\newacronym{hf}{HF}{High-Fidelity}
\newacronym{doe}{DOE}{Design of experiments}
\newacronym{ok}{OK}{Ordinary Kriging}
\newacronym{uk}{UK}{Universal Kriging}
\newacronym{hk}{HK}{Hierarchical Kriging}
\newacronym{pls}{PLS}{Partial Least Squares}
\newacronym{plsok}{PLSOK}{Partial Least Squares Ordinary Kriging}
\newacronym{plshk}{PLSHK}{Partial Least Squares Hierarchical Kriging}
\newacronym{mipt}{MIPT}{Monte Carlo-intersite-proj-th}
\newacronym{msd}{MSD}{Maximin Scaled Distance}
\newacronym{cv}{CV}{Cross-Validation}
\newacronym{cvv}{CVV}{Cross-Validation Variance}
\newacronym{cvvor}{CVVOR}{Cross-Validation-Voronoi}
\newacronym{cdm}{CDM}{Crowding Distance Metric}
\newacronym{ssa}{SSA}{Smart Sampling Algorithm}
\newacronym{gcv}{GCV}{Generalized Cross-Validation}
\newacronym{mse}{MSE}{Mean-Squared Error}
\newacronym{mmse}{MMSE}{Maximum Mean-Squared Error}
\newacronym{imse}{IMSE}{Integrated Mean-Squared Error}
\newacronym{ame}{AME}{Adaptive Maximum Entropy}
\newacronym{ei}{EI}{Expected Improvement}
\newacronym{cdf}{CDF}{Cumulative Distribution Function}
\newacronym{pdf}{PDF}{Probability Distribution Function}
\newacronym{wei}{WEI}{Weighted Expected Improvement}
\newacronym{awei}{AWEI}{Adaptive Weighted Expected Improvement}
\newacronym{eigf}{EIGF}{Expected Improvement for Global Fit}
\newacronym{haed}{HAED}{Hierarchical Adaptive Experimental Design}
\newacronym{gek}{GEK}{Gradient-Enhanced Kriging}
\newacronym{nurbs}{NURBS}{Non-Uniform Rational B-Spline}
\newacronym{lola}{LOLA}{Local Linear Approximation}
\newacronym{qbc}{QBC}{Query-By-Committee}
\newacronym{masa}{MASA}{Mixed Adaptive Sampling Algorithm}
\newacronym{lhd}{LHD}{Latin Hypercube Design}
\newacronym{tplhd}{TPLHD}{Translational Propagation Latin Hypercube Design}
\newacronym{rmse}{RMSE}{Root Mean-Squared Error}
\newacronym{rmae}{RMAE}{Relative Maximum Absolute Error}
\newacronym{cvd}{CVD}{Cross-Validation Distance}
\newacronym{le}{LE}{Lyapunov Exponents}
\newacronym{lle}{LLE}{Largest Lyapunov Exponent}
\newacronym{epe}{EPE}{Expected Prediction Error}
\newacronym{mepe}{MEPE}{Maximizing Expected Prediction Error}
\newacronym{loocv}{LOCVV}{leave-one-out cross-validation}
\newacronym{gmse}{GMSE}{Generalized Mean Square Cross-Validation Error}
\newacronym{sfcvt}{SFCVT}{Space-Filling Cross Validation Tradeoff}
\newacronym{ace}{ACE}{ACcumulative Error}
\newacronym{doi}{DOI}{Degree-of-Influence}
\newacronym{blup}{BLUP}{Best linear unbiased predictor}
\newacronym{mle}{MLE}{Maximum likelihood estimation}
\newacronym{de}{DE}{Differential evolution}
\newacronym{mivor}{MiVor}{Monte Carlo-intersite Voronoi}
\newacronym{qoi}{QoI}{Quantity of Interest}
\newacronym{mob}{MoB}{Mass-on-Belt}

\begin{document}

\begin{frontmatter}




\title{An innovative adaptive kriging approach for efficient binary classification of mechanical problems }

\cortext[cor1]{Corresponding author}
 \author[label1]{Jan N. Fuhg \corref{cor1}}

 \ead{fuhg@ikm.uni-hannover.de}
\address[label1]{Institute of continuum mechanics,
              Leibniz Universit{\"a}t Hannover,
              Appelstra{\ss}e 11,
              30167 Hannover, Germany}
\author[label2]{Am\'{e}lie Fau}
\address[label2]{Institute of mechanics and computational mechanics,
              Leibniz Universit{\"a}t Hannover,
              Appelstra{\ss}e 9A,
             30167 Hannover, Germany}



\begin{abstract}
Kriging is an efficient machine-learning tool, which allows to obtain an approximate response of an investigated phenomenon on the whole parametric space. Adaptive schemes provide a the ability to guide the experiment yielding new sample point positions to enrich the metamodel. Herein a novel adaptive scheme called \gls{mivor} is proposed to efficiently identify binary decision regions on the basis of a regression surrogate model. The performance of the innovative approach is tested for analytical functions as well as some mechanical problems and is furthermore compared to two regression-based adaptive schemes. For smooth problems, all three methods have comparable performances. For highly fluctuating response surface as encountered e.g. for dynamics or damage problems, the innovative \gls{mivor} algorithm performs very well and provides accurate binary classification with only a few observation points.
\end{abstract}


\end{frontmatter}


\printglossary

\section{Introduction}

Machine learning tools, such as decision forests \citep{criminisi2012decision}, support vector machines \citep{gunn1998support}, neural networks \citep{specht1991general,zhang2000neural} and Gaussian processes \citep{rasmussen2006gaussian}, appear nowadays very promising to study the mapping between input and output data of a black-box function (e.g. simulation or physical experiment) and to create low-cost metamodels or guide experiments. 

\begin{figure}[ht!]
\centering
\begin{subfigure}[t]{0.48\textwidth}
\includegraphics[scale=0.7]{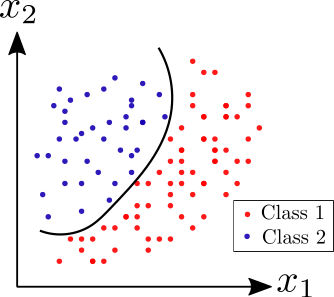}
\subcaption{Classification}
\end{subfigure}%
\begin{subfigure}[t]{0.48\textwidth}
\includegraphics[scale=0.7]{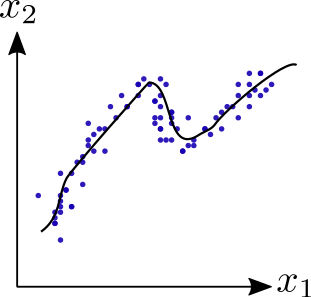}
\subcaption{Regression}
\end{subfigure}
\caption{Schemes of both types of supervised learning.}\label{fig::class_reg}
\end{figure}

Two types of supervised learning methods can be distinguished, classification and regression, see Figure \ref{fig::class_reg}. Classification deals with discrete class labels and is used e.g. in engineering for fault diagnosis of bearings 
 \citep{zhi2005support, samanta2003artificial} or treating reliability analysis as a classification task \citep{hurtado2003classification}. Regression is utilized for the prediction of continuous output and is e.g. used for design optimization \citep{liu2014efficient} or machine degradation assessment \citep{caesarendra2010application}.
 
Typically, in order to classify output data in applied mechanics and engineering a metamodel is trained by using class labels as output  \citep{zhi2005support}. In this context Gaussian process classification \citep{rasmussen2006gaussian,rasmussen2010gaussian} provides proficient surrogate classification. Various numerical approaches can be distinguished depending on the way of estimating the posterior \citep{nickisch2008approximations}. Laplace approximation obtains the Gaussian approximation of the posterior from a second-order Taylor expansion \citep{williams1998bayesian}. Expectation propagation yields an approximation through an iterative method based on marginal moments, see \cite{minka2001expectation} with extension in \cite{deisenroth2012expectation}, \cite{riihimaki2013nested}, \cite{tolvanen2014expectation} or \cite{dehaene2018expectation}. Variational bounds have been proposed in \cite{gibbs2000variational} and extended in e.g. \cite{hensman2015scalable}.

\begin{figure}[htp!]
    \centering
    \includegraphics[scale=0.7]{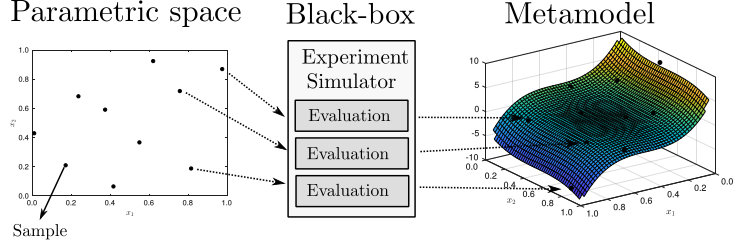}
    \caption{Creating a metamodel with supervised learning from the data of a black-box function.}
    \label{fig:my_label}
\end{figure}

However, in various applications in computational engineering a continuous function is explicitly available and needs to be transformed into binary class labels to evaluate mechanical behavior, e.g.
\begin{itemize}
    \item Defining the failure of a mechanical system (failure or no failure) based on the exceedance of a limit value with continuous quantities of interest (stress or strain), e.g. \cite{wolfe1998strain} or \cite{labuz2012mohr},
    \item Classifying the motion of a dynamic system (regular of chaotic) with a continuous indicator e.g. the largest Lyapunov exponent \citep{pesin1977characteristic, muller1995calculation, Jan_Master_thesis},
    \item Determining crack growth appearance from a pre-existing flaw, i.e. comparing the continuous energy release rate to a required energy value, see \cite{anderson2017fracture}.
\end{itemize}
This information can also be relevant to build proficient metamodels. Generally, supervised learning can be employed to generate a metamodel, which acts as an approximate of the black-box on the whole parametric space as illustrated in Figure \ref{fig:my_label} from a one-shot technique \citep{liu2017survey}.  

\begin{figure}[htp!]
\centering
  \includegraphics[trim = 110 600 180 50 , width =0.7\textwidth]{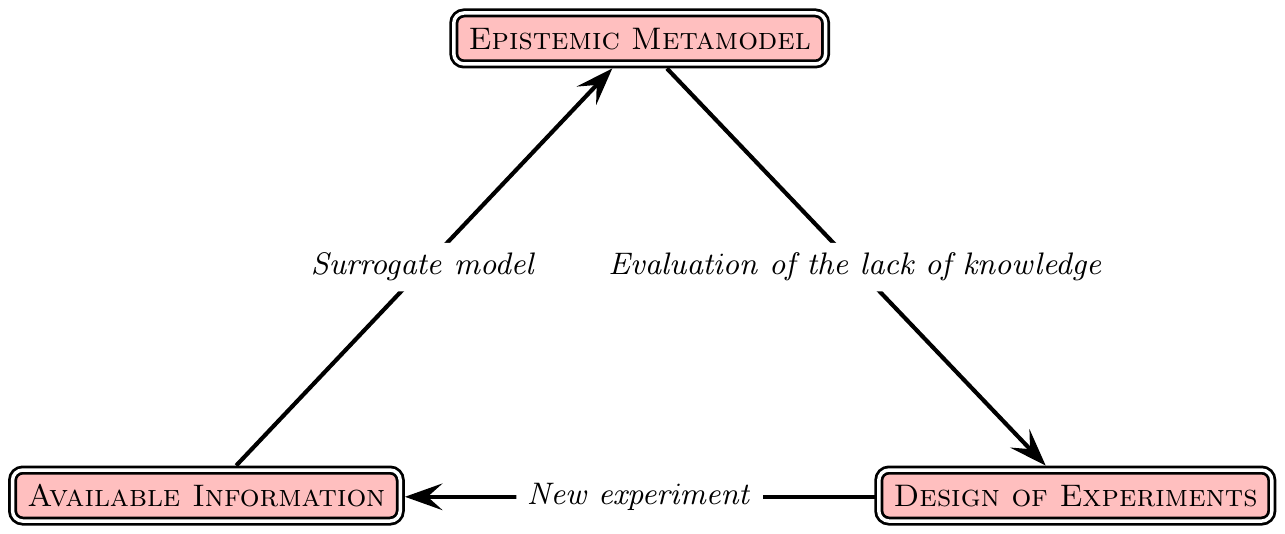}
\caption{Schematic representation of the adaptive kriging approach.}
\label{fig:adapt_scheme}       
\end{figure}

However, when the evaluation of the experiment requires high computational effort the number of utilizable evaluations is restricted by time constraints. Therefore another aim is to generate the best possible metamodel (by some error measure) with the least number of black-box evaluations. This inspired the use of adaptive sampling techniques, where samples are added to an existing dataset in an iterative procedure as schematized in Figure \ref{fig:adapt_scheme}. A surrogate model is generated from available information with an intrinsic lack of knowledge, which in turn can be investigated and used to obtain further observations. The process can be guided to be performed in an optimal manner to enlarge the dataset with new available information. 
 





The idea is to start from some initial data $\mathcal{D}_{ini} = \lbrace \left( \bm{x}_{i}, \,\bm{y}_{i} \right), \, i=1, \, \ldots  , \, m  \rbrace$ which is used to estimate a surrogate model $\tilde{\mathcal{M}}$. Through adaptive strategies new samples are obtained by solving auxiliary optimization problems, see \cite{Jan_Master_thesis}. The general optimization problem reads \citep{liu2017survey}
\begin{equation}\label{eq::}
\bm{x}_{m+1} = \arg \, \min_{\bm{x}^{\star} \in \mathbb{X}} \, Score \left( \bm{x}^{\star} \right) \, \text{,}
\end{equation}
where the score function $Score$ generally represents a tradeoff between exploration and exploitation components. The exploration contribution aims to investigate the input domain evenly in order to detect some regions of particular interest. On the contrary, the exploitation part tries to generate data points locally in already identified regions of interest with regards to the reduction of the prediction error. The literature offers different techniques to account for this tradeoff, e.g. \cite{jones1998efficient}, \cite{singh2013balanced}, \cite{turner2007multidimensional}, \cite{liu2016adaptive}. A good overview is given in \cite{liu2017survey}.

In general, new points are iteratively added to the dataset $\mathcal{D}$ until a stopping criterion is reached, which e.g. could be expressed in terms of a maximal number of experiments. New experiment refers in this context to any process which supplies additional observation. Thus, it can be provided by physical experiments, numerical experiments, or a combination of them.

In this paper the goal is to provide an accurate classifier based on the knowledge of a continuous quantity of interest with only a few effectively chosen samples. The approach is based on kriging metamodel, also called Gaussian process regression, which provides an accurate interpolating surrogate modeling technique exhibiting exact estimation at the observation points and a stochastic property, i.e. predicted variances between the black-box output and the metamodel output can be obtained. 
An overview of adaptive schemes for Kriging including a comparative review in terms of performance for benchmark functions and mechanical problems has been presented in \cite{Jan_Master_thesis}.
The problems of concern exhibit highly fluctuating response surfaces. However Gaussian process classification does not appear robust in that case. A kriging regression approach including adaptive sampling technique for classification goal is utilized instead. To the best of the authors knowledge it is the first adaptive scheme for classification purposes. It allows for easy implementation and a proficient adaptive surrogate classification for non-smooth response surfaces. The adaptive scheme is able to robustly balance the detection of the two classes and to accurately identify decision regions. The Matlab code for the presented method including a running example can be downloaded at Github under the link: \textit{https://github.com/FuhgJan/AdaptiveMIVor}. 
 
 The article is structured as follows. In Section \ref{Sec::Kriging} Gaussian process regression, on which the approach is based, is summarized. In Section \ref{sec::classification} the classification problem of interest is defined, the kriging classifier is introduced and compared with Gaussian process classification proposed in the literature for one-shot surrogate classification. Section \ref{sec:MIVor} features the proposed adaptive scheme, which aims to provide a proficient metamodel for binary classification based on few and optimised observations. Finally in Section \ref{sec::Applications} the algorithm is tested on various one- and two-dimensional numerical problems.

\section{Gaussian processes regression}
\label{Sec::Kriging}

Consider the training set with $m$ uni-variate observations $\mathcal{D} = \lbrace ( \bm{x}_{i},y_{i})| i=1, \ldots, m \rbrace$, where $\bm{x} \in \mathbb{R}^{n}$ represents the input vector and the observations are denoted by $y$. The design matrix $\bm{X} \in \mathbb{R}^{n \times m}$ collects the input vectors and the output vector $\bm{y} \in \mathbb{R}^{m}$ aggregates the observations. Gaussian process regression assumes that the functional relationship between the input and output data can be modeled by a sample path of a stationary Gaussian process.
A stationary Gaussian process is a collection of random variables with Gaussian distribution \citep{rasmussen2006gaussian}. Therefore, it is completely defined by its mean $\mu(\bm{x})$ and covariance function $k(\bm{x}, \bm{x}')$ given by
\begin{equation}
k(\bm{x}, \bm{x}') = \sigma^{2} R(\bm{x}, \bm{x}', \bm{\theta}),
\end{equation}
where $\sigma^{2}$ is the variance and the auto-correlation $R$ characterizes the correlation between every pair of points $(\bm{x}, \bm{x}')$ in the input space. The hyperparameters $\bm{\theta}$ denote a set of unknown parameters which characterize the auto-correlation structure in each dimension. 
Herein only the Mat\'{e}rn 3/2 auto-correlation function \citep{matern1960spatial}, which is given by
\begin{equation}
\begin{aligned}
R (\bm{x} - \bm{x'}, \bm{l})  = \prod_{i=1}^{n} \left( 1 + \dfrac{\sqrt{3} \abs{x_{i} - x_{i}'} }{l_{i}} \right) \, \exp \left(-\dfrac{\sqrt{3} \abs{x_{i} - x_{i}'} }{l_{i}}  \right) 
\end{aligned}
\end{equation}
is used with possibly a different correlation length $l_i$ for each dimension $i\in [1,n]$ of the parametric space.

Consider the unobserved input $\bm{x}_{\star}$ with the unknown quantity $y_{\star}$. The joint distribution of $\bm{y}$ and $y_{\star}$ reads
\begin{equation}
\begin{bmatrix}
\bm{y} \\ y_{\star}
\end{bmatrix} \sim \mathcal{N}_{m+1}  \left(\begin{bmatrix}
\bm{\mu} \\ \mu_{\star}
\end{bmatrix}, \sigma^{2} \begin{bmatrix}
\bm{1} & \bm{r}_{\star}^{T} \\
\bm{r}_{\star} & \bm{R}
\end{bmatrix} \right),
\end{equation}
where $\mathcal{N}_{m+1}(\bullet)$ denotes the $m+1$-dimensional multivariate normal distribution.
$\bm{R}$ is the correlation matrix, $\bm{1}$ is a vector of ones and $\bm{r}_{\star}$ denotes the cross-correlations between the prediction point $\bm{x}_{\star}$ and the available observations $\lbrace \bm{x}_{i} \rbrace$, which reads
\begin{equation}
r_{\star \, i} = R(\bm{x}_{\star} - \bm{x}_{i}, \bm{\theta}) \, \qquad i=1, \, \ldots \, , m.
\end{equation}
The conditional probability of $y_{\star}$ knowing $\bm{y}$ follows a Gaussian distribution of the form
\begin{equation}
\begin{aligned}
y_{\star}| \bm{y} \sim \mathcal{N}_{1} (&\hat{\mu} + \bm{r}_{\star}^{T} \bm{R}^{-1} (\bm{y}- \bm{1} \hat{\mu}) , \\&\hat{\sigma}^{2} ( 1 - \bm{r}_{\star}^{T} \bm{R}^{-1} \bm{r}_{\star} + \bm{u}_{\star}^{T} \left( \bm{1}^{T} \bm{R}^{-1} \bm{1}\right)^{-1} \bm{u}_{\star})).
\end{aligned}
\end{equation}
The best estimate for $y_{\star}$ is the mean of the distribution where the symbol $\hat{.}$ represents an estimator
\begin{equation}
\begin{aligned}
\hat{y}_{\star} &= \hat{\mu} + \bm{r}_{\star}^{T} \bm{R}^{-1} (\bm{y}- \bm{1} \hat{\mu}) 
\end{aligned}
\end{equation}
which is quantified with the uncertainty 
\begin{equation}
\begin{aligned}
\hat{\sigma}^{2}_{y_{\star}} &= \hat{\sigma}^{2} \left( 1 - \bm{r}_{\star}^{T} \bm{R}^{-1} \bm{r}_{\star} + \bm{u}_{\star}^{T} \left( \bm{1}^{T} \bm{R}^{-1} \bm{1}\right)^{-1} \bm{u}_{0}\right),
\end{aligned}
\label{eq:var_kriging}
\end{equation}
where 
\begin{equation}
\bm{u}_{0} = \bm{1}^{T} \bm{R}^{-1} \bm{r}_{\star} - \bm{1} \, \text{.}
\end{equation}
In this expression, the estimator of the mean reads
\begin{equation}
\hat{\mu} = (\bm{1}^{T} \bm{R}^{-1} \bm{1})^{-1} \bm{1}^{T} \bm{R}^{-1} \bm{y}, 
\end{equation}
and the estimator of the variance is given by
\begin{equation}
\hat{\sigma}^{2} = \frac{1}{m} \left( \bm{y} - \bm{1} \hat{\mu} \right)^{T} \bm{R}^{-1} \left( \bm{y} - \bm{1} \hat{\mu} \right).
\end{equation}

During the training phase the correlation hyperparameters $\bm{\theta}$ can be obtained by e.g. maximizing 
the reduced likelihood function $\psi$ given by
\begin{equation}
\begin{aligned}
\psi (\bm{\theta}) = \hat{\sigma}^{2} (\bm{\theta}) [\det \bm{R} (\bm{\theta})]^{1/m},
\end{aligned}
\end{equation}
which yields
\begin{equation}\label{eq::minimization_hyperparameters}
\hat{\bm{\theta}} =   \arg \, \min_{\bm{\theta}^{\star}}  \psi (\bm{\theta}^{\star}) \, \text{.}
\end{equation}
This optimization turns out to be a numerical bottleneck in the kriging algorithm due to the multimodality of the likelihood function, see \cite{bouhlel2019gradient}. Here it is solved using a hybridized particle swarm optimization algorithm, see \cite{toal2011development}. 

\section{Classification approach based on kriging regression}
\label{sec::classification}

In various applications in engineering science the interest is not on the exact value of a quantity but on decision regions, i.e. on the classification of a problem. However, binary class labels may itself be based on the knowledge of a continuous function output which is analyzed, e.g. failure or no-failure of a structure can be determined from stress values. In this this paper, classification problems are based on data obtained as numerical quantities from simulation or experiment. 

\subsection{Definition of the classification problem}
The concern is precisely on binary classification problems between the two class label $\mathcal{C}_{1}$ and $\mathcal{C}_{2}$ defined from the output quantity in accordance with the following rule
\begin{equation}
    \mathcal{C}(\bm{x}) =  \begin{cases}
    \mathcal{C}_{1}, & \text{if} \, \mathcal{M}(\bm{x}) \geq L, \\
    \mathcal{C}_{2}, & \text{else},
    \end{cases}
\end{equation}
where $L \in \mathbb{R}$, $\bm{x} \in \mathbb{R}^{n}$ and $\mathcal{M}: \mathbb{R}^{n} \rightarrow \mathbb{R}$. The goal is to provide an efficient numerical strategy to estimate the surrogate classifier $\hat{\mathcal{C}}$ of $\mathcal{C}$.

\subsection{Gaussian process classification}\label{sec::classification_GP}
Many approaches have been proposed in the literature to apply Gaussian processes to classification problems with binary output data defined as $  \lbrace -1, +1 \rbrace$. The goal is to  predict the class membership probability denoted by $p$ corresponding to a test point $\bm{x}^{0}$.
Generally the classification with Gaussian processes requires two stages
\begin{enumerate}
\item A regression step involving a "latent function" $f$ which qualitatively models the likelihood of the input value belonging to a class.
\item A second step which \cite{rasmussen2006gaussian} call "squashing" of the latent function onto $[0,1]$ by using e.g. a sigmoid function $\varphi : \mathbb{R} \rightarrow [0,1]$. Therefore the class membership probability can be written as $ p(y=1 | \bm{x}) = \varphi(f(\bm{x}))$. If the sigmoid function is symmetric it yields $p(y|\bm{x}) = \varphi(y f(\bm{x}))$.
\end{enumerate}
The aim is to compute the predictive distribution
\begin{equation}\label{eq::GPC_predictiveOutput}
p(y_{\star}| \bm{X},\bm{y},\bm{x}_{\star}) = \int p(y_{\star}| f_{\star})  p(f_{\star}| \bm{X},\bm{y},\bm{x}_{\star}) d f_{\star},
\end{equation}
where the marginalization over the latent variables gives
\begin{equation}
p(f_{\star}| \bm{X},\bm{y},\bm{x}_{\star}) = \int p(f_{\star} | \bm{X}, \bm{x}_{\star}, \bm{f}) p(\bm{f} | \bm{X}, \bm{y}) d \bm{f}
\end{equation}
with the posterior over the latent variables given by
\begin{equation}
\begin{aligned}
p(\bm{f}| \bm{X},\bm{y}) = \frac{p(\bm{y | \bm{f}}) p(\bm{f}| \bm{X})}{p(\bm{y}| \bm{X})}.
\end{aligned}
\end{equation}
$p(\bm{y} | \bm{f})$ is the likelihood term, for which generally no closed form exists.  Several numerical approaches have been proposed in the literature to estimate the posterior \citep{williams1998bayesian,dehaene2018expectation,nickisch2008approximations}.

Using Gaussian process classification, the predictive output mean as obtained by equation (\ref{eq::GPC_predictiveOutput}), does not necessarily yield accurate representation for all the available observations but rather tends to average out fringe cases. Therefore, it is proposed to use a classification-oriented but regression-based kriging approach. 

\subsection{kriging classifier}

A kriging surrogate denoted by $\hat{\mathcal{M}}$ is utilized to approximate the  problem $\mathcal{M}$ on the whole parametric space and estimate the surrogate classifier $\hat{\mathcal{C}}$ as
\begin{equation}\label{eq::Ourapproach}
\hat{\mathcal{C}}(\bm{x}) =  \begin{cases}
\mathcal{C}_{1}, & \text{if} \, \hat{\mathcal{M}}(\bm{x}) \geq L \\
\mathcal{C}_{2}, & \text{otherwise}.
\end{cases}
\end{equation}
Kriging appears promising as it ensures an exact representation at the observation points and it generally performs well for low number of observations. Its ability in this context is investigated in comparison with Gaussian process classification.

\subsection{Comparaison between kriging classifier and Gaussian classification}

\begin{figure}[htbp!]
\centering
\begin{subfigure}[t]{0.45\textwidth}
\includegraphics[width=0.9\textwidth]{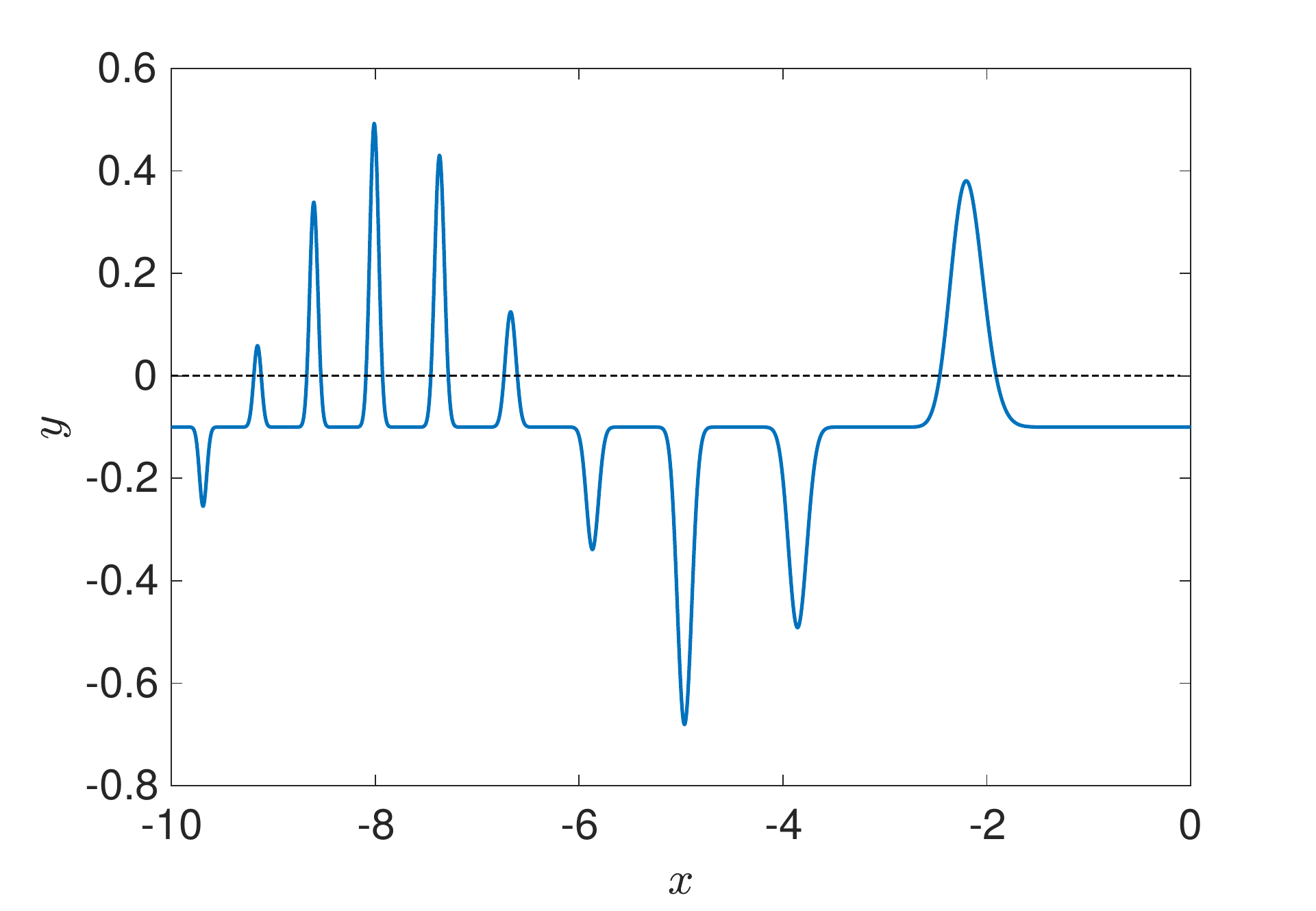} 
\subcaption{Michalewicz function $\mathcal{M}_{M}$}\label{fig::Michal_problem}
\end{subfigure}
\begin{subfigure}[t]{0.45\textwidth}
\includegraphics[width=0.9\textwidth]{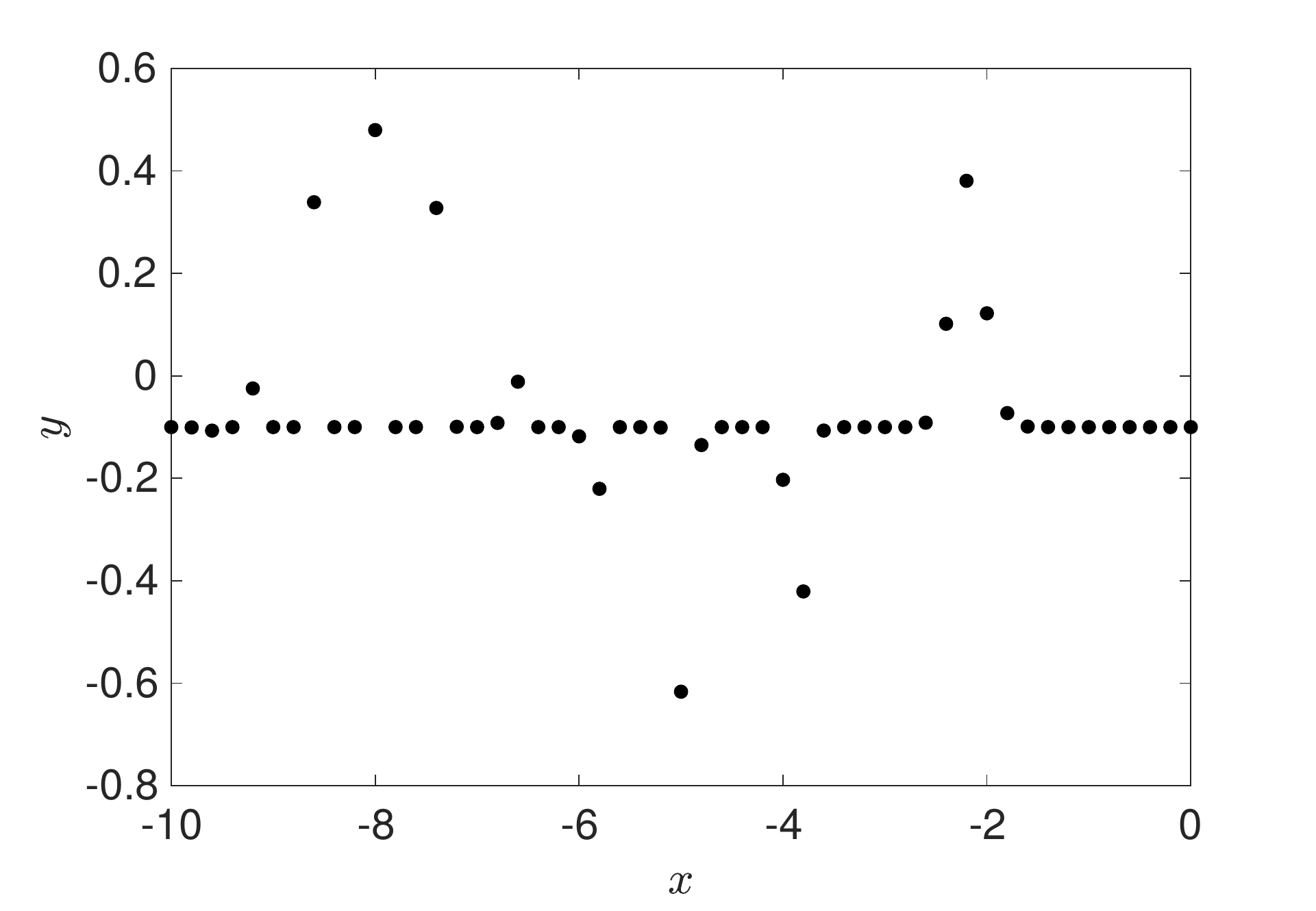}
\subcaption{Dataset}\label{fig::Michal_Samples}
\end{subfigure} \\
\begin{subfigure}[t]{0.55\textwidth}
\includegraphics[width=0.9\textwidth]{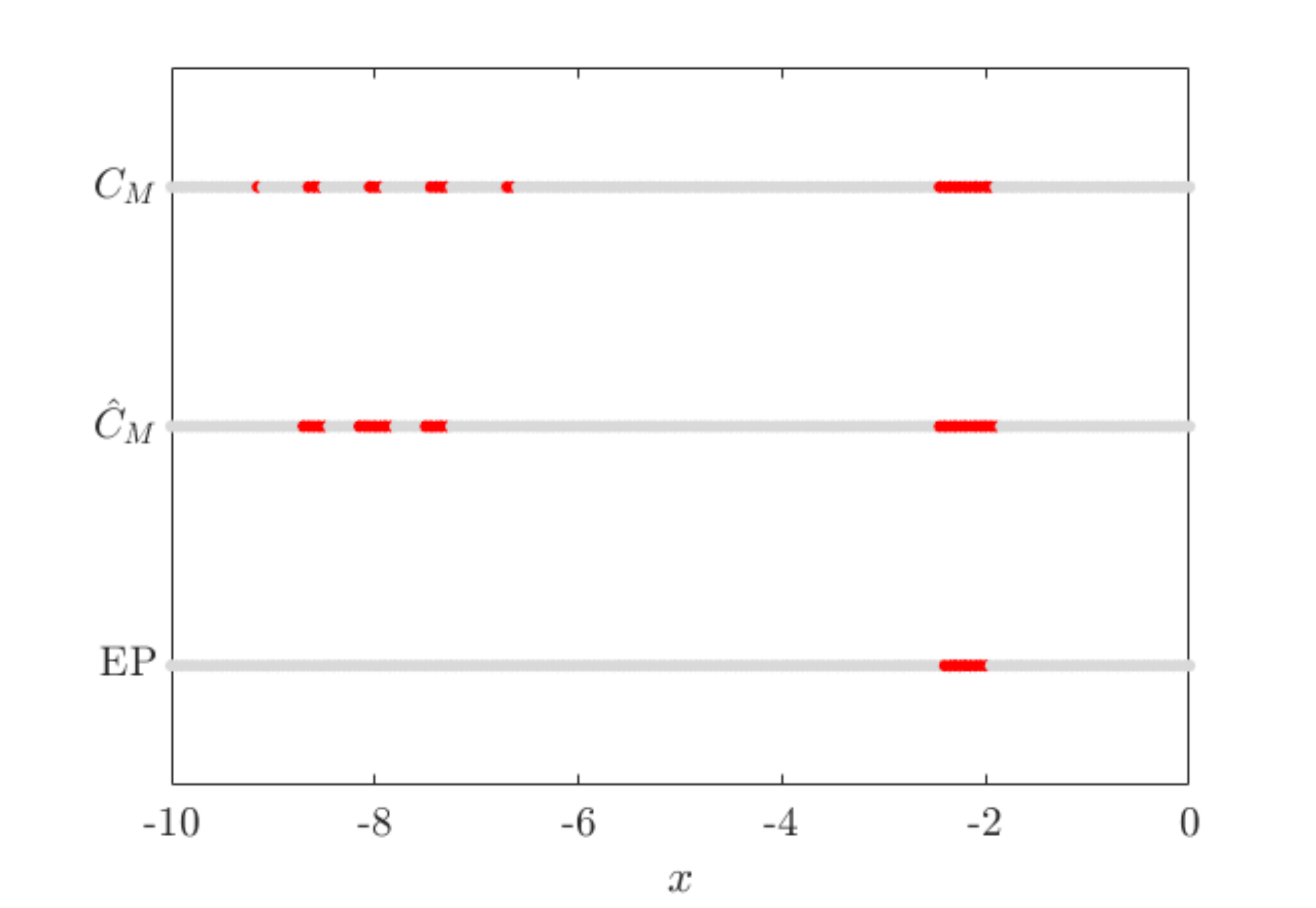}
\subcaption{Classification results}\label{fig::Test_Classif}
\end{subfigure} \\
\begin{subfigure}[t]{0.45\textwidth}
\includegraphics[width=0.9\textwidth]{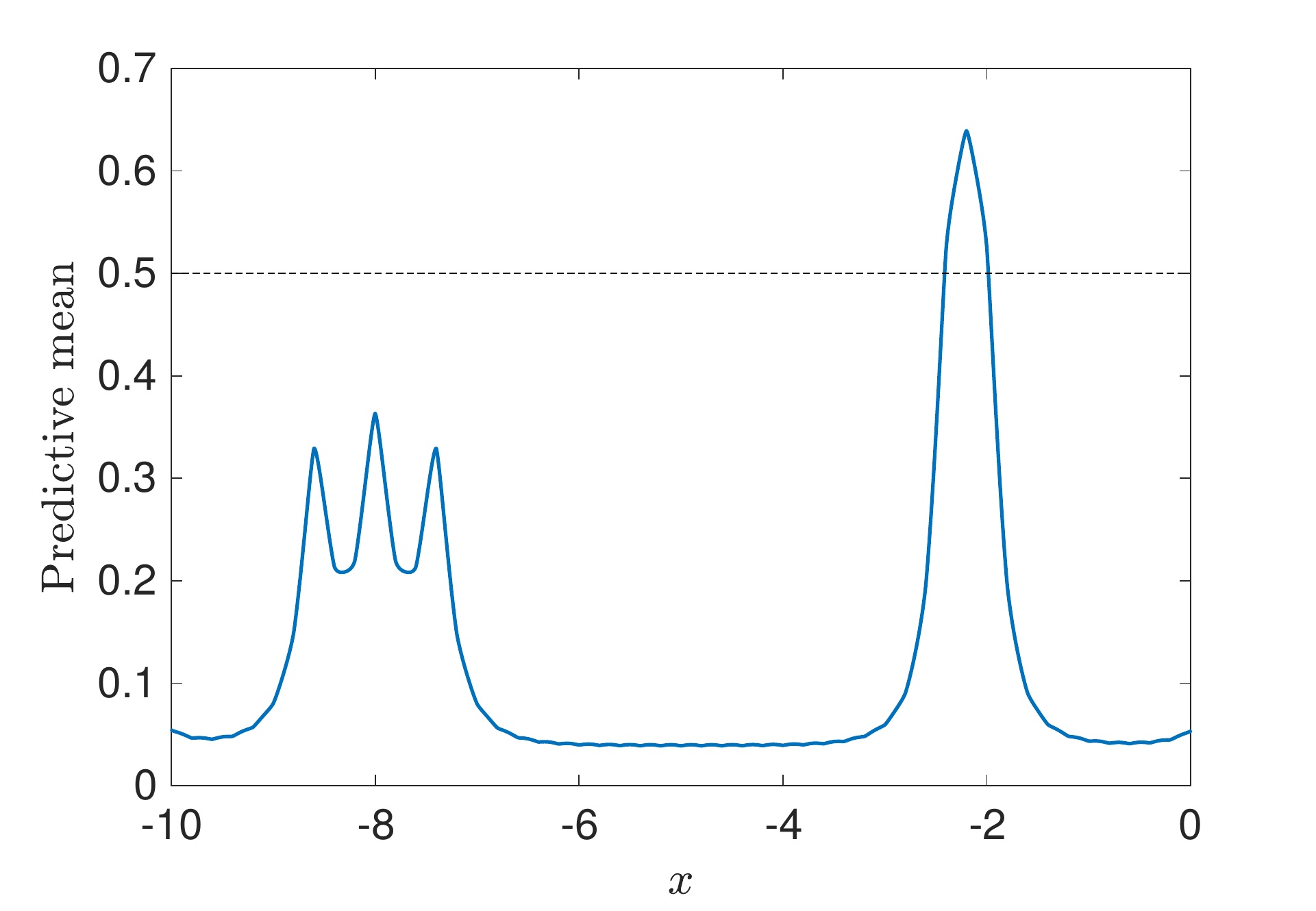} 
\subcaption{Expectation propagation surrogate classifier} \label{fig::Test_EP_mean}
\end{subfigure}
\begin{subfigure}[t]{0.45\textwidth}
\includegraphics[width=0.9\textwidth]{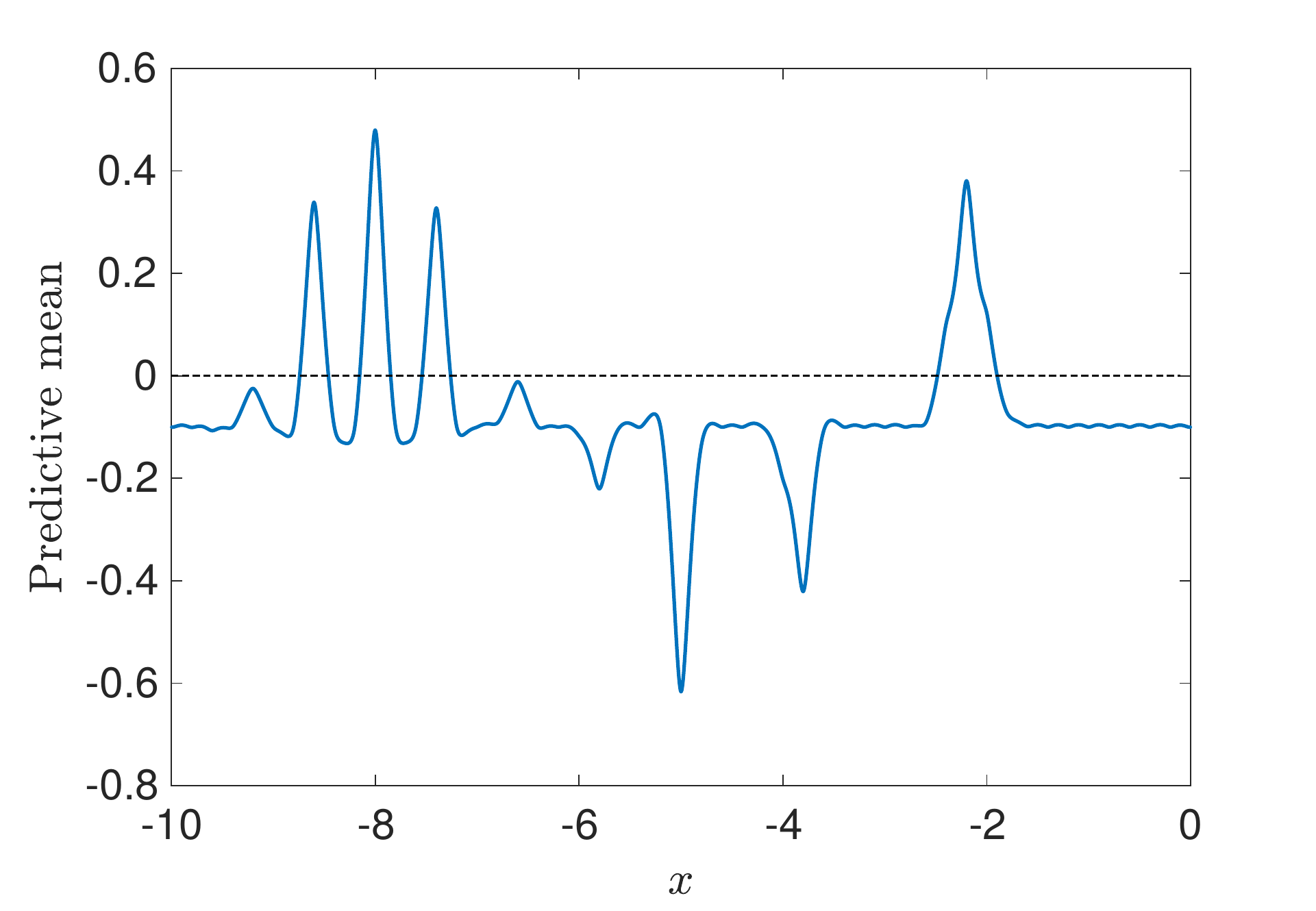} 
\subcaption{Mean $\hat{\mathcal{M}}_{M}$}\label{fig::Test_mean}
\end{subfigure}
\caption{Comparison of Gaussian process classification technique by expectation propagation algorithm $EP$, with a classification surrogate method based on a regression surrogate model $\hat{\mathcal{C}}_M$. }\label{fig_Gauss_classification}
\end{figure}

To compare the kriging classifier and Gaussian process classification, assume $\mathcal{M}$ being given by the Michalewicz function \citep{michalewicz2013genetic} 
with
\begin{equation}
    \mathcal{M}_{M}(\bm{x}) = -0.6 \left( \sin(x) \sin^{20}\left(\frac{i x^{2}}{\pi}\right)\right) - 0.1,
\end{equation}
and $L=0$ as well as $x \in [-10,0]$. The function, which is shown in Figure \ref{fig::Michal_problem}, fluctuates with large local gradients. 

Based on 51 training data points, as displayed in Figure \ref{fig::Michal_Samples}, two surrogate classification models are compared and evaluated:
\begin{enumerate}
    \item a Gaussian process classifier trained with the classified labels $\lbrace \mathcal{C}_{1}, \mathcal{C}_{2} \rbrace = \lbrace -1,1 \rbrace$ as input and consisting of an approximation of (\ref{eq::GPC_predictiveOutput}) using the expectation propagation algorithm as detailed in \cite{minka2001expectation}, which is a commonly utilized classification approach with Gaussian processes \citep{nickisch2008approximations}. The machine-learning toolbox introduced in \cite{rasmussen2010gaussian} is employed.
        \item the kriging classifier $\hat{\mathcal{C}}_{M}$, as described in equation (\ref{eq::Ourapproach}), based on the regression metamodel $\hat{\mathcal{M}}_{M}$ for $\mathcal{M}_{M}$. 
\end{enumerate}

It can be seen that 6 samples out of 51 yield a value that indicates a membership to class $\mathcal{C}_{1}$. We observe how this information is propagated in the surrogate classifiers for both cases. In Figure \ref{fig::Test_Classif}, the reference classifier $\mathcal{C}_{M}$ and the two surrogate classifiers $\hat{\mathcal{C}}_M$ and $\text{EP}$ estimated on 5000 evaluation points are represented. Red dots represent output value belonging to class $\mathcal{C}_{1}$, whereas gray points symbolize $\mathcal{C}_{2}$ membership. The output of the expectation propagation algorithm is a class probability between zero and one, as displayed in Figure \ref{fig::Test_EP_mean}. To be consistent it is chosen to translate the class probability information into membership to either class  $\mathcal{C}_{1}$ or  $\mathcal{C}_{2}$, i.e. if the predicted output mean is larger or equal than $0.5$ the output class is $\mathcal{C}_{1}$, for the complementary condition, the output class is $\mathcal{C}_{2}$. The surrogate Gaussian classification denoted EP as depicted in Figure \ref{fig::Test_Classif} shows a large error comparing to the reference classification. Only one $\mathcal{C}_1$ subdomain is detected, whereas the reference solution comprises six disconnected subdomains. The mean of the Gaussian Process classifier is depicted in Figure \ref{fig::Test_EP_mean}. It can be seen that the mean of the estimator is an average over the observations on the nearby observation points, this appears particularly problematic for the observations in the neighborhood of $x=-8$, where the output value jumps between two classes, leading to a non-detection of the local $\mathcal{C}_1$ subdomains. The surrogate kriging classifier $\hat{\mathcal{C}}_M$ is able to detect four $\mathcal{C}_1$ subdomains, among the six subdomains of the reference estimation. Indeed, since Gaussian process regression is exact at the observations points, the six samples yielding  $\mathcal{C}_1$ class are able to transmit exact local information, as illustrated in Figure \ref{fig::Test_mean}. The kriging classifier appears then more proficient for classification based on highly fluctuating response surfaces. 

It should be mentioned that in this paper it has also been decided to not use the class labels as an output for the kriging classifier $\hat{\mathcal{M}}_{M}$ because numerical issues have appeared with output values showing steep jumps, and results with greater variance of the estimator have been obtained, when trying to train the surrogate model from the class label instead of the numerical response surface values. 

To sum up, a kriging metamodel is built based on regression, but designed for classification purpose. To benefit at best from the information provided by the observations, a dedicated adaptive scheme is designed. It can be noticed that this adaptive sampling technique is introduced for the presented kriging classifier, however, it could generally also be used for any classical Gaussian process classification scheme.

\section{Monte Carlo-intersite Voronoi (MiVor) adaptive scheme}

\label{sec:MIVor}

In general the feature space for binary classification is divided into decision regions (or classification subdomains) in a way that a pattern that falls into the decision region $\mathcal{R}_{i}$ ($i=1,2$) is assigned to class $\mathcal{C}_{i}$ \citep{rasmussen2006gaussian}. Therefore there can be more than one decision region for a specific class.
It is observed that for many problems of interest the classification subdomains cluster together in possibly disconnected subgroups as observed e.g. in \cite{krig_stick_slip} for chaotic/regular motion classification. 

Therefore, adaptive sampling techniques for this purpose should first aim at identifying the different subdomains by chance i.e. randomly scanning the whole parametric domain, and then efficiently sample around some areas of specific interest to accurately demarcate the decision regions. The presented scheme combines \gls{mipt} algorithm \citep{crombecq2011efficient}, which is a proficient sequential explorative scheme and an exploitation contribution based on the analysis of a Voronoi tessellation to proficiently mark out the class boundary. In details, Voronoi tessellation is employed to find the largest cell that indicates a change of class behavior in the parameter space. This cell is then used to feature an exploitation character. Therefore, the innovative hybrid scheme is called \acrfull{mivor}.

During the adaptive sampling scheme as schematized in Figure \ref{fig:adapt_scheme}, all available observations, obtained either as initial data set or as a new experiment through the previous steps of the adaptive scheme, are analyzed. Only binary classification is here considered. The parametric domain comprises two classes $\mathcal{C}_1$ and $\mathcal{C}_2$. Let $n_{1}$ be the number of samples within the class $\mathcal{C}_1$. Define the set $\mathcal{X}_{1}$, that contains all samples indicating the class $\mathcal{C}_1$ behavior. Let $\mathcal{X}_{2} = \mathcal{X} \setminus \mathcal{X}_{1}$ be the set containing the parametric values yielding the complementary behavior. For sake of clarity, it is assumed that $\mathcal{C}_2$ correspond to the largest part of the parametric domain. Therefore, it is expected that most of the \gls{mipt} sample points correspond with that class. As soon as one sample yields an output within the $\mathcal{C}_1$ class, an exploitative contribution is added to the adaptive algorithm using a random switching strategy with decreasing criterion. However, the algorithm remains totally general and could also be used considering first $\mathcal{C}_1$ is explored, and further boundaries between $\mathcal{C}_1$ and $\mathcal{C}_2$ are precisely described.

\subsection{Initial data}

The adaptive sampling procedure begins from some initial sample points obtained either through one-shot or sequential space-filling sampling procedures. 

\gls{lhd} is a commonly used data generation technique \citep{kleijnen2009kriging}. 
Assume that the input space $\mathbb{X}$ is a $\left[0, k-1 \right]^{n}$ hypercube. The
$n$-dimensional \gls{lhd} comprising $k$ points, is a set of $k$ points of the form $x_{i} = \left( x_{i1}, \, \ldots, \,  x_{in} \right) \in \lbrace 0, \, \ldots , \, k-1 \rbrace^{n}$, such that for each dimension $j$ all $x_{ij}$ are distinct \citep{husslage2011space}. The popularity of \gls{lhd} is mainly due to the two following properties:
\begin{enumerate}
\item \gls{lhd} is a space-filling procedure, see \cite{crombecq2011efficient}. This characteristic is particularly interesting when no details of the mapping is available. Hence, it is important to gain information from the entire input space $\mathbb{X}$. To further improve the space-filling property, \gls{lhd} can be combined with the maximin criterion \citep{van2007maximin}.
\item \gls{lhd} is a non-collapsing procedure \citep{husslage2011space}. The collapsing property, see e.g. \cite{janssen2013monte}, describes the fact that when one of the design parameters has almost no influence on the response, then two sample points that are only different in this parameter can be considered as the same point. Hence, they are evaluated twice to create the surrogate model, which creates ill-conditioned matrices for kriging. The non-collapsing property is enforced by \gls{lhd}, which means that after removing one or more parameters the spatial design is still useable.
\end{enumerate} 

A variant of \gls{lhd} called \gls{tplhd} is used to create the initial data set. \gls{tplhd} is obtained via the translational propagation algorithm with a one-point seed \citep{viana2010algorithm}. 
It \label{page::TPLHD} is able to obtain near optimal Latin hypercube
designs without using formal optimization, which leads to less computational effort and so results basically provided in real time.

Involving initial samples requires furthermore an assumption on the number of initial observations considered before starting the adaptive sampling procedure. Compromise is necessary as, on one hand, if the initial sample size is very small, the resulting initial metamodel is of poor quality and the adaptive sampling technique may generate points in unwanted regions of interest due to a lack of exploration performance of the initial data set  \citep{kim2009construction,ghoreyshi2009accelerating}, but on the other hand, considering large initial sample size leads to high computational cost, which could have been avoided by reducing the set and performing pertinent observations from the knowledge of the adaptive machine-learning strategy \citep{crombecq2011novel}.

From the initial set, \gls{mivor} algorithm aims at choosing among a set of Monte Carlo points the most interesting observations to be performed for gaining efficiently knowledge for the surrogate model. Both complementary strategies are included, exploration and exploitation.

\subsection{Exploration component of the \gls{mivor} algorithm based on \gls{mipt}}
\label{sec::MIPT}

The first part of the process is purely exploration-based, i.e. new
samples are added to explore the whole parametric domain. Samples are defined using \gls{mipt} \citep{crombecq2011efficient}, an explorative scheme which has shown its ability to create a proficient space-filling set of points with a low computational cost for various benchmark tests  \citep{Jan_Master_thesis}. 

It is based on a Monte Carlo approach in the input domain to define a set of $n_{MC}$ points $P = \lbrace \bm{p}_{1}, \ldots, \bm{p}_{n_{MC}} \rbrace$ where $\bm{p}_{i} \in \mathbb{R}^{n}$. All Monte Carlo points are evaluated and ranked with respect to their distance to the existing sample points. A distance threshold is defined to create the space-filling character of the method as 
\begin{equation}
d_{min} = \frac{1}{n_{MC}}.
\end{equation}
A new sample is selected by optimizing a discrete optimization problem defined by ranking each point $\bm{p}_{i}$ over the set of Monte Carlo points $P$ according to
\begin{equation}
MIPT(P, \bm{p}) = \begin{cases}
0, & \text{if} \, \min\limits_{\bm{p}_{i} \in P} \norm{\bm{p}_{i} - \bm{p}}_{-\infty} < d_{min}, \\
\min\limits_{\bm{p}_{i} \in P} \norm{\bm{p}_{i} - \bm{p}}_{2}, &\text{if} \, \min\limits_{\bm{p}_{i} \in P} \norm{\bm{p}_{i} - \bm{p}}_{-\infty} \geq d_{min}.
\end{cases}
\end{equation}
Here, $\norm{\bullet}_{2}$ describes the euclidean distance and $\norm{\bullet}_{-\infty}$ is the negative infinity norm defined by  $\norm{\bm{p}}_{-\infty} = \min (p_{i}), \, \forall i \in \left[ 1,n\right] $.

Due to the chosen convention, most of \gls{mipt} points are expected to belong to class $\mathcal{C}_2$. As soon as one \gls{mipt} point is detected belonging to the complementary class, the exploitation part of the algorithm is activated to feature an adaptive scheme aiming at describing precisely the boundary between the two classes.

\subsection{Exploitation-based adaptive step}
\label{sec:exploitation}

It has been observed that the boundaries between two classes need to be sampled sufficiently in order to create a proficient metamodel for binary classification. Therefore the exploitation component of \gls{mivor} aims to sample close to these edge, based on the following ideas:
\begin{itemize}
\item The volume fraction of the Voronoi cells corresponding with samples $\bm{x}^{i}$ of $\mathcal{C}_1$ behavior provides an information about the `density' of samples relative to that class. The larger the volume of cells corresponding with it the more uncertain is the $\mathcal{C}_1$ subdomain around this sample. Therefore, a new observation point needs to be added to feature more precisely the limits between both classes.
\item Within the neighborhood of a sample with $\mathcal{C}_1$ behavior $\bm{x}^{i} \in \mathcal{X}_{1}$, more points are in the set of $\mathcal{C}_2$ behavior $\mathcal{X}_{2}$, closer $\bm{x}^{i}$ is to an edge.
\end{itemize}

The general procedure of the exploitation step combines both of these ideas. First the parametric space is divided into cells by Voronoi tessellation. 

\subsubsection{Voronoi tessellation}
\label{sec::Voro}

The input parametric space is tessellated by a set of $m$ cells $\lbrace Z_{1}, \ldots, Z_{m} \rbrace$ around the existing $m$ sample points by employing the well-known Voronoi tessellation \citep{aurenhammer1991voronoi}. The so-called dominance function is used to define a set of points relative to cell $i$ with respect to another cell $j$
\begin{equation}
dom(\bm{x}^{i}, \bm{x}^{j} ) = \lbrace \bm{x} \in \mathbb{R}^{n} | \norm{\bm{x}- \bm{x}^{i}} \leq \bm{x - \bm{x}^{j}} \rbrace, \, \forall (i,j) \in (1,m)^2, \, \text{with} \, j \neq i.
\label{eq:dom}
\end{equation}
Thus, a point $\bm{x}$ belongs to the cell relative to $\bm{x}^{i}$ if it is at least as close to $\bm{x}^{i}$ as to any other sampled points $\left\lbrace \bm{x}^{j}\right\rbrace_{\substack{j\in \left[1,m\right] \\ j \neq i}}$, i.e. all the points for which $\bm{x}^{i}$ is dominant over $\bm{x}^{j} $. Then, the Voronoi cell corresponding to the point $\bm{x}^{i}$ is defined as
\begin{equation}
Z_{i} = \underset{\bm{x}^{j} \in \mathcal{X} \setminus \bm{x}^{i}} \cap dom(\bm{x}^{i}, \bm{x}^{j}).
\end{equation}
The computation of the Voronoi tessellation requires high computational effort especially in higher dimensions \citep{crombecq2011novel}. However as shown in \cite{crombecq2011novel} the volume of each corresponding cell can be estimated by employing the Monte Carlo method. The volume integration process of the Voronoi cells relative to the existing $m$ sample points is summarized in Box \ref{alg::EstimationVolume}. First the parametric space is randomly filled with Monte Carlo points. It can be noticed that the same set of points previously generated as sample candidates can easily be reused here. Then, a Monte Carlo point is considered in the influence zone of point $\bm{x}^{i}$ using definition (\ref{eq:dom}). Intuitively, the Voronoi cell with the largest volume is the one with the highest number of randomly sampled points in its influence domain. The volume of each Voronoi cell can therefore be estimated from the amount of points in each cell, as detailed in Box \ref{alg::EstimationVolume}. This method provides an effective numerical tool to calculate the fraction of the parametric volume $Vol_{i}, i=1, \ldots,n$ corresponding with the Voronoi cell of each sample point of the dataset.
\begin{kasten}[ht!]
\begin{Algorithmus}
\begin{itemize}
\itemsep0.5em 
\item Initial sample data given by $\mathcal{X}= \lbrace \bm{x}^{1}, \ldots, \bm{x}^{m} \rbrace$. \item Sample $n_{MC}$ Monte Carlo points in the input domain and define $P = \lbrace \bm{p}_{1}, \ldots, \bm{p}_{n_{MC}} \rbrace$.
\item Define $\bm{Vol} = \lbrace Vol_{i}, \, i=1,\ldots, m \rbrace$. Initially set all $Vol_{i}=0$.
\item[] for $\bm{p} \in P$ 
\begin{itemize}[topsep=-50px,partopsep=0px]
\item[] Find closest point $\bm{x}^{i}$ for $p$ in $\mathcal{X}$ with $$\bm{x}^{i} =  \min_{\bm{x}^{\star} \in \mathcal{X}} \norm{ \bm{p} - \bm{x}^{\star}}_{2}.$$
\item[] Set $Vol_{i} = Vol_{i} + 1$.
\end{itemize}
end
\item Set $Vol_{i} = \frac{Vol_{i}}{\norm{\bm{Vol}}_{1}}$ for all $i=1, \ldots m$.
\end{itemize}
\end{Algorithmus}
\captionof{kasten}{Estimation of normalized Voronoi cell volume using Monte Carlo method}\label{alg::EstimationVolume}
\end{kasten}

\subsubsection{Ranking of Voronoi cells}

Then the sample points with $\mathcal{C}_1$ behavior are ranked through a score $R$ defined as 
\begin{equation}\label{eq::SamplePointRank}
\forall \bm{x}^{i} \in \mathcal{X}_{1}, \qquad R(\bm{x}^{i}) = Vol_{i} \, \mathcal{N}_{i},
\end{equation}
i.e. a combination of the volume of the corresponding Voronoi cell and the number of points $\mathcal{N}_{i}$ belonging to the set $\mathcal{X}_{2}$ in the neighborhood of $\bm{x}^{i}$, which is defined as the set of $2 \times n$ nearest points by euclidean distance of the set $\mathcal{X} \setminus \bm{x}^{i}$, with $n$ the dimension of $\bm{x}^{i}$. 

\begin{figure}[ht!]
\centering
\begin{subfigure}[t]{0.49\textwidth}
\includegraphics[width=\textwidth]{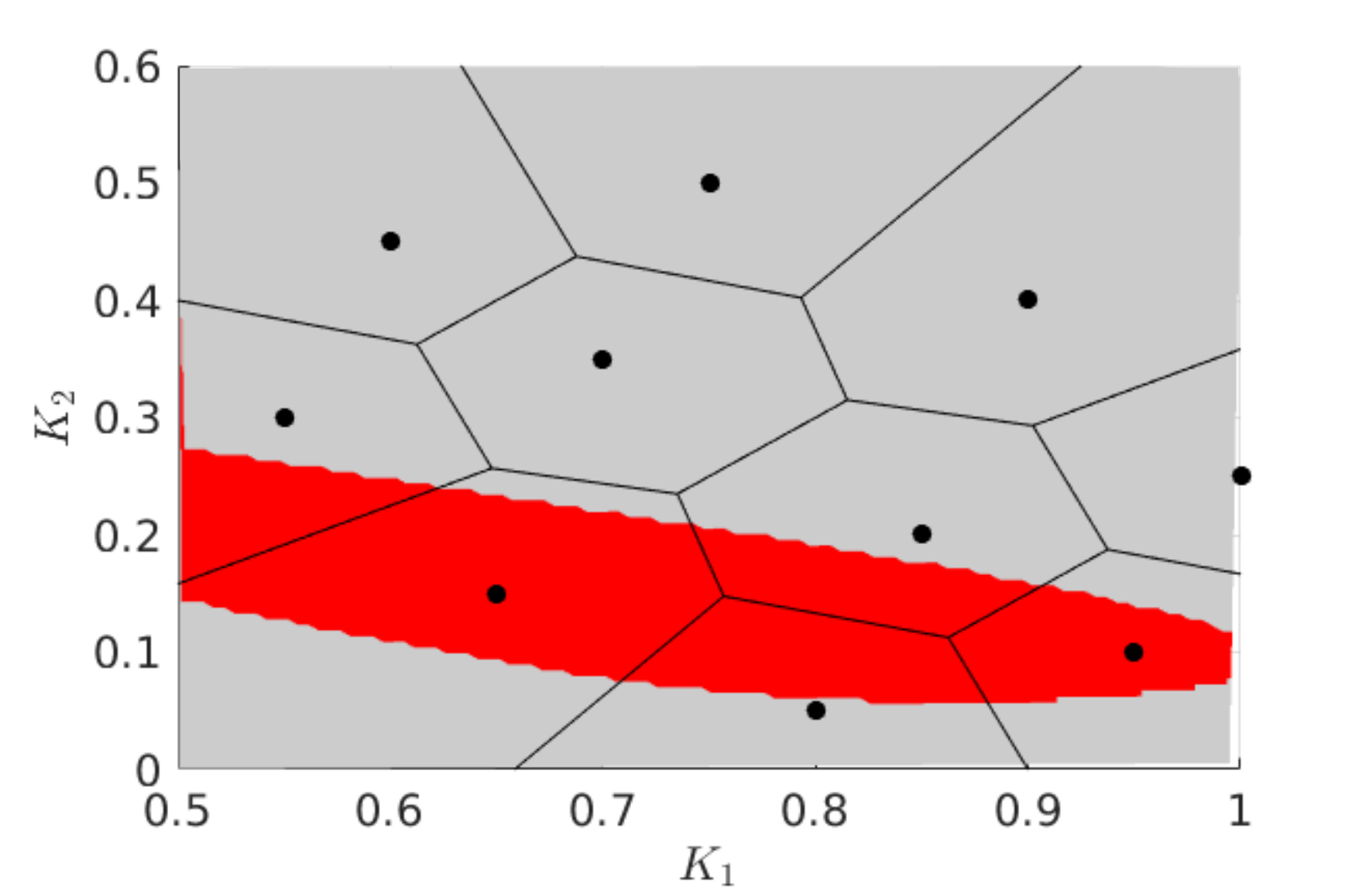} 
\subcaption{Existing samples with their corresponding Voronoi cells}\label{fig:MIVOINITIAL}
\end{subfigure}%
\begin{subfigure}[t]{0.49\textwidth}
\includegraphics[width=\textwidth]{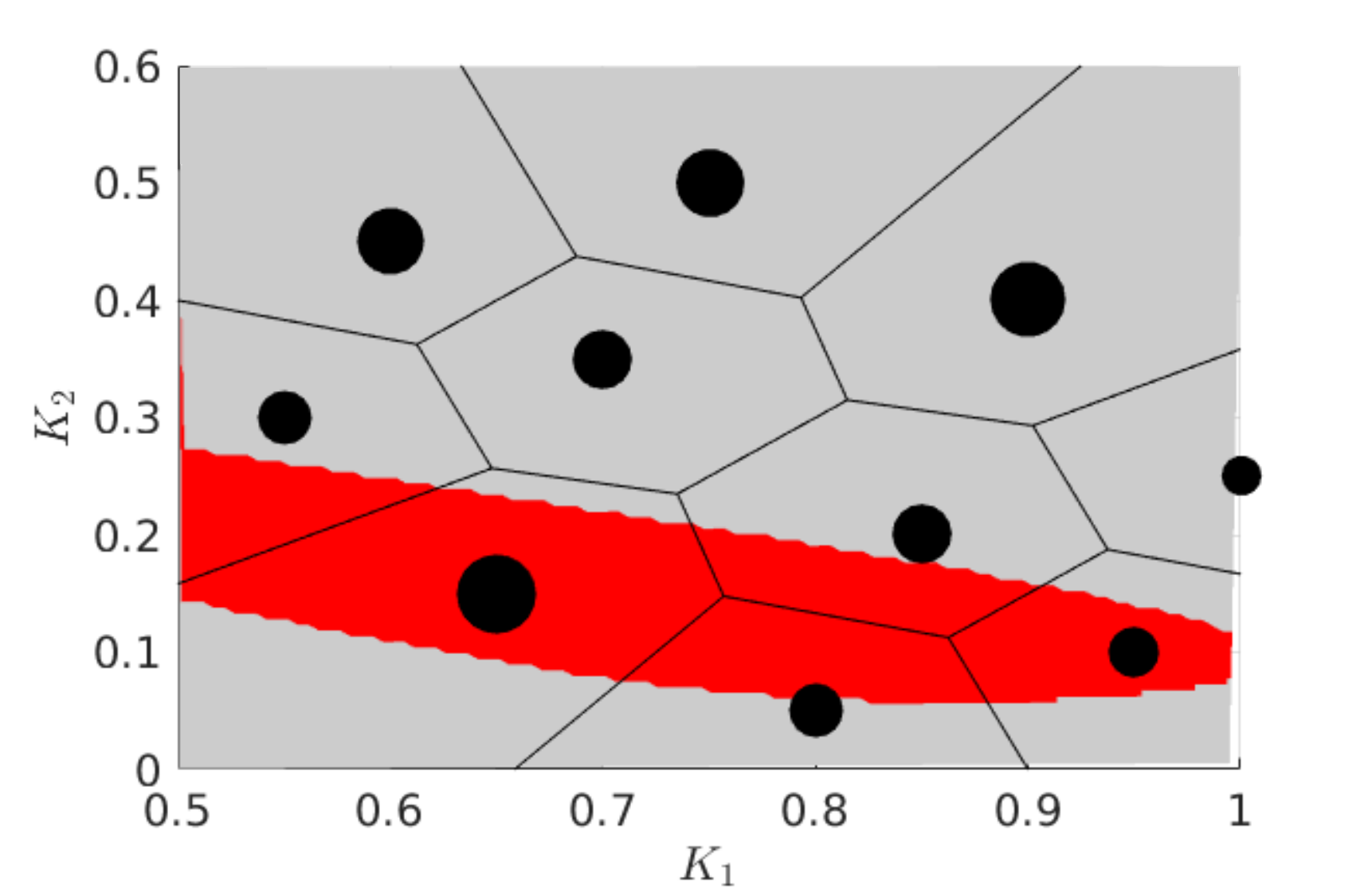} 
\subcaption{Area of sample points proportional to the volume of the corresponding Voronoi cell}\label{fig:MIVOVolume}
\end{subfigure}
\begin{subfigure}[t]{0.49\textwidth}
\includegraphics[width=\textwidth]{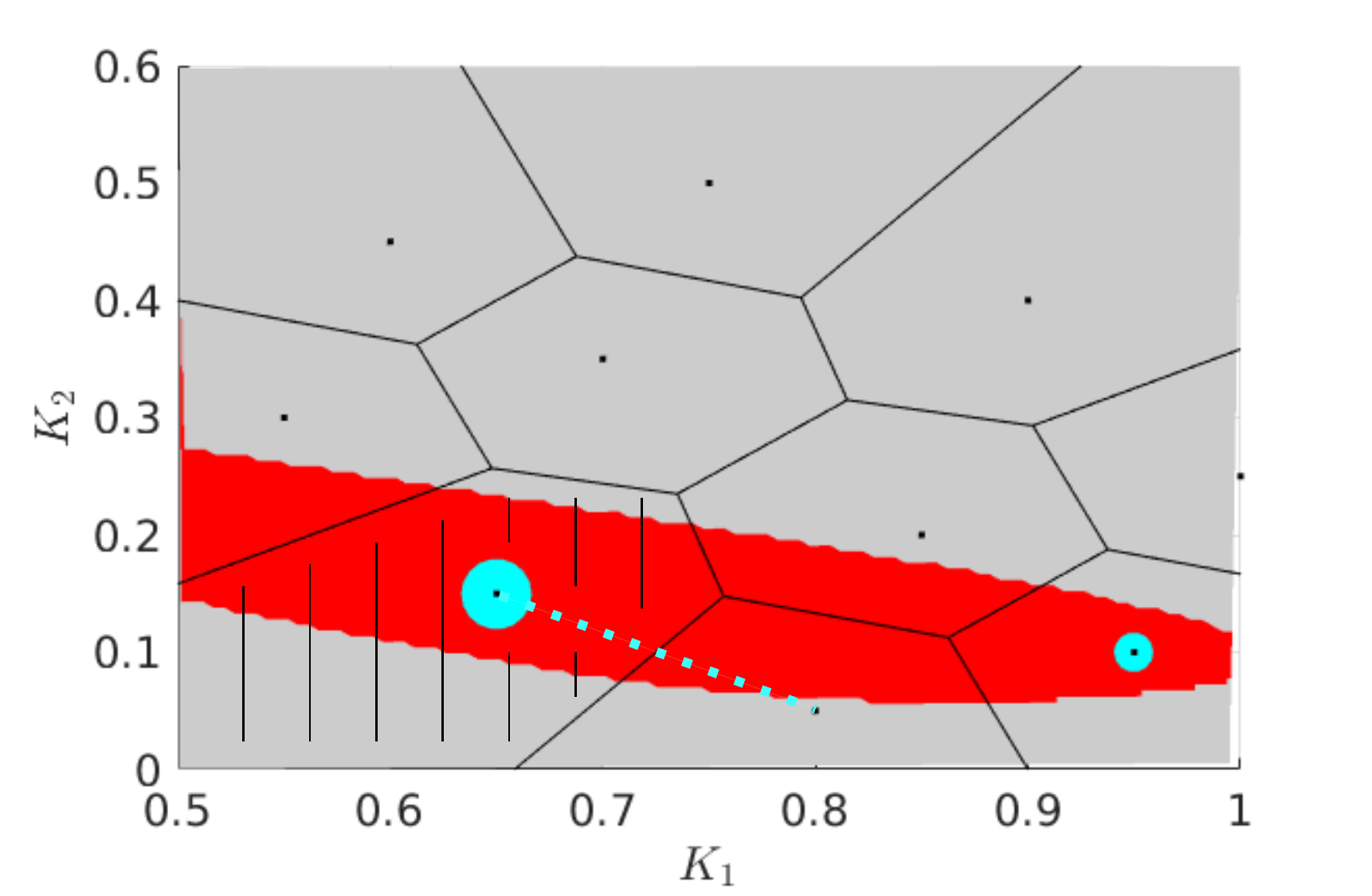} 
\subcaption{Size of $\mathcal{C}_1$ samples proportional to their Score value (in cyan). Dotted line indicates the closest point to the highest scoring sample}\label{fig:MIVOScore}
\end{subfigure}%
\begin{subfigure}[t]{0.49\textwidth}
\includegraphics[width=\textwidth]{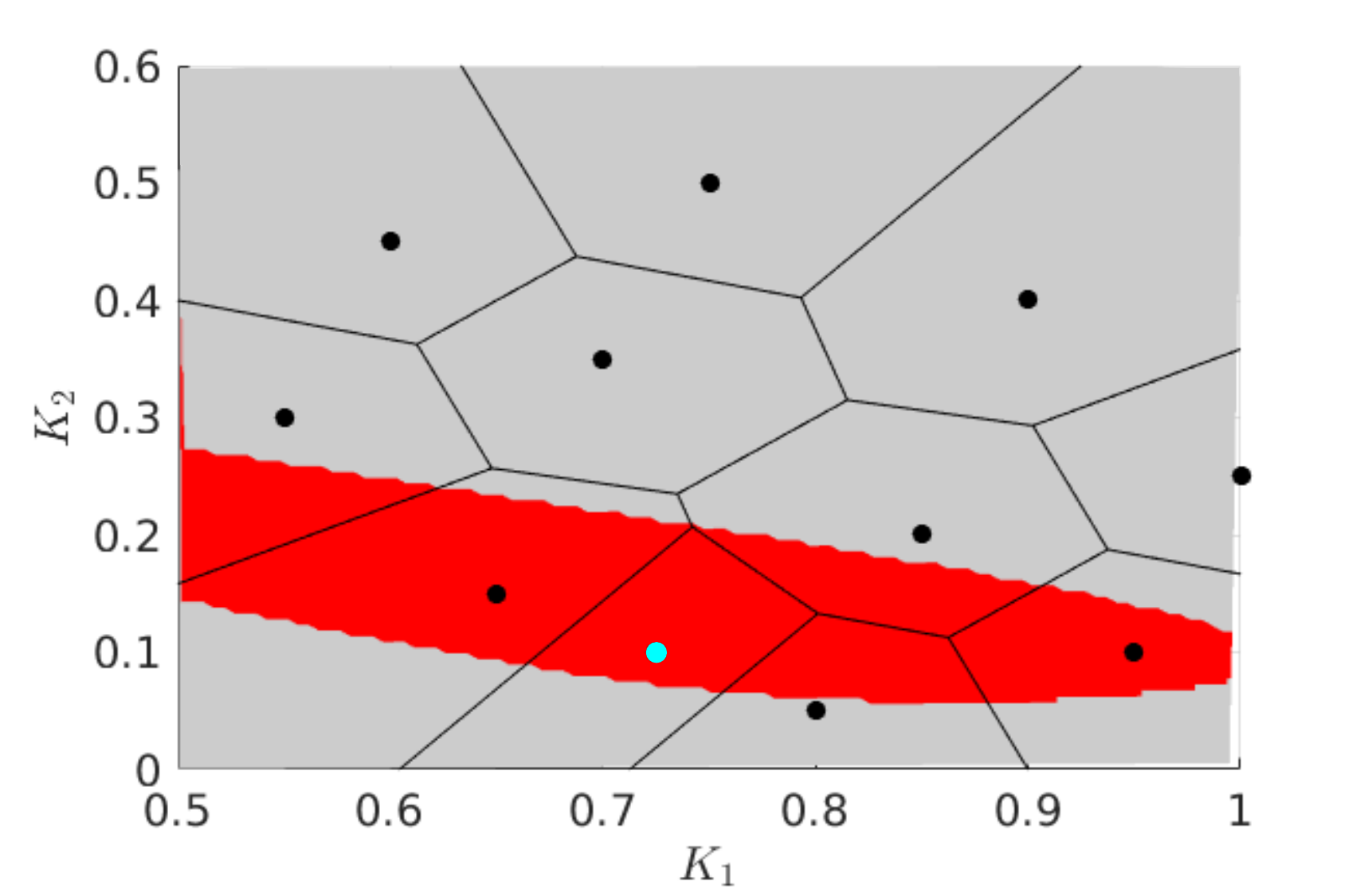} 
\subcaption{New point sampled near to the boundary between the two cells on the dotted line of (c)}\label{fig:MIVONewPoint}
\end{subfigure}
\caption[Process of sampling a new point with \gls{mivor}]{Process of sampling a new point with \gls{mivor} over the contour of a minor class (represented in red) embedded in a regular behavior (represented in grey).}\label{fig:MIVORONEStep}
\end{figure}

The steps of the procedure are illustrated in Figure \ref{fig:MIVORONEStep}, where the $\mathcal{C}_1$ behavior is indicated by the red color domain, whereas the class $\mathcal{C}_2$ is represented by the grey subdomain. First, from 10 initial samples, the Voronoi tessellation provides a set of cells as shown in Figure \ref{fig:MIVOINITIAL}. It can be observed that two points correspond with class $\mathcal{C}_1$. So, the exploration part of the algorithm is activated if the realization of $u$ is larger than $r$. Thus the volumes of the Voronoi cells are evaluated using a Monte Carlo approach (see Box \ref{alg::EstimationVolume}). For sake of illustration, Figure \ref{fig:MIVOVolume} shows the areas of the sample points proportional to the volume of the relative cell. As a sidenote, all the cell volumes are represented here, however in the adaptive algorithm only the volume relative to the two cells with minor behavior would indeed be evaluated. The most crucial Voronoi cell denoted by $Z_{max}$ is defined as the cell around the existing sample point with highest score
\begin{equation}
\bm{x}^{max} = \max_{\bm{x}^{\star} \in \mathcal{X}_{1}} R(\bm{x}^{\star}).
\end{equation}
In the example illustrated in Figure~\ref{fig:MIVOScore}, only two cells have to be ranked from their score values, their relative importance has been schematized by the cyan area around them. The Monte Carlo points corresponding to the cell $Z_{max}$ define the candidate set $P_{max} \subset P$. In order to estimate the location of the edge between the two classes as accurately as possible the Monte Carlo point in the set $P_{max}$ that is closest to a sample point of class $\mathcal{C}_2$ is taken as candidate denoted $\bm{x}^{cand}$ for the next sample point
\begin{equation}
\bm{x}^{cand} = \min_{\substack{\bm{p}  \in P_{max}\\ \forall \bm{x}^{i} \in \mathcal{X}_{2}}} \norm{\bm{p} - \bm{x}^{i}} \, .
\end{equation}

For instance, in Figure \ref{fig:MIVONewPoint}, the candidate point can be seen. It has been chosen as the Monte Carlo point in the Voronoi cell of the sample with the highest score, that is the closest to the regular neighbor. Through this process it is expected that this point is close to the boundary between the two class regions. 

Even if a nugget term, see \cite{booker1999rigorous}, is included in the algorithm, it is preferred to enhance the quality of sampling by preventing clustering between sample points.

\subsubsection{Prevention of kriging point clustering during exploitation step}\label{sec::Clustering}

 In order to avoid clustering of sample points and hence numerical issues, a control distance, which needs to be exceeded between two samples before accepting the candidate point, is included in the design of experiments. 
For this the space-filling metric $S$ is introduced as
\begin{equation}
S = 0.1 \max\limits_{\forall \bm{x}^{i} \in \mathcal{X}} (ds(\bm{x}^{i}) ),
\end{equation}
with
\begin{equation}
ds(\bm{x}^{i}) = \min\limits_{\forall \bm{x}^{i} \in \mathcal{X} \cap (i \neq j)} \left( \norm{\bm{x}^{i} - \bm{x}^{{j}}} \right)  .
\end{equation}
If the candidate point $\bm{x}^{cand}$ is not closer than $S$ to an existing sample point, i.e.
\begin{equation}
\forall \, \bm{x}^{i} \in \mathcal{X}, \qquad \norm{ \bm{x}^{cand} - \bm{x}^{i} } > S,
\label{eq:condition_point}
\end{equation} 
then, this candidate point is accepted, a new experiment is performed and the surrogate model is updated to benefit from new information. If the condition (\ref{eq:condition_point}) is not fulfilled, the candidate point is rejected, and a new candidate point referred to as substitute point is contemplated. The substitute point $\bm{x}^{subs}$ is the Monte Carlo point within the set $P_{max}$ with the highest prediction variance as defined by Equation (\ref{eq:var_kriging})
\begin{equation}
\bm{x}^{subs} = \max_{\bm{p} \in P_{max}} \, \sigma_{\hat{y}_{\star}}^{2}(\bm{p}).
\end{equation}
 Its distance admissibility is checked through Equation (\ref{eq:condition_point}) considering now the substitute point instead of the candidate point. If the condition is fulfilled, then the substitute point is accepted as the new sample point. If $\bm{x}^{subs}$ violates also the distance constraint, \gls{mipt} as defined in Section \ref{sec::MIPT} is employed to define the new point of observation.
 
\begin{figure}[ht!]
\tikzstyle{decision} = [rectangle, draw,  
    text width=20em, text centered, rounded corners, node distance=2cm, inner sep=0pt,minimum height=3em]
\tikzstyle{block} = [rectangle, draw, fill=blue!20, 
    text width=20em, text centered, rounded corners, minimum height=4em,node distance=4cm,]
\tikzstyle{blockA} = [rectangle, draw, fill=blue!10, 
    text width=20em, text centered, rounded corners, minimum height=3em]
\tikzstyle{line} = [draw, -latex']
\tikzstyle{cloud} = [draw, ellipse,fill=red!20, node distance=8.0cm,
    minimum height=2em]
 \centering 
\begin{tikzpicture}[node distance = 1.5cm, auto,scale=0.8, transform shape]
    \node [block] (init) {
\begin{tabular}{lr}
Candidate point & $\bm{x}^{cand}$ \\
Space-filling metric & $S$ \\
Set of Monte Carlo points & $P_{max}$ \\
Set of samples & $\mathcal{X}$
\end{tabular}
};
    \node [decision, below of=init,node distance=3.0cm] (start) {Check if: \\
    $\exists \, \bm{x}^{i} \in \mathcal{X} \, \text{with} \, \norm{\bm{x}^{i} - \bm{x}^{cand}}< S$};
    
 \node [cloud, right of=start] (Done1) {$\bm{x}^{m+1} = \bm{x}^{cand}$};    
    
    \node [blockA, below of=start,node distance=2.5cm] (Evaluate) {$\bm{x}^{subs} = \max\limits_{\bm{p} \in P_{max}} \, \sigma_{\hat{Y}_0}^{2}(\bm{p})$  };
    
   \node [decision, below of=Evaluate,node distance=2.5cm] (Check1) {Check if: \\
    $\exists \, \bm{x}^{i} \in \mathcal{X} \, \text{with} \, \norm{\bm{x}^{i} - \bm{x}^{subs}}< S$};

     \node [cloud, right of=Check1] (Done2) {$\bm{x}^{m+1} = \bm{x}^{subs}$};    
    
    \node [cloud, below of=Check1,node distance=2.0cm] (Update) {Find $\bm{x}^{m+1}$ with MIPT.};
  
    \path [line] (init) -- (start);
     \path [line] (start) -- node {no}(Done1);
     \path [line] (start) -- node {yes}(Evaluate);
    \path [line] (Evaluate) -- (Check1);
      \path [line] (Check1) -- node {no}(Done2);
     \path [line] (Check1) -- node {yes}(Update);

\end{tikzpicture}
\caption[Workflow for acceptance of a \gls{mivor} candidate point.]{Workflow for acceptance of a \gls{mivor} candidate point as a new sample point $\bm{x}^{m+1}$ to prevent point clustering.}\label{fig::flowchart_MIVOr_Acceptance}
\end{figure}
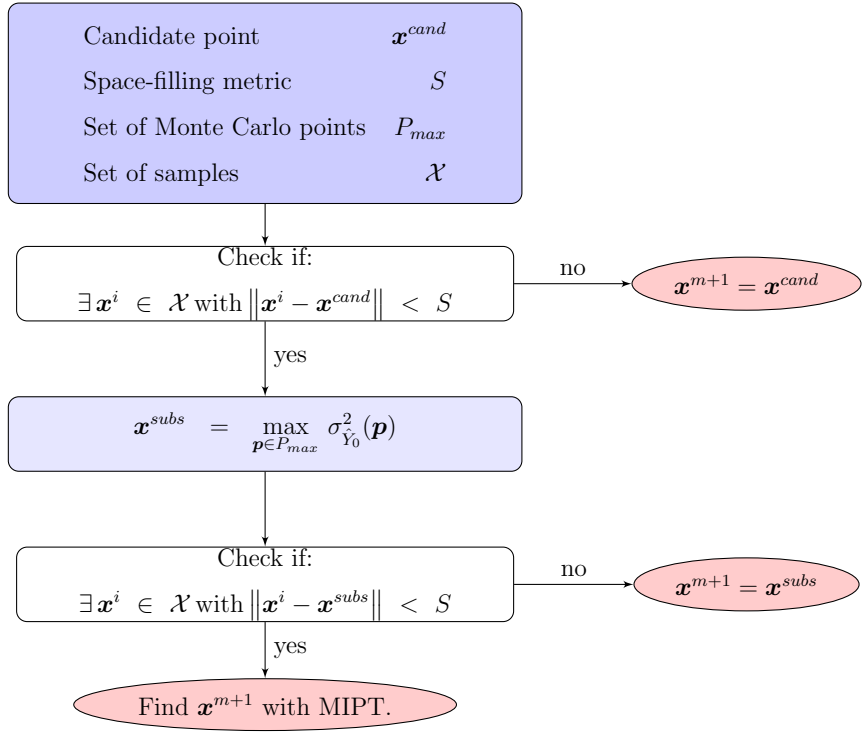

The workflow for acceptance of the candidate point is summarized in Figure~\ref{fig::flowchart_MIVOr_Acceptance}.

The goal is to feature an adaptive scheme able to both detect eventually disconnected subdomains relative to class $\mathcal{C}_1$, ans also describe precisely the boundary between the two classes.

\subsection{Random switching strategy with decreasing criterion to balance exploration and exploitation}

\begin{figure}[ht!]
\centering
\begin{subfigure}[t]{0.33\textwidth}
\begin{tikzpicture}[scale = 0.55, transform shape]
\begin{axis}
[
 axis equal,
xlabel={Iteration},
ylabel={Weights},
y label style={anchor=north},
axis lines=middle,
xtick={0},
xticklabels={},
    ytick={0},
    ymin=-0.1,ymax=1.1,
    xmin=-0.1,xmax=1.1,
    axis on top=false,
every axis x label/.style={
    at={(ticklabel* cs:0.5)},
    anchor=north,
},
every axis y label/.style={
    at={(ticklabel* cs:1.01)},
    anchor=south,
},
legend style={at={(axis cs:1.0,0.5)},anchor=west}
]
\draw[dashed] (axis cs:0.0,1.0) -- (axis cs:1.0,1.0);
\addplot [black, nodes near coords=$1.0$,every node near coord/.style={anchor=0},forget plot] coordinates {( 0, 1)};
\addplot [domain=0.0:1.0,red,line width = 0.5mm]{tanh(2.8*x)};
\addplot [domain=0.0:1.0,dashed,blue,line width = 0.5mm]{1-tanh(2.8*x)};
\end{axis}
\end{tikzpicture}
\subcaption{Decreasing strategy}\label{fig:Ex_ex_decreasing_strategy}
\end{subfigure}%
\begin{subfigure}[t]{0.33\textwidth}
\begin{tikzpicture}[scale = 0.55, transform shape]
\begin{axis}
[
 axis equal,
xlabel={Iteration},
ylabel={Weights},
y label style={anchor=north},
axis lines=middle,
xtick={0},
xticklabels={},
    ytick={0},
    ymin=-0.1,ymax=1.1,
    xmin=-0.1,xmax=1.1,
    axis on top=false,
every axis x label/.style={
    at={(ticklabel* cs:0.55)},
    anchor=north,
},
every axis y label/.style={
    at={(ticklabel* cs:1.01)},
    anchor=south,
},
]
\draw[dashed] (axis cs:0.0,1.0) -- (axis cs:1.0,1.0);
\addplot [black, nodes near coords=$1.0$,every node near coord/.style={anchor=0}] coordinates {( 0, 1)};
\draw[red, line width = 0.5mm] (axis cs:0.0,0.0) -- (axis cs:0.3,0.0);
\draw[red,line width = 0.5mm] (axis cs:0.3,0.0) -- (axis cs:0.36,1.0);
\draw[red,line width = 0.5mm] (axis cs:0.36,1.0) -- (axis cs:0.63,1.0);
\draw[red,line width = 0.5mm] (axis cs:0.63,1.0) -- (axis cs:0.7,0.0);
\draw[red,line width = 0.5mm] (axis cs:0.7,0.0) -- (axis cs:1.0,0.0);

\draw[dashed,blue,line width = 0.5mm] (axis cs:0.0,1.0) -- (axis cs:0.3,1.0);
\draw[dashed,blue,line width = 0.5mm] (axis cs:0.3,1.0) -- (axis cs:0.36,0.0);
\draw[dashed,blue,line width = 0.5mm] (axis cs:0.36,0.0) -- (axis cs:0.63,0.0);
\draw[dashed,blue,line width = 0.5mm] (axis cs:0.63,0.0) -- (axis cs:0.7,1.0);
\draw[dashed,blue,line width = 0.5mm] (axis cs:0.7,1.0) -- (axis cs:1.0,1.0);

\end{axis}
\end{tikzpicture}
\subcaption{Greedy strategy}\label{fig:Ex_ex_greedy_strategy}
\end{subfigure}%
\begin{subfigure}[t]{0.33\textwidth}
\begin{tikzpicture}[scale = 0.52, transform shape]
\begin{axis}
[
xlabel={Iteration},
ylabel={Weights},
y label style={anchor=north},
axis lines=middle,
xtick={0},
xticklabels={},
    ytick={0},
    ymin=-0.1,ymax=2.1,
    xmin=-0.1,xmax=1.1,
    axis on top=false,
every axis x label/.style={
    at={(ticklabel* cs:0.5)},
    anchor=north,
},
every axis y label/.style={
    at={(ticklabel* cs:1.01)},
    anchor=south,
},
legend style={at={(axis cs:1.0,0.7)},anchor=west}
]
\draw[dashed] (axis cs:0.0,2.0) -- (axis cs:1.0,2.0);
\addplot[domain=0:1.0,dashed,blue,line width = 0.5mm]{cos(deg(9*x-(pi/0.5)))+1};%
\addplot[domain=0:1.0,red,line width = 0.5mm]{sin(deg(9*x-pi/2))+1};
\addlegendentry{$w_{local}$}
\addlegendentry{$w_{global}$}
\addplot [black, nodes near coords=$1.0$,every node near coord/.style={anchor=0}] coordinates {( 0, 2.0)};
\end{axis}
\end{tikzpicture}
\subcaption{Switch strategy}\label{fig:Ex_ex_switch_strategy}
\end{subfigure}
\caption[Adaptive strategies to balance local exploitation and global exploration]{Adaptive strategies to balance local exploitation and global exploration.}%
\label{fig:Ex_ex_strategy}%
\end{figure}
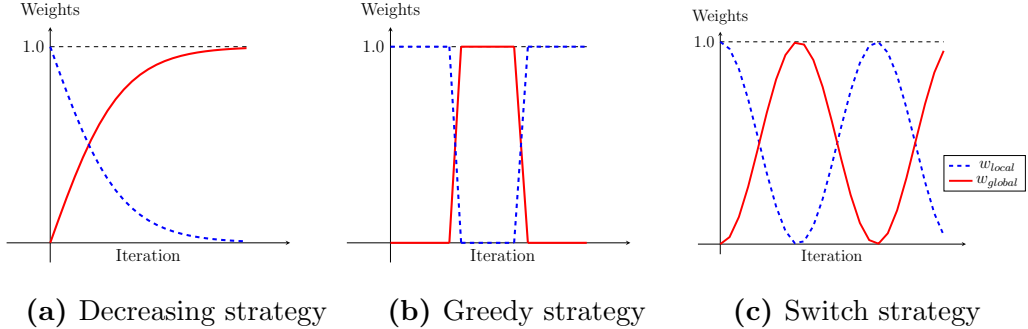

Some established adaptive strategies to balance global exploration and local exploitation based on the weights $\omega_{local}$ and $\omega_{global}$ corresponding with local and global scores, respectively, are represented in Figure \ref{fig:Ex_ex_strategy}. In all the cases the local and global weights initially have the values $(\omega_{global},\omega_{local}) = (1.0,0.0)$. Using a decreasing strategy, the local weight decreases with the iterations while the global weight increases until reaching the values $(\omega_{global},\omega_{local}) = (0.0,1.0)$ (see Figure \ref{fig:Ex_ex_decreasing_strategy}). In case of a greedy algorithm, the local and global weights oscillate between the two values, either $0.0$ or $1.0$ (see Figure \ref{fig:Ex_ex_greedy_strategy}). The switch strategy allows the local and global weights to have any value between $ 0.0$ and $1.0$, and decreasing as well as increasing evolution while satisfying the condition $\omega_{global}+\omega_{local}=1.0$ (see Figure \ref{fig:Ex_ex_switch_strategy}).

In this paper an innovative, randomized approach is proposed, which can be described as a combination of decreasing and switch strategies. In details, initially an exploration rate $r_0 \in \left[ 0 , 1\right]$ is chosen by the user as well as a decreasing factor $\alpha>1.0$. The value of the exploration rate denoted by $r$ will evolve from the initial value $r_0$ in the course of the adaptive sampling process.

For any step, from one sample value $u$ of a uniformly distributed random variable $ U[0,1]$, the exploration or exploitation character of the relative step of the adaptive scheme is chosen as follows:
\begin{itemize}
\item if $u< r$, the new observation is decided through exploration, by employing the \gls{mipt} algorithm previously presented in Section \ref{sec::MIPT}, and the exploration rate is divided by the decreasing factor $\alpha$. 
\item if $u \geq r$, the new sample is obtained by exploitation as proposed in Section \ref{sec:exploitation}, i.e. the step aims at sampling a new point close to a predicted edge between two decision regions. 
\end{itemize}

\begin{figure}[htbp!]
\centering
\begin{subfigure}[t]{0.49\textwidth}
\includegraphics[width=\textwidth]{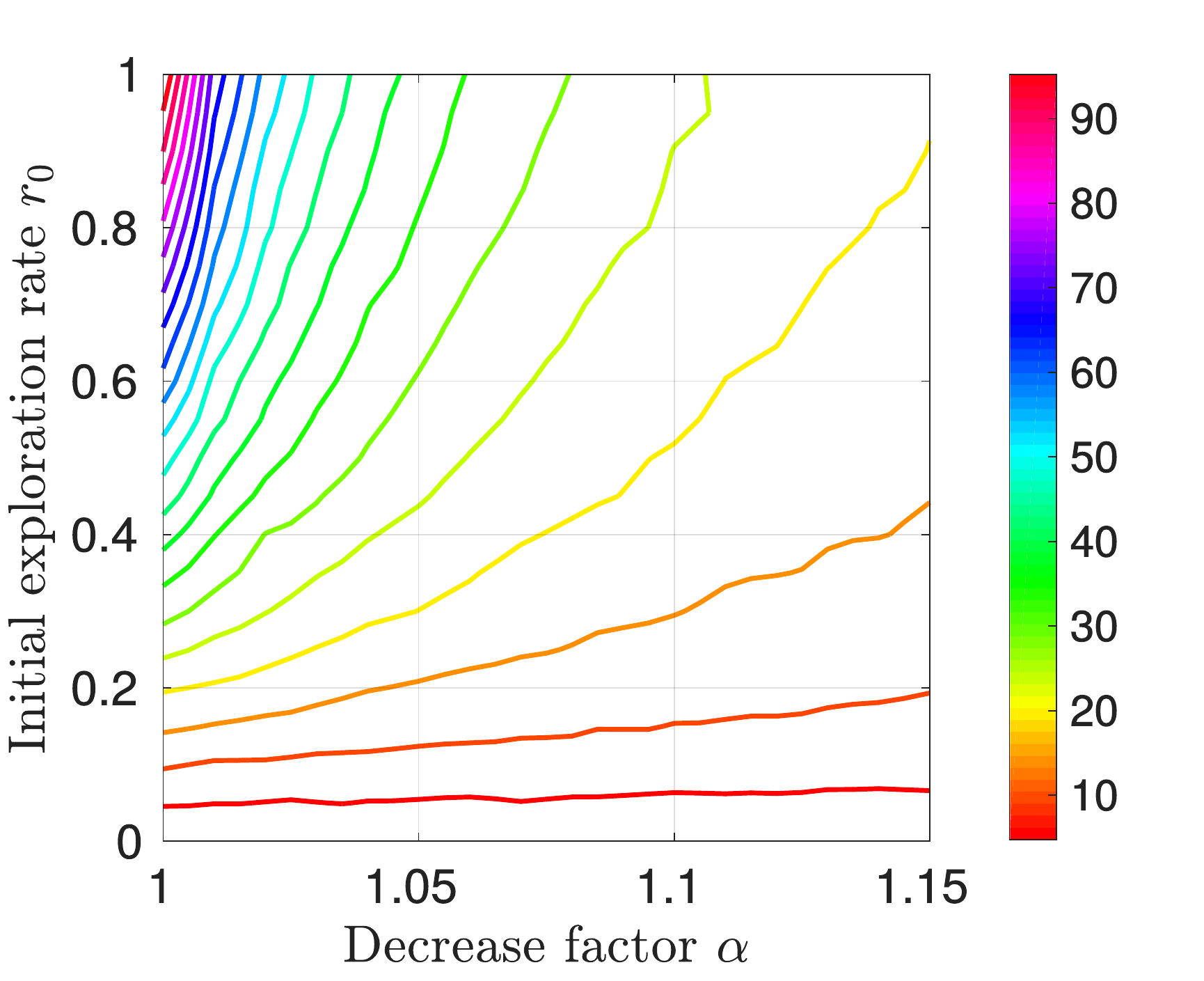}
\subcaption{Average number of \gls{mipt} samples among 100 added samples}\label{fig:MIVor_nb_MIPT}
\end{subfigure}%
\begin{subfigure}[t]{0.49\textwidth}
\includegraphics[width=\textwidth]{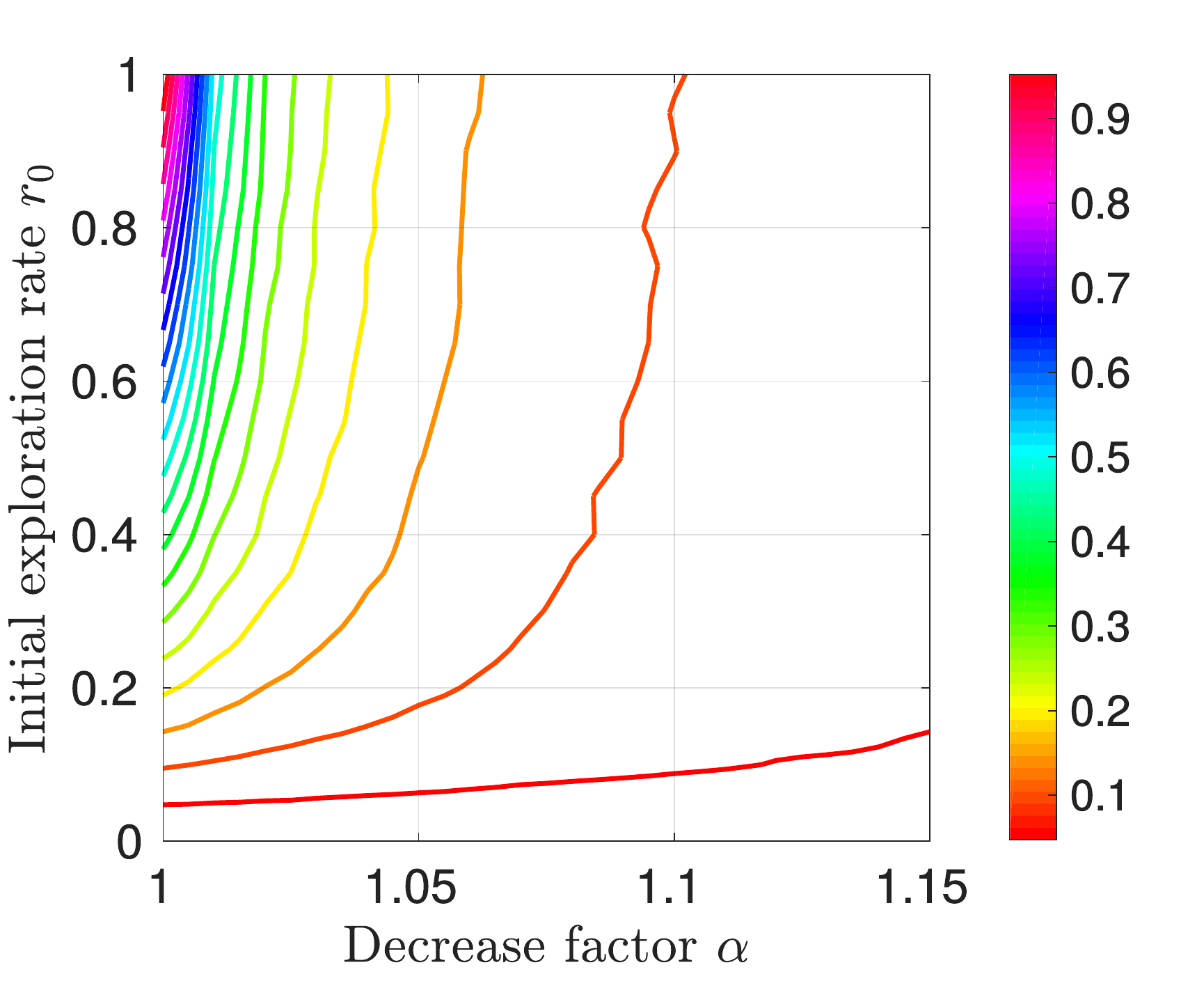} 
\subcaption{Average remaining exploration rate $r$ after adding 100 samples}\label{fig:MIVor_explor_rate}
\end{subfigure}

\begin{subfigure}[t]{0.5\textwidth}
\includegraphics[width=\textwidth]{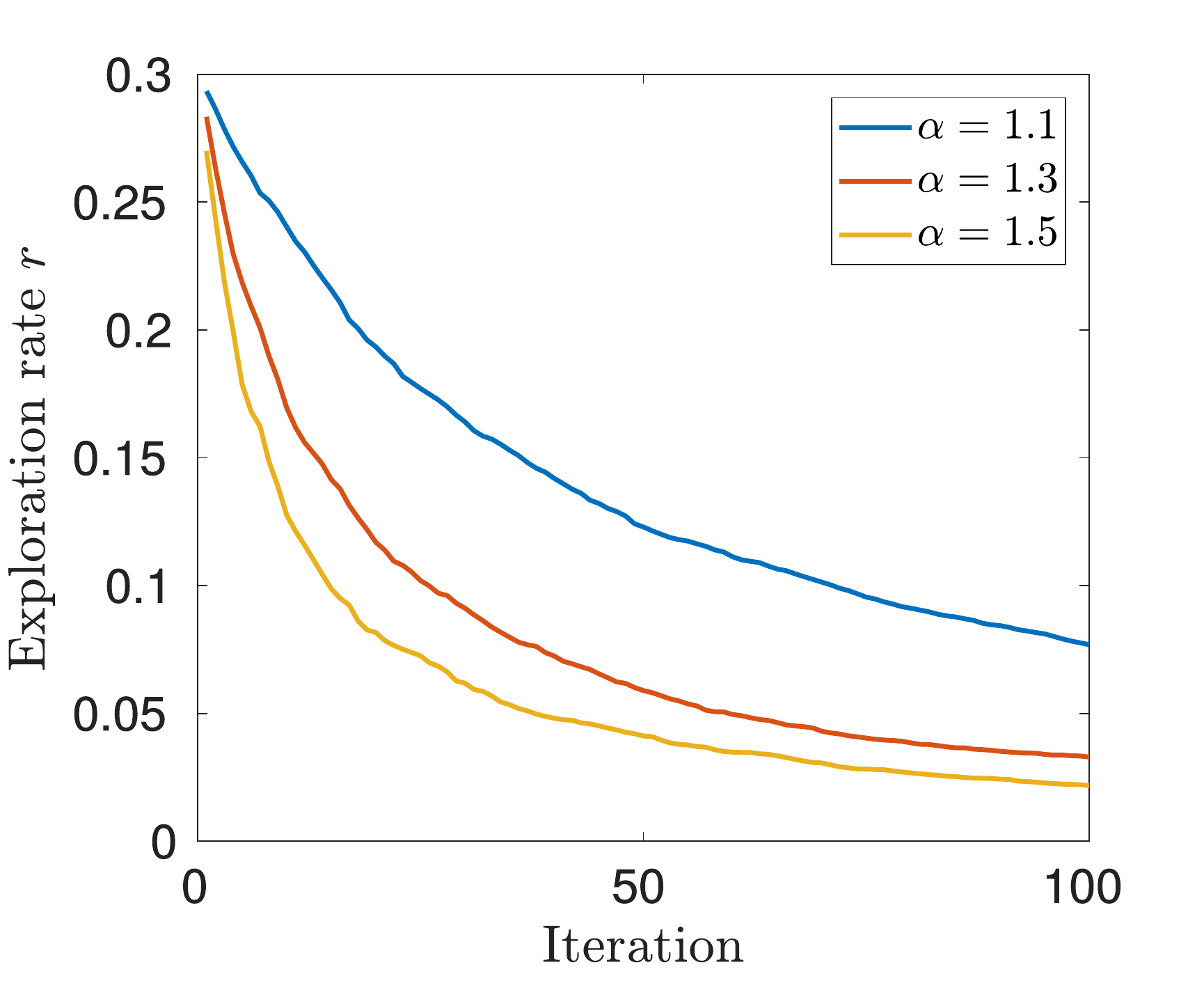} 
\subcaption{Decrease of exploration rate with iterations considering different decrease factor values}\label{fig:MIVor_rate_iteration}
\end{subfigure}%
\caption[\gls{mivor} process parameters after 100 samples.]{Investigation of the effect of \gls{mivor} process parameters while adding 100 samples (results averaged over 100 realizations).}\label{fig:MIVORExplorationRateAndAlpha}
\end{figure}

The effect of the initial exploration rate and the decreasing rate on the behaviour of \gls{mivor} is illustrated in Figure \ref{fig:MIVORExplorationRateAndAlpha}. All the represented results have been averaged over 100 realization paths. On Figure \ref{fig:MIVor_nb_MIPT} it can be observed how the number of samples added by exploration, i.e. using \gls{mipt}, among 100 added samples can be tuned depending on the initial exploration rate and the decrease rate.
It can be seen that the higher the decrease factor the lower the exploration character of the adaptive algorithm. Besides, as expected, a higher initial exploration rate yields a higher exploration character of the adaptive scheme. Figure \ref{fig:MIVor_explor_rate} shows the remaining exploration rate after adding 100 observation points displayed over $\alpha$ and $r_{0}$. The oscillating behavior which can be seen for some curves on both figures is mainly due to the low number of realization paths, i.e. 100 realization paths. As expected, it may be observed that the remaining exploration rate is smaller with higher decrease factor and lower initial exploration rate. Considering a fixed value of the initial exploration rate, after 100 supplementary samples, the remaining exploration rate may vary a lot, which features largely different algorithm behavior. On Figure \ref{fig:MIVor_rate_iteration} the decrease of the exploration rate from that initial value can be observed. For each \gls{mipt} iteration, the exploration rate decreases and its domain ranges lies between $r_0$ and $0$. The evolution clearly depends on the decrease factor value chosen by the user. Therefore depending on the expected output the method offers substantial variability.
However for an unknown classification output a decrease factor of $1.1$ and an initial exploration rate of $0.4$ have been identified as the pair of numerical parameters generally yielding the most proficient adaptive scheme.

\subsection{Overview of \gls{mivor} adaptive scheme}

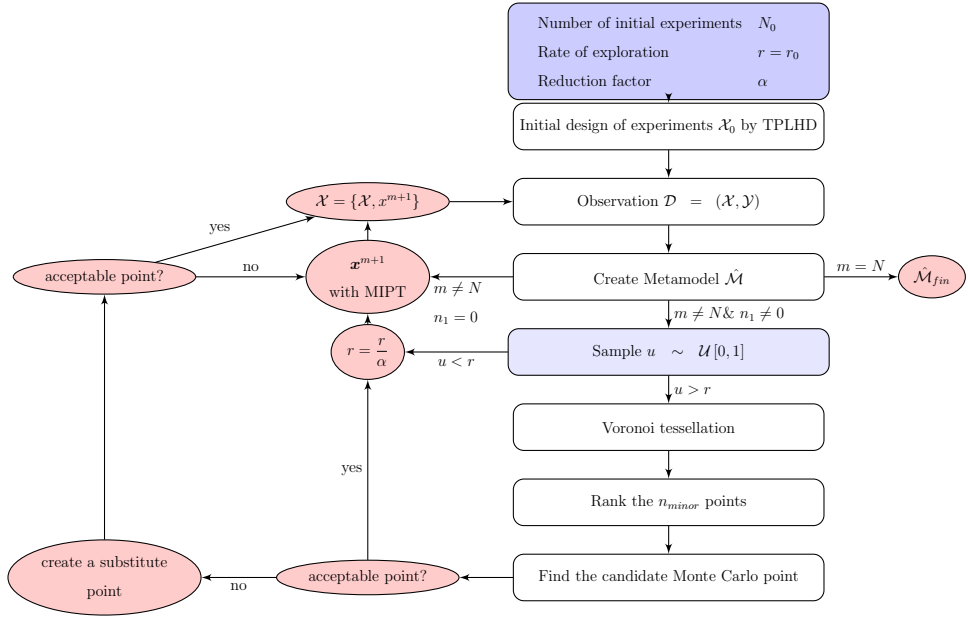
\begin{figure}[ht!]
\tikzstyle{decision} = [rectangle, draw,  
    text width=20em, text centered, rounded corners, node distance=2cm, inner sep=0pt,minimum height=3em]
\tikzstyle{block} = [rectangle, draw, fill=blue!20, 
    text width=20em, text centered, rounded corners, minimum height=4em,node distance=4cm,]
\tikzstyle{blockA} = [rectangle, draw, fill=blue!10, 
    text width=20em, text centered, rounded corners, minimum height=3em]
\tikzstyle{line} = [draw, -latex']
\tikzstyle{cloud} = [draw, ellipse,fill=red!20, node distance=8.0cm,
    minimum height=2em]
 \centering 
\begin{tikzpicture}[node distance = 1.5cm, auto,scale=0.5, transform shape]
    \node [block] (init) {
\begin{tabular}{ll}
Number of initial experiments & $N_0$ \\
Rate of exploration & $r=r_0$ \\
Reduction factor & $\alpha$ \\
\end{tabular}
};
\node [decision, below of=init,node distance=2.0cm] (init_DOE) {Initial design of experiments $\mathcal{X}_0$ by \gls{tplhd}};

\node [decision, below of=init_DOE,node distance=2.0cm] (start) {Observation $\mathcal{D} = \left( \mathcal{X}, \mathcal{Y} \right)$};
    
\node [decision, below of=start,node distance=2.0cm] (metamodel) {Create Metamodel $\hat{\mathcal{M}}$};

\node[cloud, right of=metamodel,node distance=7.0cm] (fin_fin) {$\hat{\mathcal{M}}_{fin}$};

\node[cloud, left of=metamodel,node distance=8.0cm,align=center] (new_MIPT) {$\bm{x}^{m+1}$\\with MIPT};

\node[cloud, left of=new_MIPT,node distance=7.0cm](subs_acceptable) {acceptable point?};

\node[cloud, left of=start] (rien){$\mathcal{X} = \left\lbrace \mathcal{X},x^{m+1}\right\rbrace$};
        
\node [blockA, below of=metamodel,node distance=2.0cm] (Sample_u) {Sample $u \sim \mathcal{U}\left[0,1 \right]$};

\node[cloud, left of =Sample_u] (Update_r) {$r = \dfrac{r}{\alpha}$};
                
    \node [decision, below of=Sample_u,node distance=2.0cm] (Voronoi) {Voronoi tessellation};
    
        \node [decision, below of= Voronoi,node distance=2.0cm] (Ranking) {Rank the $n_{minor}$ points};
        
        \node [decision, below of= Ranking,node distance=2.0cm] (Closest) {Find the candidate Monte Carlo point};
        
        \node[cloud, left of=  Closest](acceptable){acceptable point?};
        
         \node[cloud, left of=  acceptable, align=center,node distance=7.0cm](substitute){create a substitute\\point};
        
     \path [line] (init) -- (init_DOE);
    \path [line] (init_DOE) -- (start);
    \path [line] (start) -- (metamodel);
     \path [line] (metamodel) -- node{$m \neq N$ \newline \& $n_{1} \neq 0$} (Sample_u);
     \path [line] (metamodel) -- node {\parbox{2.0cm}{$m \neq N$ \\ $n_{1} = 0$}} (new_MIPT);
     \path [line] (Sample_u) -- node {$u>r$}(Voronoi);
     \path [line] (metamodel) -- node{$ m = N$} (fin_fin);
     \path [line] (Sample_u) -- node{$u < r$}(Update_r);
\path [line] (Update_r) -- (new_MIPT);
\path [line] (new_MIPT) -- (rien);
\path [line](rien) -- (start);
\path [line](Voronoi) -- (Ranking);
\path [line] (Ranking) -- (Closest) ;
\path [line] (Closest) -- (acceptable) ;

\path[line](acceptable) -- node{yes}(Update_r);
\path[line](acceptable) -- node{no}(substitute);
\path[line](substitute) -- (subs_acceptable);
\path[line](subs_acceptable) -- node{no} (new_MIPT);
\path[line](subs_acceptable) -- node{yes} (rien);

\end{tikzpicture}
\caption[Workflow of finding a new sample point with \gls{mivor}.]{Workflow for finding a new sample point with \gls{mivor}.}\label{fig::flowchart_MIVOrSample}
\end{figure}

To sum up, the whole \gls{mivor} procedure is defined by the following steps:
\begin{enumerate}
\item[] \textbf{Create initial metamodel}.\\ From the initial design of experiments $\mathcal{D} = \lbrace \mathcal{X}, \mathcal{Y} \rbrace$ generate an initial metamodel. Define a reduction factor $\alpha$ and an initial value of the rate of exploration $r = r_0$. 
\item[] \textbf{Check number of samples indicating minor $\mathcal{C}_1$ behavior.}\\ Among the input set, check the number of samples indicating $\mathcal{C}_1$ behavior $n_{1}$. As long as  $n_{1} = 0$, sample new points with \gls{mipt}. As soon as  $n_{1}>0$ go to the next step.
\item \textbf{Exploration or Exploitation.}\\
 Sample a uniform random variable $u \sim U[0,1]$, if $u <r$, find the next sample point by \gls{mipt} and reduce $r$ by factor $\alpha$ else go to step 2.  
\item \textbf{Rank the existing sample points of set $\mathcal{X}_{1}$.} \\ Evaluate the volumes of the Voronoi cells of samples in $\mathcal{X}_{1}$. Rank the $n_{1}$ points according to the formula of equation (\ref{eq::SamplePointRank}). Identify the highest scoring sample $\bm{x}^{max}$. Store the Monte Carlo points $\bm{x}^{max}$ in $P_{max}$.
\item \textbf{Sample in Voronoi cell with highest score.} \\Find the closest point in $\mathcal{X}_{2}$ in the neighborhood of $\bm{x}^{max}$. Set this point as the candidate point $\bm{x}^{cand}$.
Find the new sample point by following the workflow of Figure \ref{fig::flowchart_MIVOrSample}. Go back to step 2.
\end{enumerate}

For simplicity the adaptive scheme is stopped by reaching an a priori chosen number of experiments denoted by $N$. It can be noticed that the presented algorithm allows to maintain some exploration contribution in \gls{mivor} procedure even once a point in the minor domain has been found. This seems of particular importance for cases to provide a robust scheme for cases where the minor parametric domain is disconnected in several subdomains.

\section{Numerical tests}
\label{sec::Applications}

The strategy is tested for different parametric problems. First one-dimensional and two-dimensional analytical functions are explored for sake of simplicity and clarity. Then the method is investigated for two mechanical applications.

 For every case, the performance of the metamodel is evaluated with respect to a reference solution obtained with 5000 samples points per input dimension placed in the parametric domain using \gls{tplhd}.  The metamodel for classification is evaluated according to the following metric. The percentage of correctly predicted points in the $\mathcal{C}_{i}$ regime ($i=1,2$) denoted as $ a^{p}_{C_{i}} $ is defined as
 \begin{equation}
   a^{p}_{C_{i}}  = \dfrac{\hat{n}^{C_{i}}_{ref,C_{i}}}{\text{n}_{ref,C_{i}}},
 \end{equation}
where $\text{n}_{ref,C_{i}}$ is the number of reference points that have an output in the $\mathcal{C}_{i}$ regime, and $\hat{n}^{C_{i}}_{ref,C_{i}}$ denotes the number of reference points in class $\mathcal{C}_{i}$, which are actually predicted in class $\mathcal{C}_{i}$ by the metamodel.
Therefore $a^{p}_{C_{i}} $ is equal to zero when none of the reference output class is in accordance with the surrogate classification. When $a^{p}_{C_{i}} $ is $1.0$ all the points of that class in the reference solution are equivalent to the surrogate classification. It is highlighted that the percentage of correctly identified points is not globally analyzed as the challenge lying in identifying the minor class would be clouded by a global indicator. Final results are averaged values over 20 realizations of the adaptive sampling processes. Sample positions are possibly shown for one process realization, which is selected randomly.

The \gls{mivor} approach is compared to two adaptive sampling techniques classically used in context of Gaussian process regression, namely \gls{eigf} as presented by \cite{lam2008sequential} as well as the \gls{mepe} technique proposed by \cite{liu2017adaptive}. Comparisons with more adaptive methods on several reference problems have been proposed in \cite{Jan_Master_thesis}.

In order to avoid numerical issues in the scale of the input space the parametric space is normalized for each input dimension. The normalized input value is estimated from $x_{i}$ as
\begin{equation}
    \bar{x}_{i} = \frac{x_{i} - x_{i}^{l}}{x_{i}^{u} - x_{i}^{l}}
\end{equation}
with $x_{i}^{u}$ and $x_{i}^{l}$ the upper and lower limits of the parametric domain in dimension $i$ respectively. Thus the normalised quantities indicated by an upper bar belong to $\left[ 0,1 \right]$.

\subsection{Analytical reference functions}

First the innovative kriging algorithm is investigated for classification based on some one-dimensional and two-dimensional analytical functions. It allows to easily analyse the performances for different properties of the response surface.

\subsubsection{Higdon function} \label{subsub:Higdon}
\begin{figure}[ht!]
\centering
\begin{subfigure}[t]{0.5\textwidth}
\includegraphics[width = \textwidth]{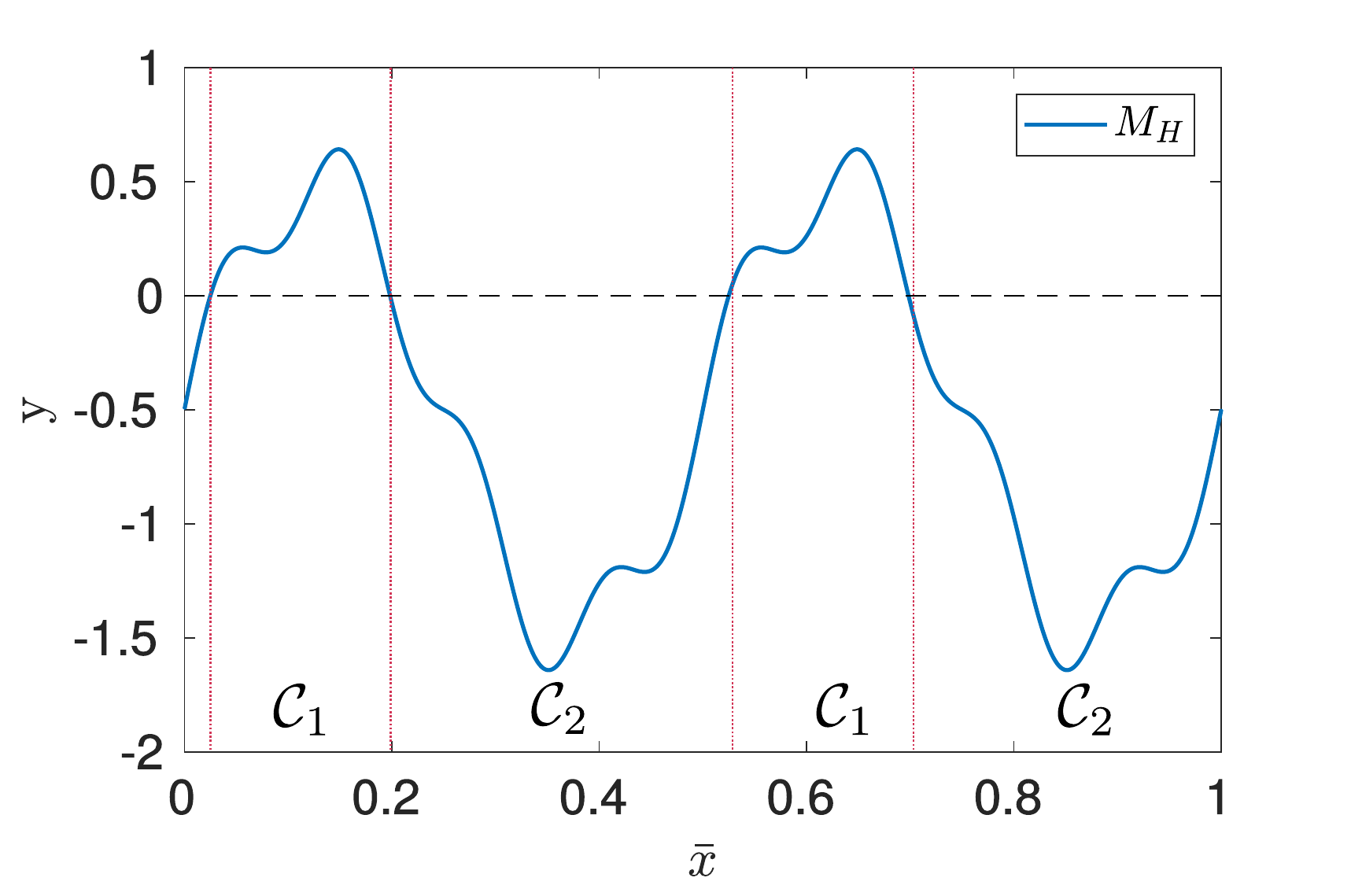}
\subcaption{$\mathcal{M}_{H}$}\label{fig::Higdon_surf}
\end{subfigure}%
\begin{subfigure}[t]{0.5\textwidth}
\includegraphics[width = \textwidth]{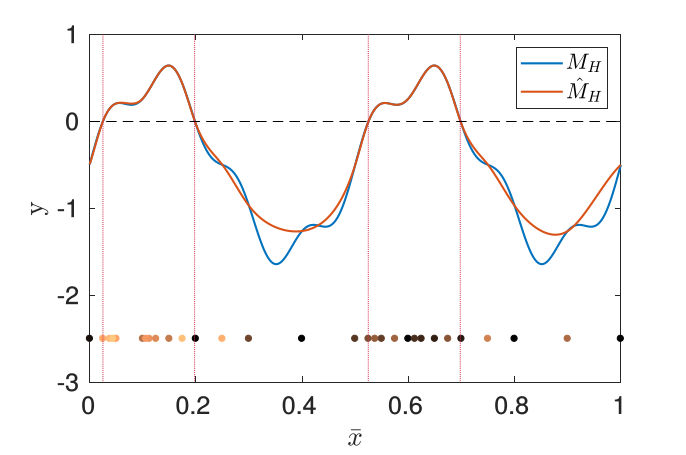}
\subcaption{$\hat{\mathcal{M}}_{H}$ and 35 samples with \gls{mivor}}\label{fig::Higdon_meta}
\end{subfigure}
\caption{Classification of Higdon function response. (a) response over normalized parametric space with limit chosen at $L=0$, (b) \gls{mivor} surrogate and sample positions.}\label{fig_class_Higdon}
\end{figure}

Consider a modified version of the smooth Higdon function \citep{higdon2002space} as a first example. The function reads
\begin{equation}
    \mathcal{M}_{H}(x) = \sin \left( \frac{2 \pi x}{10} \right) + 0.2 \sin \left( \frac{2 \pi x}{2.5} \right) -0.5,
    \end{equation}
where $x \in [-10,10]$, and so $\mathcal{M}_{H}(x) \in [-1.7,0.7]$. The limit output value for class $\mathcal{C}_{1}$ is set to $L=0$. The function response over the normalized domain is depicted in Figure \ref{fig::Higdon_surf}. Five initial samples are created with \gls{tplhd}. None of them correspond to an output value in class $ \mathcal{C}_{1}$. Then 30 samples are progressively added with the presented \gls{mivor} technique.

The resulting sample positions (colored dots) and the resulting surrogate model $\hat{\mathcal{M}}_{H}$ are plotted in Figure \ref{fig::Higdon_meta}. The brighter the point color of the samples the later they have been added to the dataset in the adaptive \gls{mivor} process. The sample positions reveal a concentration of samples near to the boundaries between classes $\mathcal{C}_{1}$ and $\mathcal{C}_{2}$, therefore the surrogate model yields proficient results in these areas whereas the surrogate does not necessarily yield accurate results in internal zones of the $\mathcal{C}_{2}$ domain.

\begin{figure}[htbp!]
\centering
\includegraphics[width=0.78\textwidth]{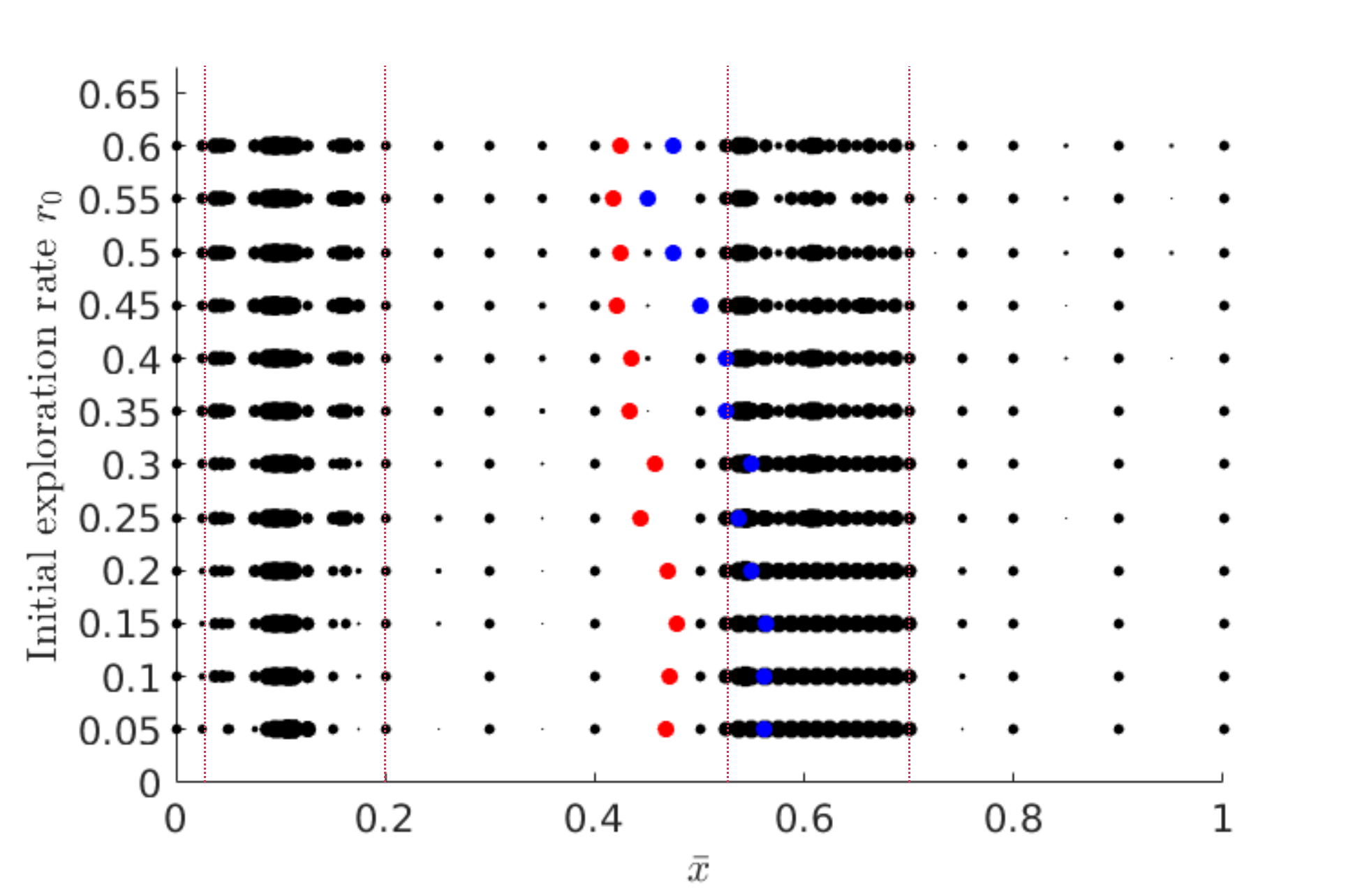}
\caption{Influence of the initial exploration rate on the sampling position for the Higdon example.}\label{fig::Higdon_influe}
\end{figure}

The influence of the initial exploration rate on the resulting sample position has been investigated. 20 realizations of the \gls{mivor} processes are processed until adding 30 samples for each of them. All of the (600) added samples are shown in Figure \ref{fig::Higdon_influe} over the normalized parametric domain for 12 different initial exploration rates between 0.05 and 0.60 and a fixed decrease factor equal to $1.1$. In order to enhance the visualization the size of each scatter point is increased proportionally to the number of points that are located in the neighborhood of each sample defined as the subdomain of maximal distance of $0.02$ around the sample.  
The mean and median sample positions are shown in red and blue color, respectively. It can be observed that small exploration rates yield an adaptive process which primarily samples in the $\mathcal{C}_{1}$ domain. When increasing the initial exploration rate, the \gls{mivor} algorithm creates more points with \gls{mipt} and therefore more points are added exploratively in the input domain.

\begin{figure}[htbp!]
\centering
\begin{subfigure}[t]{0.49\textwidth}
\includegraphics[width=\textwidth]{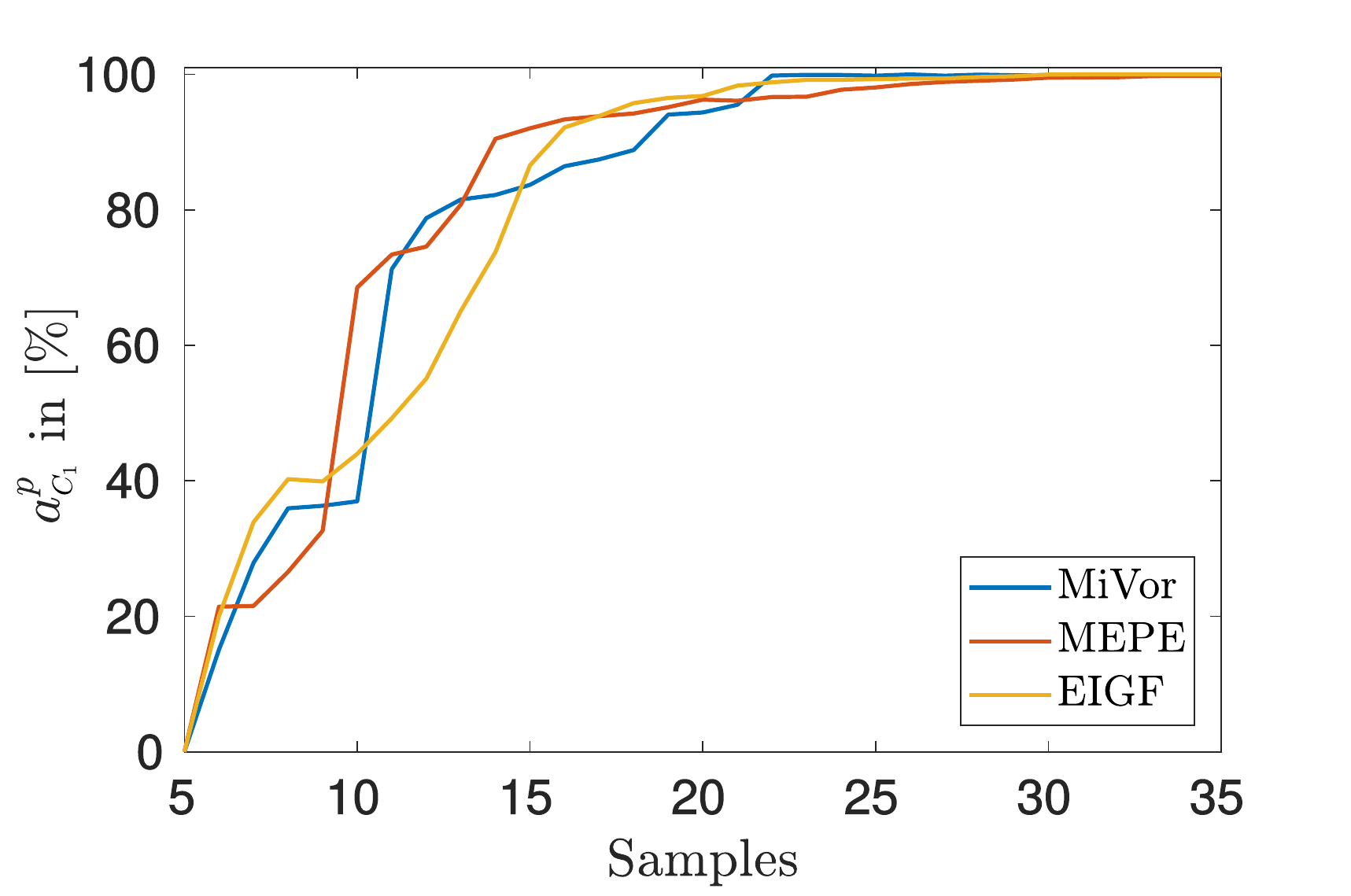}
\subcaption{$a^{p}_{C_{1}}$}\label{fig::Higdon_Above}
\end{subfigure}
\begin{subfigure}[t]{0.49\textwidth}
\includegraphics[width=\textwidth]{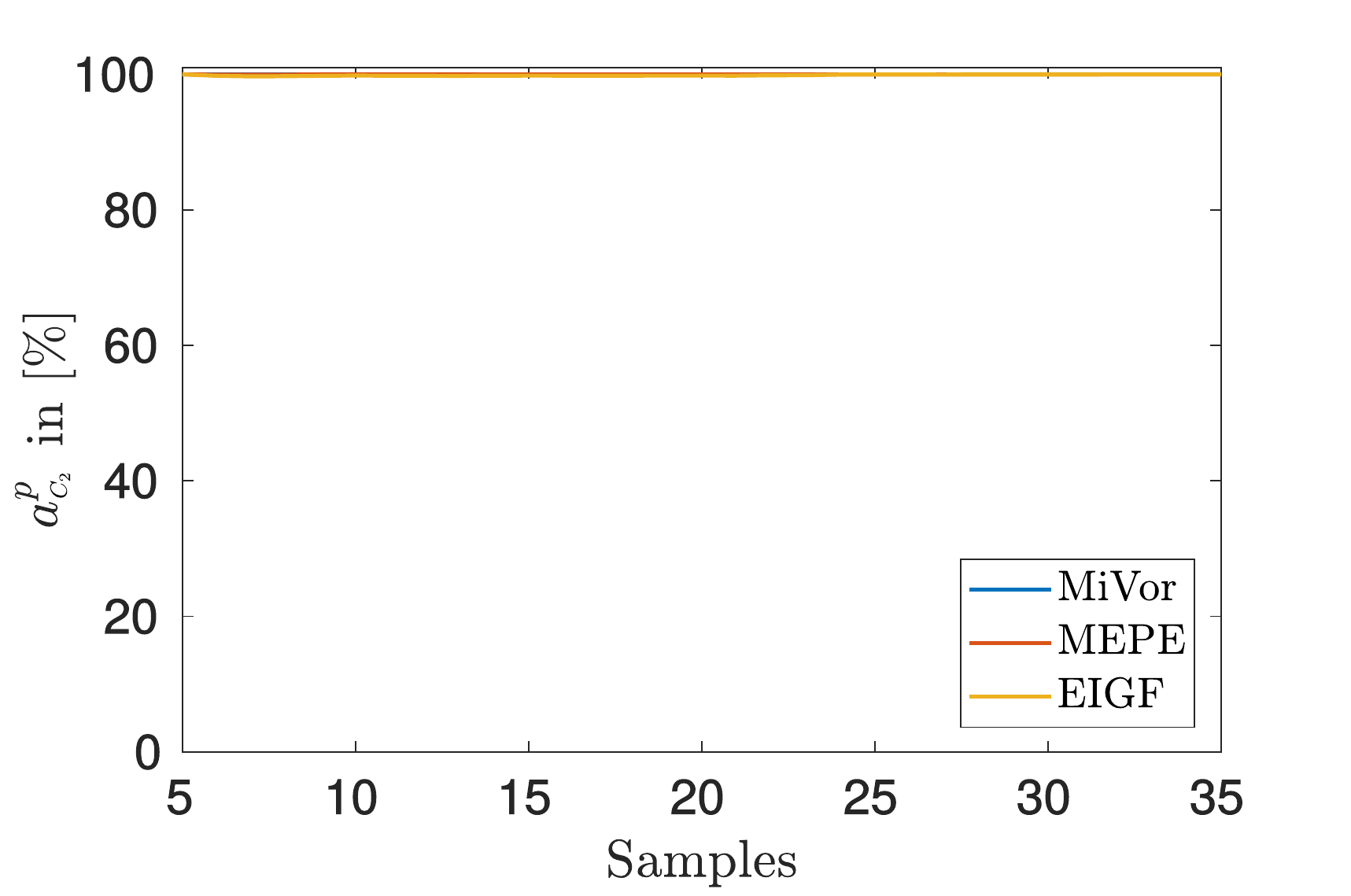}
\subcaption{$a^{p}_{C_{2}}$}\label{fig::Higdon_below}
\end{subfigure}%
\caption{Averaged error values for Higdon function with binary output.}\label{fig::Higdon_error}
\end{figure}

The error averaged over 20 realizations corresponding with the approximation $\hat{\mathcal{C}}_{H} $, as defined in Equation~\ref{eq::Ourapproach}, with respect to the reference solution is depicted in Figure \ref{fig::Higdon_error}. It can be noticed on Figure \ref{fig::Higdon_below} that the performances in terms of percentage of points correctly classified in the dominant class $\mathcal{C}_2$ do not provide valuable information, as even a rough surrogate model identifies all the points in the dominant class. For the problems of interest the challenge lies indeed in detecting the minor class subdomain. Concerning class $\mathcal{C}_1$, even if \gls{mivor} first reaches a value of $a^{p}_{C_{1}}$ upper than $99.99\%$, it can be seen in Figure \ref{fig::Higdon_Above} that all three methods yield around the same performances for this example. The output of the simple Higdon model could satisfactory be classified with any of the three surrogate models considering 25 sample points. 

The Higdon function exhibits a smooth evolution, whereas in applied mechanics and engineering highly fluctuating and non-smooth function responses may also be encountered.

\subsubsection{Modified Higdon function}
\begin{figure}[htbp!]
\centering
\begin{subfigure}[t]{0.5\textwidth}
\includegraphics[width=\textwidth]{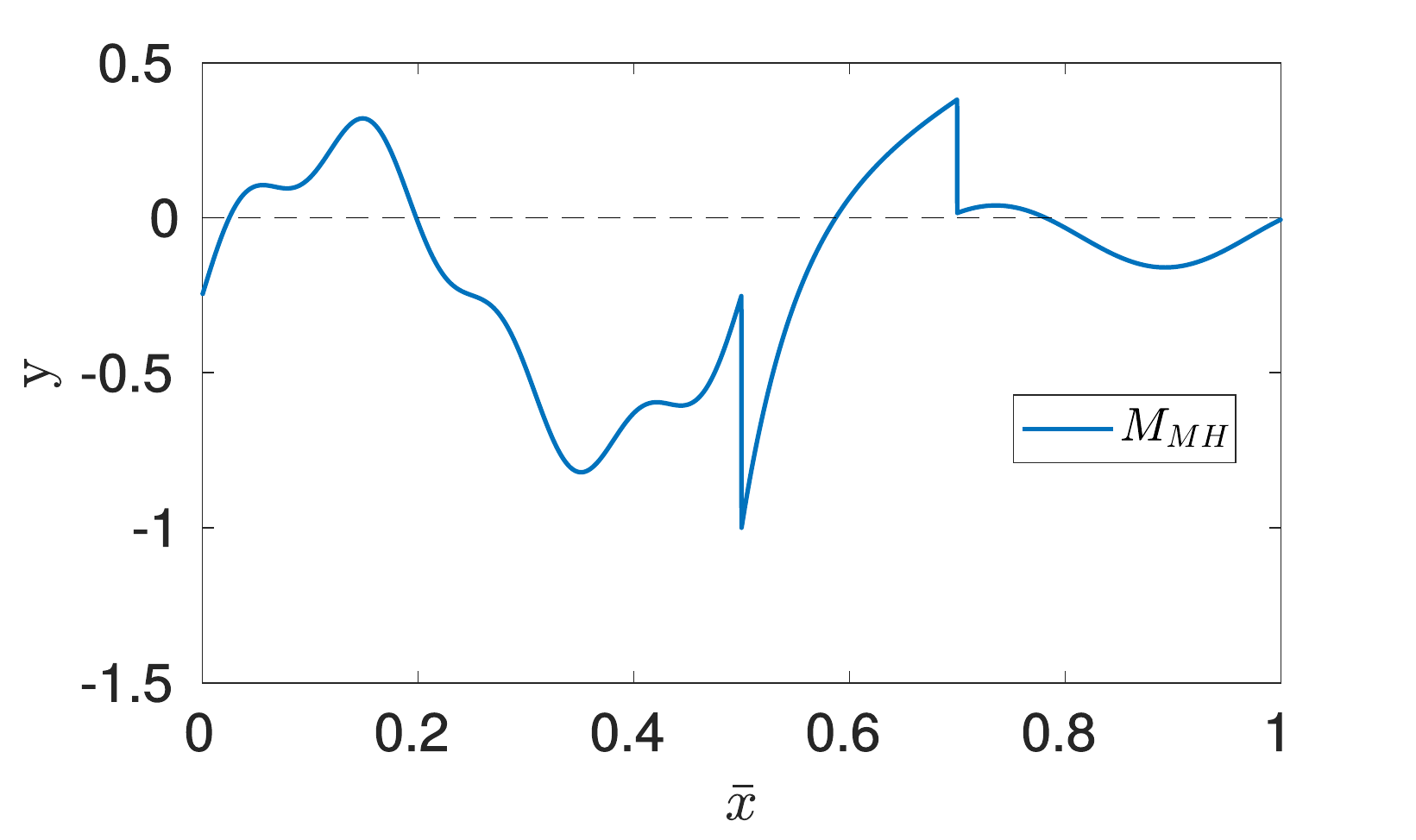}
\subcaption{$\mathcal{M}_{MH}$}\label{fig::Higdon_var_surf}
\end{subfigure}%
\begin{subfigure}[t]{0.5\textwidth}
\includegraphics[width=\textwidth]{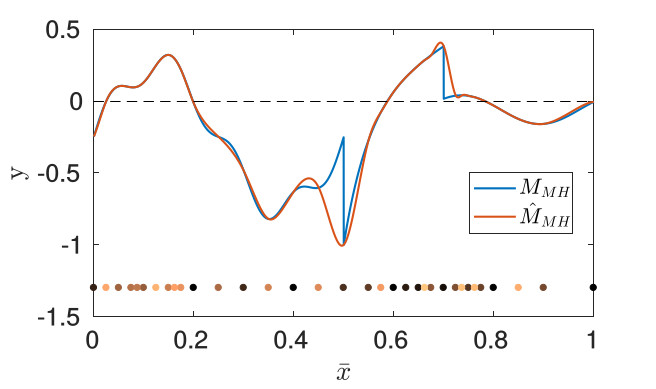}
\subcaption{$\hat{\mathcal{M}}_{MH}$ and samples}\label{fig::Higdon_var_meta}
\end{subfigure}
\caption{Classification of modified Higdon function response with discontinuities after 35 samples (a) response over normalized parametric space with limit chosen at $L=0$, (b) surrogate model based on 35 sample points.}\label{fig::class_mod_Higdon}
\end{figure}

Let consider, for further investigation, a modified version of the Higdon function given by
\begin{equation}
\mathcal{M}_{MH}(x) = \begin{cases}
    \sin \left( \frac{2 \pi x}{10} \right) + 0.2 \sin \left( \frac{2 \pi x}{2.5} \right) -0.5, & \text{if} \, x < 0, \\
    - \exp(-x) + 0.1 x, & \text{if} \, 0 \leq x < 4, \\
    - 0.1 \sin (x)-0.06 , & \text{if} \, x \geq 4, 
    \end{cases}
\end{equation}
where $x \in [-10,10]$, as illustrated over the normalized input domain in Figure \ref{fig::Higdon_var_surf}. The response surface is now discontinuous and non-periodic. The limit value between the two classes is kept to $L = 0$. 

The same five initial TPLHD sample points are considered, none of them lies in the class $\mathcal{C}_1$ domain, and the \gls{mivor} algorithm is employed until adding 30 samples to the initial dataset. A set of sample locations as well as a resulting metamodel is displayed in Figure \ref{fig::Higdon_var_meta}. It can similarly be observed that \gls{mivor} predominantly adds samples in the $\mathcal{C}_{1}$ domain, and particularly near to the boundary with the class $\mathcal{C}_{2}$ domain. It is able to create a metamodel which accurately classifies the response surface in the two class labels.

\begin{figure}[!htbp]
\centering
\includegraphics[width=0.7\textwidth]{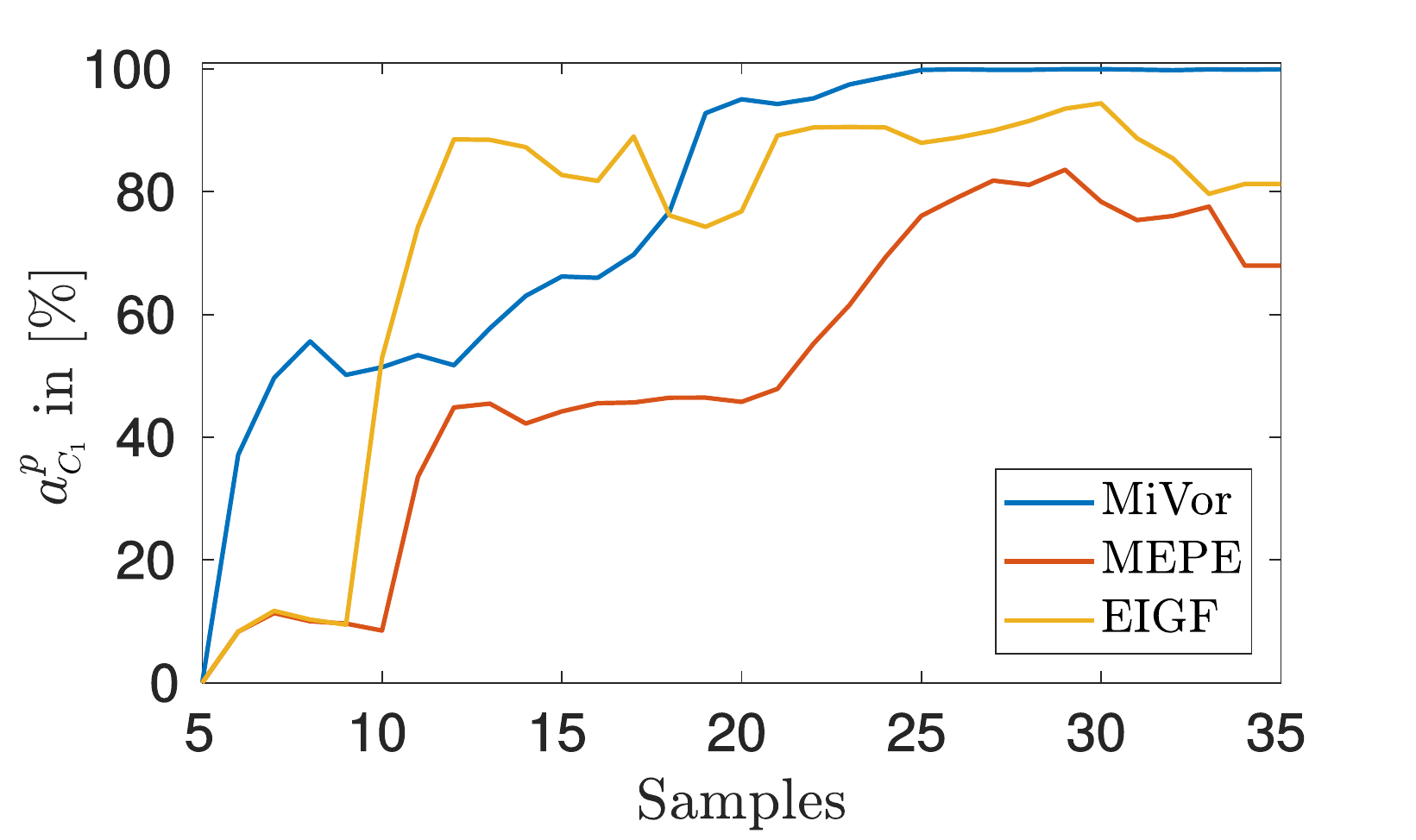}
\caption{Evolution of the averaged error value $a^{p}_{C_{1}}$ for modified Higdon function with number of samples.}\label{fig::Hig_variation_errorData}
\end{figure}

The evolution of the averages error measures during the adaptive process is shown in Figure \ref{fig::Hig_variation_errorData}. As previously, only class $\mathcal{C}_1$ is of interest. It can be seen that \gls{mivor} yields much better results in comparison to the other two methods. \gls{mivor} is able to reach an accuracy of $99.5 \%$ with a metamodel based on 25 observations, whereas both \gls{mepe} and \gls{eigf} do not seem to reach convergence even with 35 samples with performance results around $80 \, \%$ of points correctly identified in class $\mathcal{C}_1$.

\begin{figure}[ht!]
\centering
\begin{subfigure}[t]{0.5\textwidth}
\includegraphics[width=\textwidth]{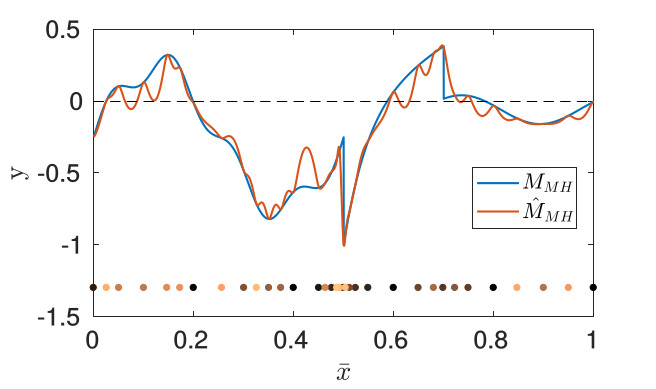}
\subcaption{\gls{mepe}}\label{fig::Hig_Variation_MEPE}
\end{subfigure}%
\begin{subfigure}[t]{0.5\textwidth}
\includegraphics[width=\textwidth]{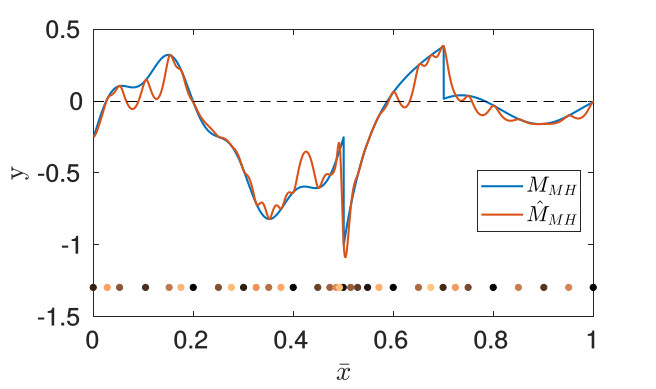}
\subcaption{\gls{eigf}}\label{fig::Hig_Variation_EIGF}
\end{subfigure}
\caption{Classification of modified Higdon function response with discontinuities after 45 samples (a) surrogate and sample positions for \gls{mepe}, (b) surrogate and sample positions for \gls{eigf}.}\label{fig::Higdon_Var_EIGFMEPE_meta}
\end{figure}

The difference in performances is due to the problems faced by the traditional methods to deal with discontinuities as shown in Figure \ref{fig::Higdon_Var_EIGFMEPE_meta} through two example sample sets and the resulting metamodels for \gls{mepe} and \gls{eigf}. It can be noticed that both sampling techniques run into numerical problems due to point clustering and so numerical problems in the inversion of the auto-correlation matrix. Besides, in both cases a large number of points are sampled in the area around $\bar{x}= 0.5$, which do not improve the accuracy of the final metamodel in terms of classification performances.

\subsubsection{Modified Drop-Wave function}

\begin{figure}[htbp!]
\centering
\begin{subfigure}[t]{0.5\textwidth}
\includegraphics[width=\textwidth]{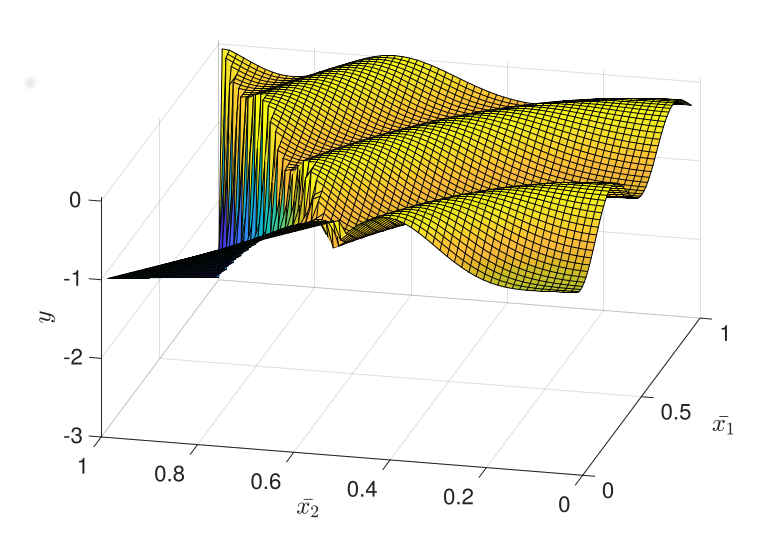}
\subcaption{$\mathcal{M}_{MDV}$ response surface}\label{fig::disc_surf}
\end{subfigure}%
\begin{subfigure}[t]{0.5\textwidth}
\includegraphics[width=\textwidth]{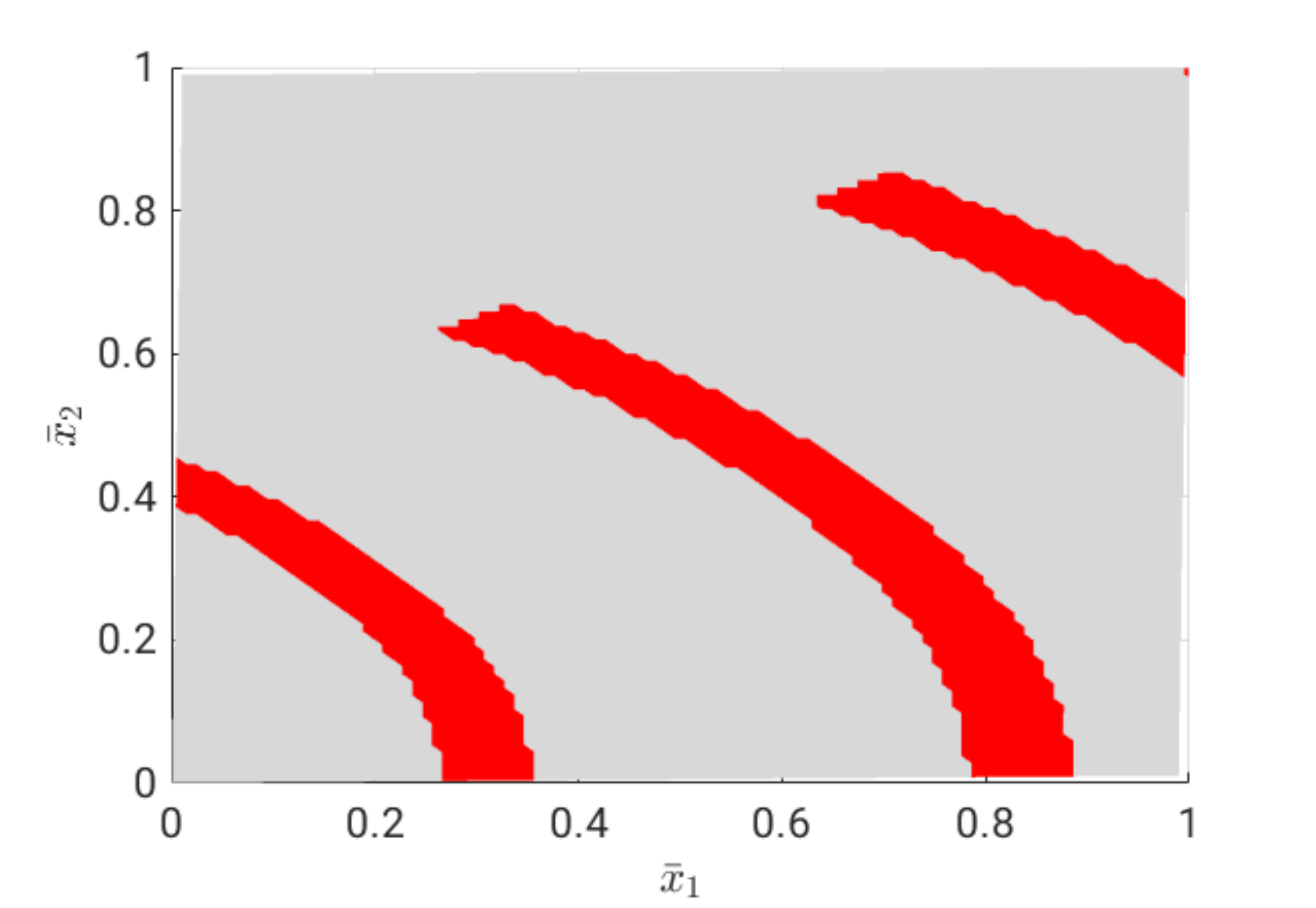}
\subcaption{Output classification of $\mathcal{M}_{MDV}$}\label{fig::disc_class}
\end{subfigure} \\
\begin{subfigure}[t]{0.5\textwidth}
\includegraphics[width=\textwidth]{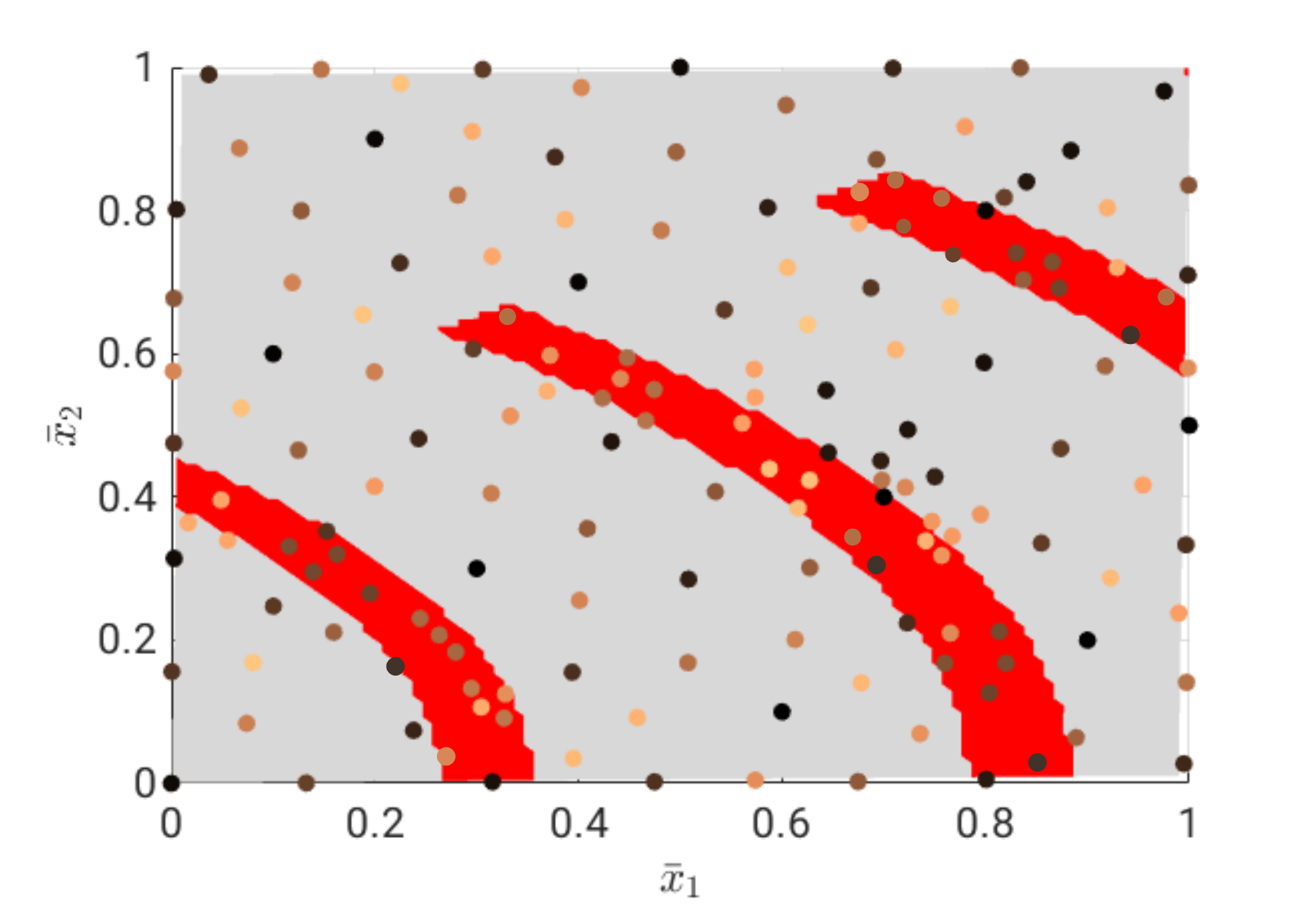}
\subcaption{Example set of 150 \gls{mivor} samples}\label{fig::disc_samples}
\end{subfigure}%
\begin{subfigure}[t]{0.5\textwidth}
\includegraphics[width=\textwidth]{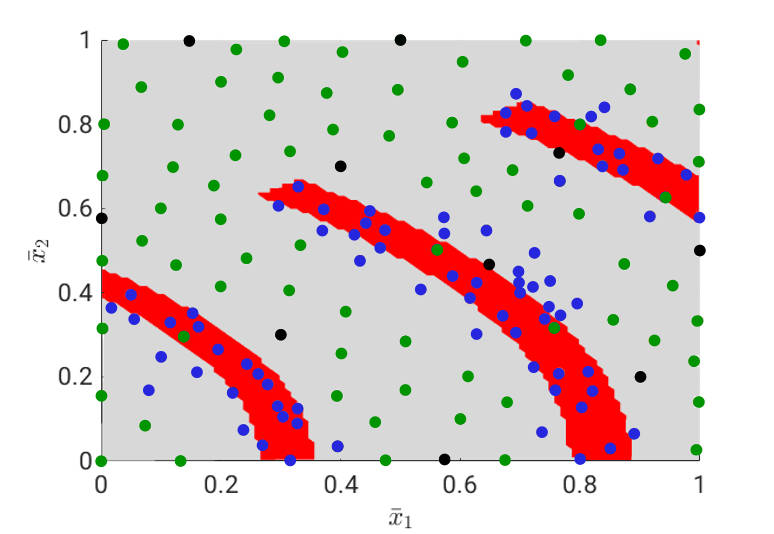}
\subcaption{Samples based on initial set in black, exploration steps in green and exploitation steps in blue}\label{fig::disc_Expo}
\end{subfigure}
\begin{subfigure}[t]{0.5\textwidth}
\includegraphics[width=\textwidth]{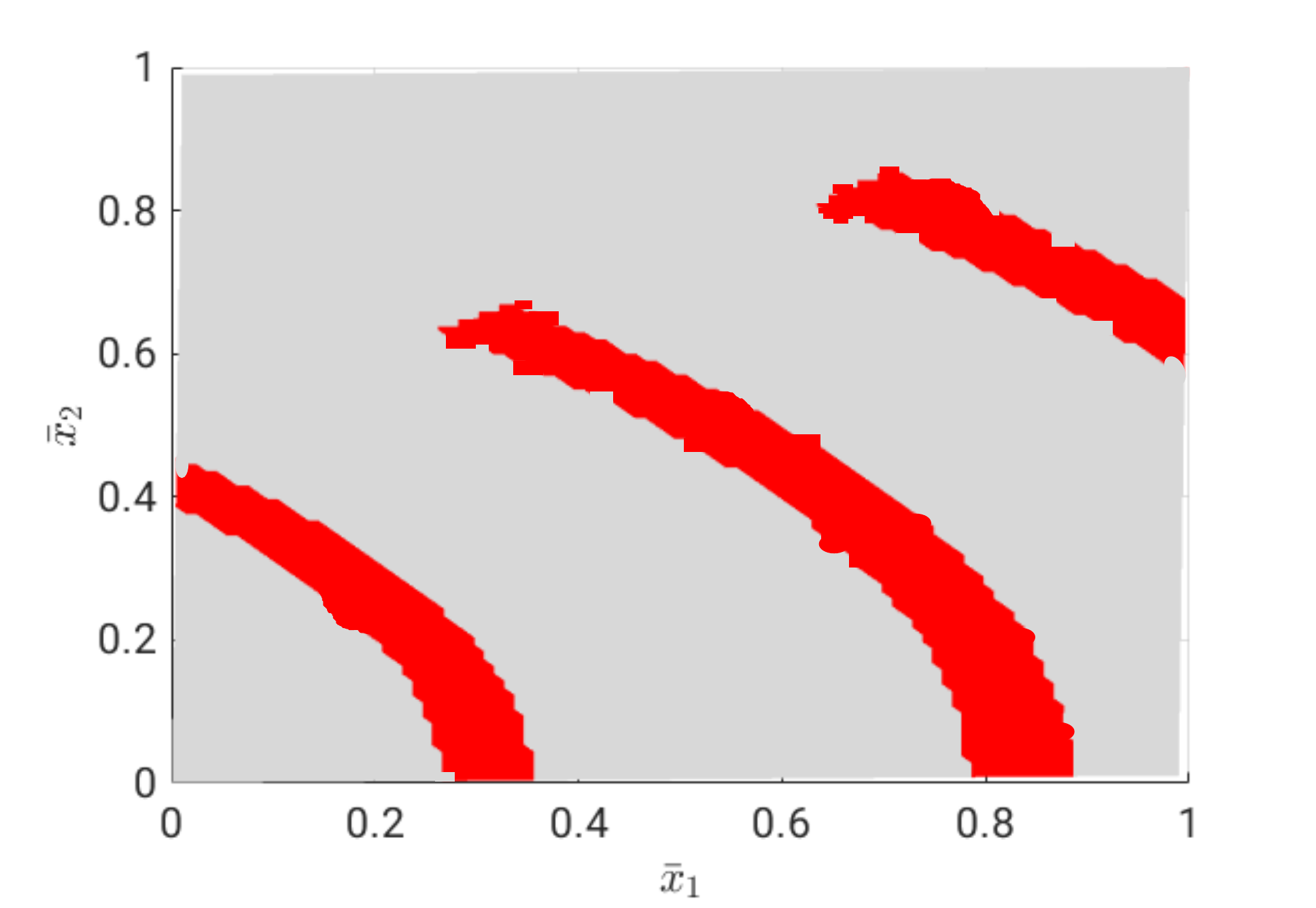}
\subcaption{Classification metamodel evaluated at reference points}\label{fig::disc_meta}
\end{subfigure}
\caption{\gls{mivor} for modified Drop-Wave function.}\label{fig::class_mivor_drop_wave}
\end{figure}

Consider a discontinuous modification of the two-dimensional Drop-Wave function defined as
\begin{equation}
\begin{aligned}
\mathcal{M}_{MDV}(\bm{x}) =
\begin{cases}
1- \abs{x_{1} x_{2}}, & \text{if} \,  x_{2} > x_{1}, \\
-\frac{1+ \cos \left( 12 \sqrt{x_{1}^{2} + x_{2}^{2}} \right)}{0.5 \left(1^{2} + x_{2}^{2} \right) +2} + 0.05, & \text{else},
\end{cases}
\end{aligned}
\end{equation}
on the parametric space $(x_1,x_2) \in [1,0] \times [2,2]$. Let the limit $L$ be defined by $L=0.0$. The response of the function over the normalized space is plotted in Figure \ref{fig::disc_surf}. The corresponding class labels are shown in Figure \ref{fig::disc_class}, where red subdomains correspond to $\mathcal{C}_{1}$ class and gray subdomains to $\mathcal{C}_{2}$ class. An example of set of samples randomly picked among the 20 realizations is depicted in Figure \ref{fig::disc_samples} over the target classification output. Initially 10 samples are created with \gls{tplhd}, none of which yield an output value corresponding to $\mathcal{C}_{1}$. Next, $140$ \gls{mivor} samples are added to the dataset. In Figure \ref{fig::disc_Expo} it can be seen that the \gls{mipt} steps spread the points all over the parametric domain whereas the exploitation component adds points on and around the $\mathcal{C}_{1}$ domain. The final metamodel is evaluated in terms of classification outcome at 10000 reference points, as shown in Figure \ref{fig::disc_meta}. The classification metamodel appears to be rather in good accordance with the reference classification solution shown in Figure \ref{fig::disc_class}. The boundary localization is not exactly accurate, but using only 140 observation points, the metamodel is able to identify the three minor disconnected subdomains and to localize them in a rough manner, which can be acceptable for first engineering design and/or optimization processes.

\begin{figure}[htbp!]
\centering
\includegraphics[width=0.6\textwidth]{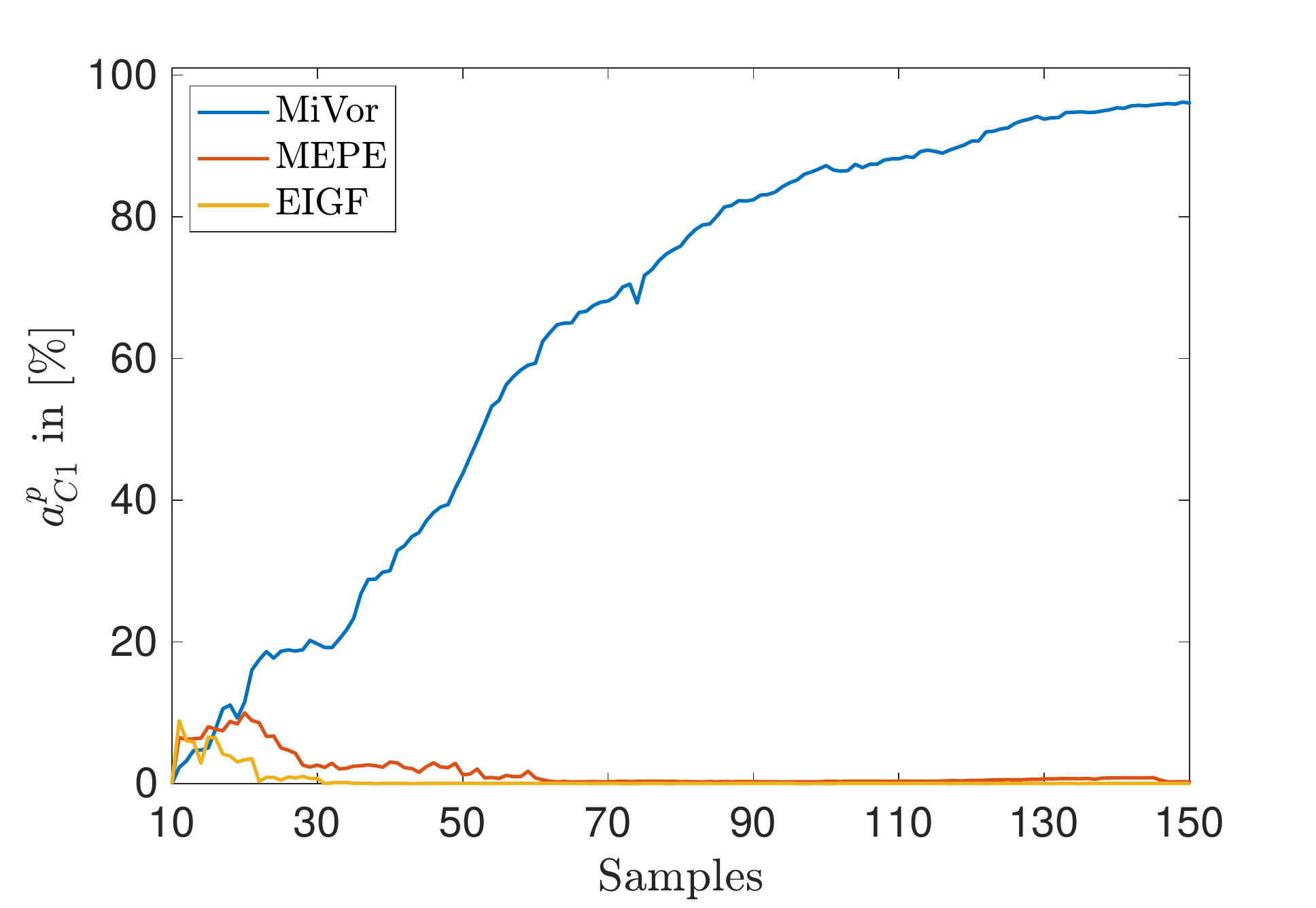}
\caption{Averaged error values concerning the class $\mathcal{C}_{1}$ for modified Drop-Wave function with binary output for different adaptive sampling techniques.}\label{fig::Disc_data}
\end{figure}

The error in terms of $a^{p}_{C_{2}}$ remains almost constant with values very close to $1.0$ while adding sample points. The evolution of $a^{p}_{C_{1}}$ along the adaptive process for \gls{mivor}, \gls{mepe} and \gls{eigf} is plotted in Figure \ref{fig::Disc_data}, in terms of averaged values over 20 realization paths. From the initial metamodel based on 10 observation points for which $a^{p}_{C_{1}} = 0.0$, the performance of the adaptive metamodel built using \gls{mivor} approach improves quite regularly until reaching a value of around $0.97$ using 150 observation points, which means considering 10000 reference points comprising 1475 points in class $\mathcal{C}_1$. Among them 97$\, \%$ i.e. 1436 points are correctly classified and only 39 points are wrongly estimated as belonging to class $\mathcal{C}_2$ by the surrogate model. In view of the computational effort involved in that surrogate model, these performances appear totally satisfactory. On the contrary, both \gls{mepe} and \gls{eigf} are not able to improve the classification performances by adding some new experiments, $a^{p}_{C_{1}}$ remains at values close to $0.0$ for both cases. 

It can be concluded that despite the fact that usual regression adaptive methods are able to capture classification feature for smooth functions as shown in Example \ref{subsub:Higdon}, they fail for classification problems corresponding with complex response surface, as they are not designed for that goal.

\begin{figure}[htbp!]
\centering
\begin{subfigure}[t]{0.5\textwidth}
\includegraphics[width=\textwidth]{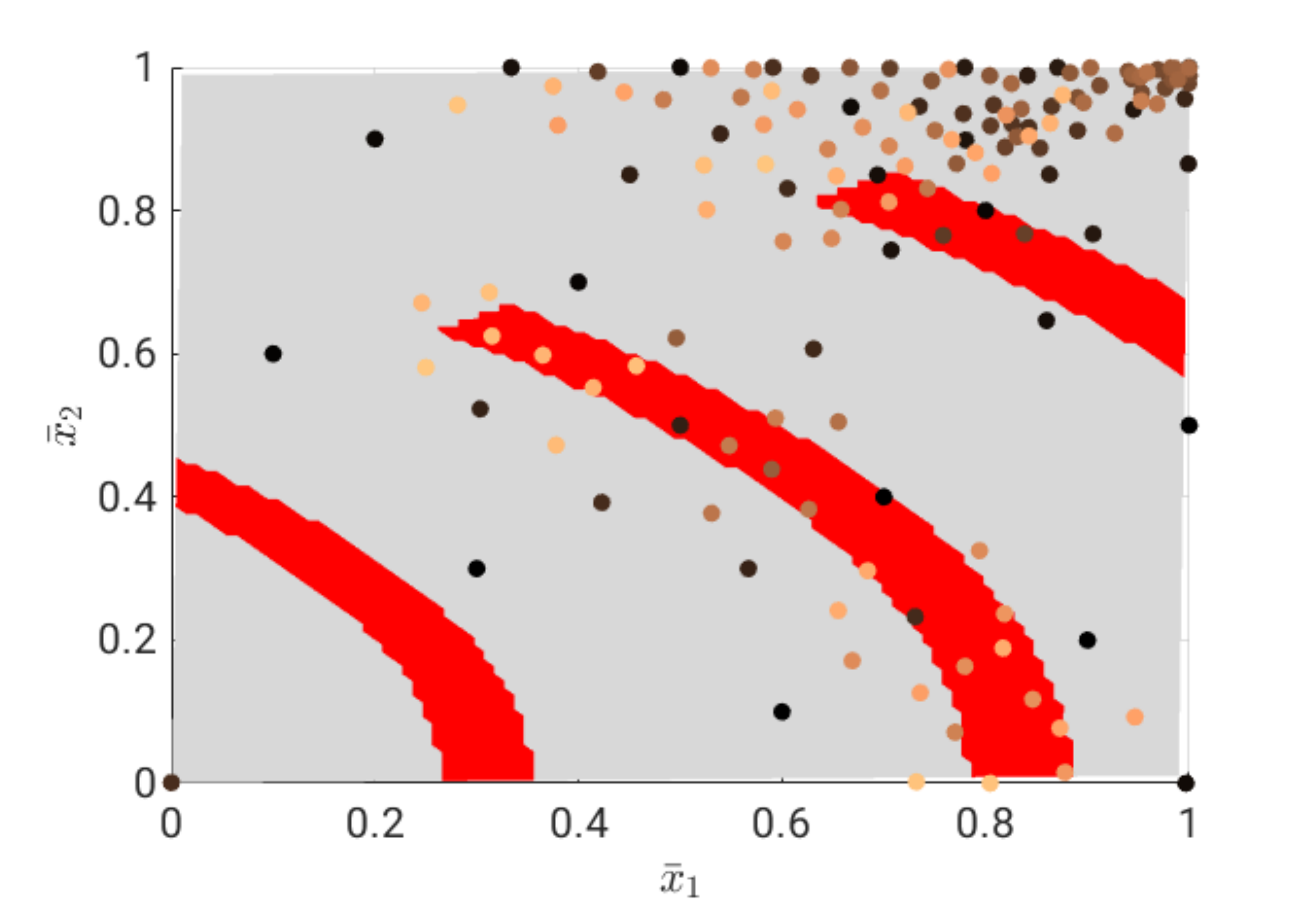}
\subcaption{Example set of 150 \gls{mepe} Samples}\label{fig::Drop_wave_sample}
\end{subfigure}%
\begin{subfigure}[t]{0.5\textwidth}
\includegraphics[width=\textwidth]{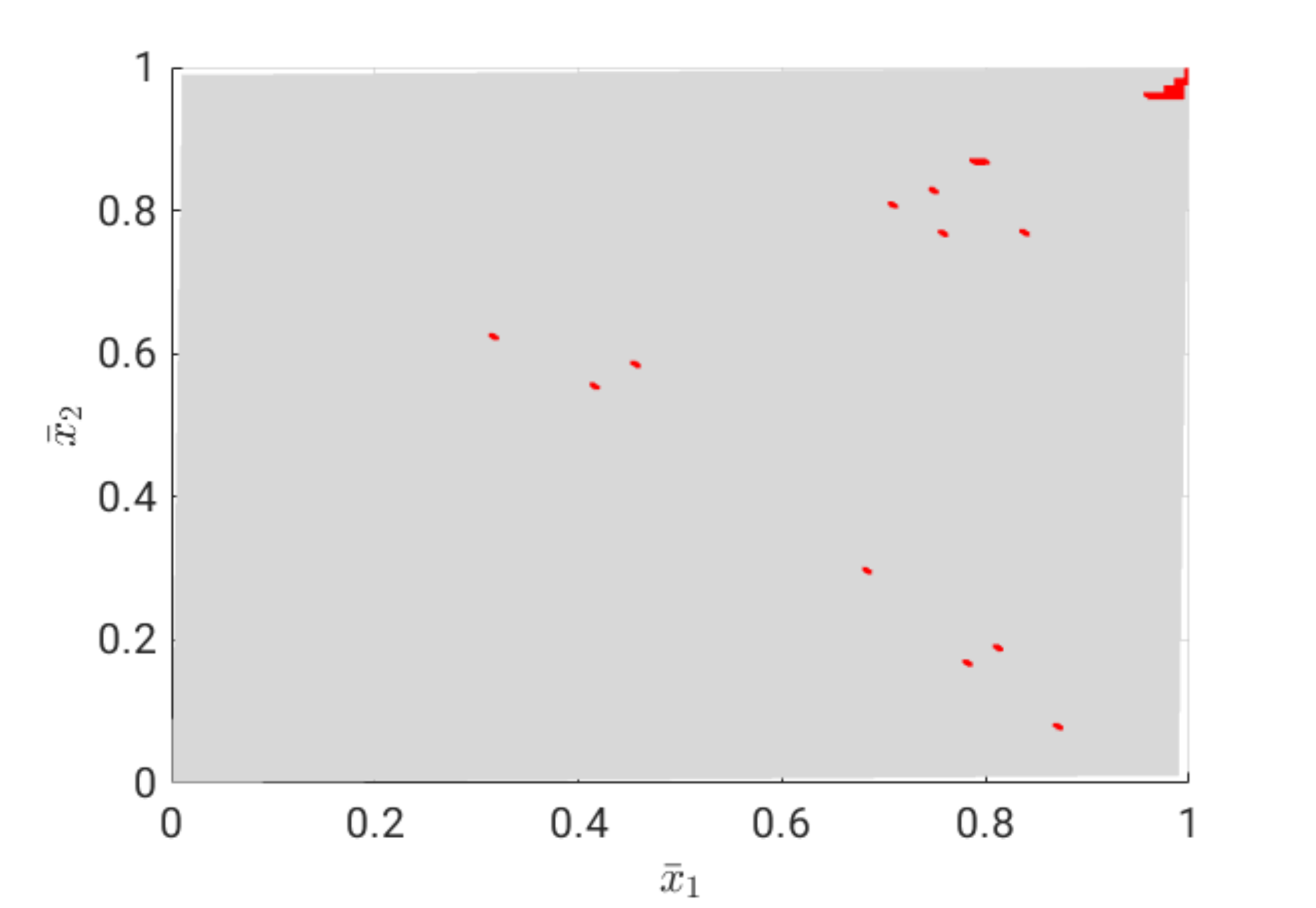}
\subcaption{MEPE classification metamodel}\label{fig::Drop_wave_MEPE_class}
\end{subfigure}
\caption{\gls{mepe} for modified Drop-Wave function after 150 samples.}\label{fig::Drop_wave_MEPE}
\end{figure}

The detailed behavior of \gls{mepe} can be analyzed in Figure \ref{fig::Drop_wave_MEPE}. It can be seen in Figure \ref{fig::Drop_wave_sample} that \gls{mepe} samples around the discontinuous drop of the response surface, which results in a poor predication performance with regard to the classification as shown in Figure \ref{fig::Drop_wave_MEPE_class}. A similar phenomenon can be observed with \gls{eigf}.

Finally the results obtained by the kriging metamodel are compared with Gaussian process classification based on either Laplace approximation or expectation propagation. Relative error measures provided in Table \ref{table::disc_compare} have been estimated in comparison with a reference solution based on 10000 \gls{tplhd} points designed in one step. Here the goal is not to investigate the sampling scheme, but the surrogate approaches. For different sample sizes, identical sample sets obtained by \gls{tplhd} are considered for building the three alternative metamodels. It can be seen that for lower sample sizes kriging classification outperforms the two other methods. However when the sample size increases, i.e. with data set larger than 200 observation, the other two methods perform as good or slightly better than kriging approach. However it can be seen that all these methods based on \gls{tplhd} sampling need at least 500 samples to reach results as accurate as \gls{mivor} with around 140 samples. This shows the efficiency of the presented adaptive sampling technique for highly fluctuating classification problems.
 \begin{table}[htbp!]
\begin{center}
 \resizebox{\textwidth}{!}{  
\begin{tabular}{c| c c |c c|c c|c c}
\cline{6-9}
\multicolumn{3}{c}{}  & \multicolumn{2}{c}{} & \multicolumn{4}{|c|}{Gaussian process classification } \\ \hline
\multicolumn{3}{|c|}{Samples}  & \multicolumn{2}{|c|}{Kriging} & \multicolumn{2}{|c|}{Laplace} &\multicolumn{2}{|c|}{Expectation} \\
\multicolumn{3}{|c|}{classification}  & \multicolumn{2}{|c|}{} & \multicolumn{2}{|c|}{approximation} &\multicolumn{2}{|c|}{propagation} \\
\hline
Total & $n_{C1}$ & $n_{C2}$ & $a^{p}_{C_{1}}$ [$\%$] & $a^{p}_{C_{2}}$ [$\%$] & $a^{p}_{C_{1}}$ [$\%$] & $a^{p}_{C_{2}}$ [$\%$] & $a^{p}_{C_{1}}$ [$\%$] &$a^{p}_{C_{2}}$ [$\%$] \\
\hline
\hline
50 &9 & 41 & 15.18 & 93.31 & 0.00 & 100.00 & 0.00 & 100.00  \\
100 &16& 84 &  77.15 & 98.19 & 0.00 & 100.00 & 0.00 & 100.00  \\
150 &22& 128& 73.96 & 99.22 & 41.68 & 99.66& 63.61 & 98.82 \\
200 &31 & 169 &76.47 & 99.63 & 68.66 & 99.43 & 75.9 & 98.15  \\
\hdashline
250 &35 &215 &74.03 & 99.34 & 74.77 & 99.09 & 77.22& 98.70  \\
300 &46& 254& 80.33 & 99.34 & 86.18 & 98.31 & 87.03 & 98.11\\
350 &51 & 299 & 87.11 & 99.39 &  57.73 & 99.77 & 86.99 & 98.66 \\
400 &58 & 342& 87.18 & 99.37 & 83.87 & 98.73 & 87.44& 98.75\\
450 &65&385 & 87.86& 99.46 & 84.82 & 98.89& 88.15& 97.46\\
500 &74 & 426 & 89.83 & 99.39 & 90.76 & 98.44 & 90.90& 98.81\\
550 &82&468 & 90.16 & 99.44& 91.27 & 98.18& 91.07& 98.58 \\
\hline
\end{tabular}
}
\end{center}
\caption{Error values for different Gaussian process classification methods for the modified Drop-Wave function and different sample sizes generated with \gls{tplhd}}\label{table::disc_compare}
\end{table}

 In view of $n_{C1}$ and $n_{C2}$ the total number of samples that yield an output of the respective classes, it can be seen for \gls{tplhd} sample size of 50 or 100 samples, that with few samples detected in class $\mathcal{C}_1$, the kriging approach is able to build a classification metamodel describing this minor class subdomain, whereas both Laplace approximation and expectation propagation are not able to correctly detect even one point in class $\mathcal{C}_1$ among the reference points.

\subsection{Mechanical problems}

In the following \gls{mivor} is tested for the generation of classification surrogate models in the context of two mechanical applications. The first case is the classification of a damage measure for maintenance decision, distinguishing between required maintenance or non-required maintenance as classification output. The second surrogate model aims at classifying a dynamic system into stable and chaotic motion.

\subsubsection{Classification for maintenance decision making}

In this first problem a classification of a two-dimensional parametric domain for maintenance decision is of interest. It is assumed that maintenance is required when the damage parameter $D$ exceeds a value of $0.02$.

A two-scale damage model dedicated to high-cycle fatigue in the context of continuum damage mechanics is considered. The damage accumulation is governed by micro-plasticity phenomena, the details of it, which are not here required, are given in \cite{Damage_model}. Only one Gauss point is considered to evaluate the effect of an overload during cyclic fatigue damage. The goal is to predict the parametric subdomain which requires maintenance after a total number of 100000 loading cycles depending on the frequency and the amplitude of the overload cycles.

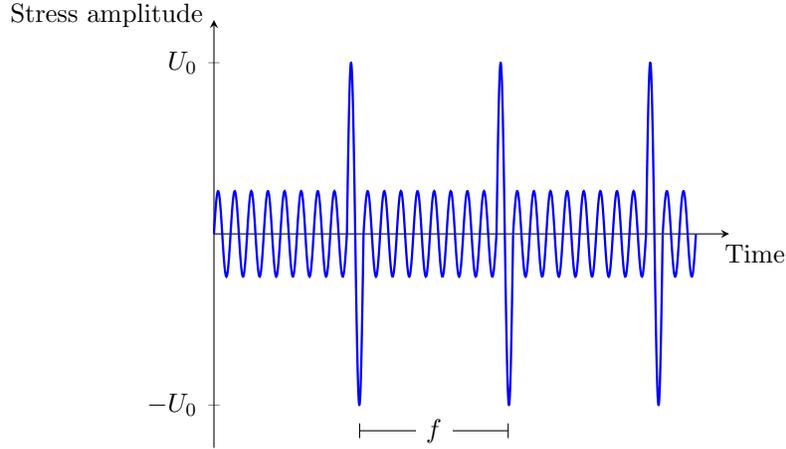
\begin{figure}[htbp!]
\centering
\begin{tikzpicture}[scale = 1.0, transform shape]
\begin{axis}
[
xlabel={\footnotesize{Time}},
ylabel={\footnotesize{Stress amplitude}},
y label style={anchor=north},
axis lines=middle,
xtick={0},
xticklabels={},
    ytick={-1.00,0,1.00},
    ymin=-1.250,ymax=1.250,
    xmin=-0.1,xmax=62*pi,
    axis on top=false,
    yticklabels={\footnotesize{$-U_{0}$},0,\footnotesize{$U_{0}$}},
every axis x label/.style={
    at={(ticklabel* cs:1.05)},
    anchor=north,
},
every axis y label/.style={
    at={(ticklabel* cs:1.01)},
    anchor=east,
},
]
\addplot[domain=0:16*pi,blue,line width = 0.3mm,samples=300]{0.25*sin(deg(x))};%
\addplot[domain=16*pi:18*pi,blue,line width = 0.3mm,samples=300]{1.00*sin(deg(x))};%
\addplot[domain=18*pi:34*pi,blue,line width = 0.3mm,samples=300]{0.25*sin(deg(x))};%
\addplot[domain=34*pi:36*pi,blue,line width = 0.3mm,samples=300]{1.00*sin(deg(x))};%
\addplot[domain=36*pi:52*pi,blue,line width = 0.3mm,samples=300]{0.25*sin(deg(x))};%
\addplot[domain=52*pi:54*pi,blue,line width = 0.3mm,samples=300]{(1.00*sin(deg(x))};%
\addplot[domain=54*pi:58*pi,blue,line width = 0.3mm,samples=300]{0.25*sin(deg(x))};%
\draw[|-|] (17.5*pi, 10.0) -- (35.5*pi,10.0) node[midway, fill=white] {\footnotesize{$f$}};
\end{axis}
\end{tikzpicture}
\caption{Two-dimensional parametric fatigue loading} \label{2D_load}
\end{figure}

The parametric study focuses thus on load description. It consists, as illustrated in Figure \ref{2D_load}, on perfectly periodic loading of amplitude 233 MPa with some regular overload cycles, which have a constant amplitude $U_{0} \in [233, 370] \text{MPa} $ and occur regularly after each block of amplitude 233 MPa comprising $f_{0} \in [50, 10000]$ cycles.

\begin{figure}[ht!]
\centering
\begin{subfigure}[t]{0.5\textwidth}
\includegraphics[width=\textwidth]{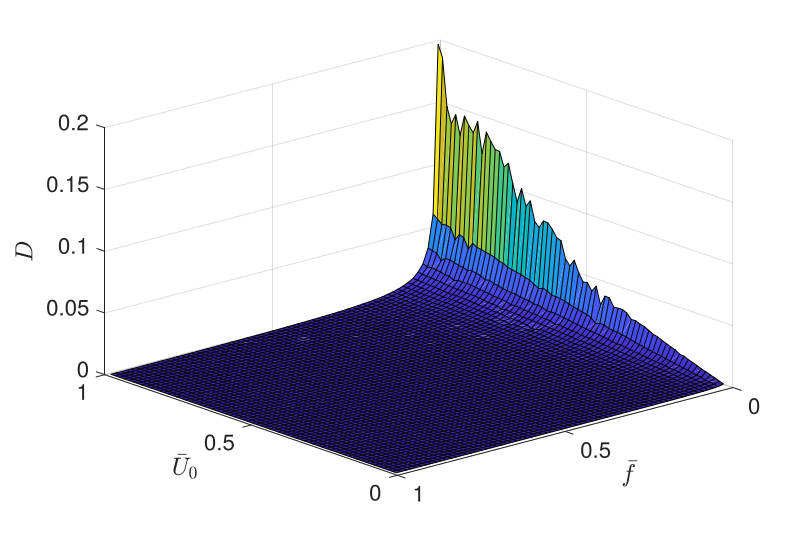}
\subcaption{Damage response surface}\label{fid::damage_surf}
\end{subfigure}%
\begin{subfigure}[t]{0.5\textwidth}
\includegraphics[width=\textwidth]{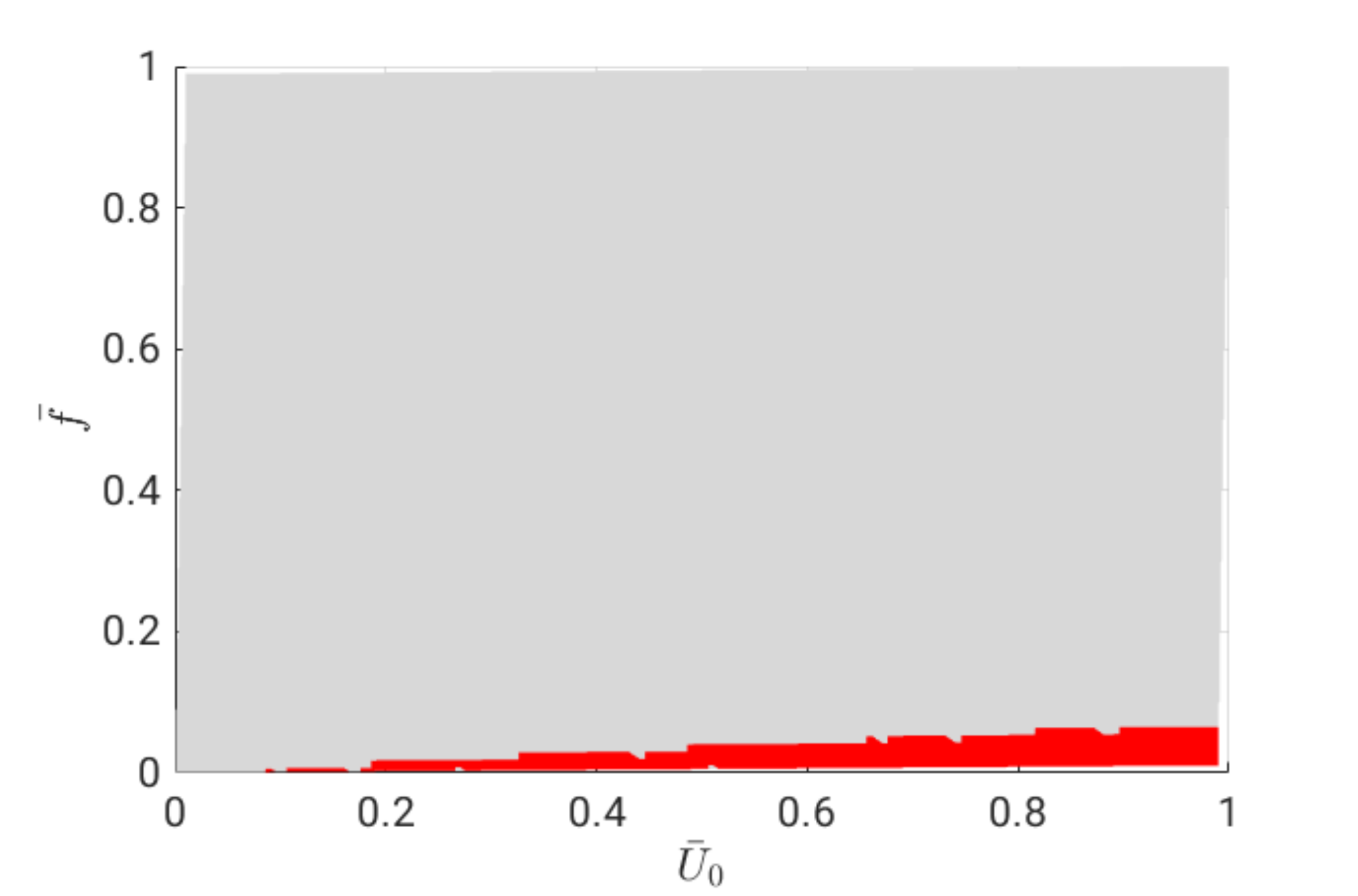}
\subcaption{Reference classification}\label{fid::damage_class}
\end{subfigure} \\
\begin{subfigure}[t]{0.5\textwidth}
\includegraphics[width=\textwidth]{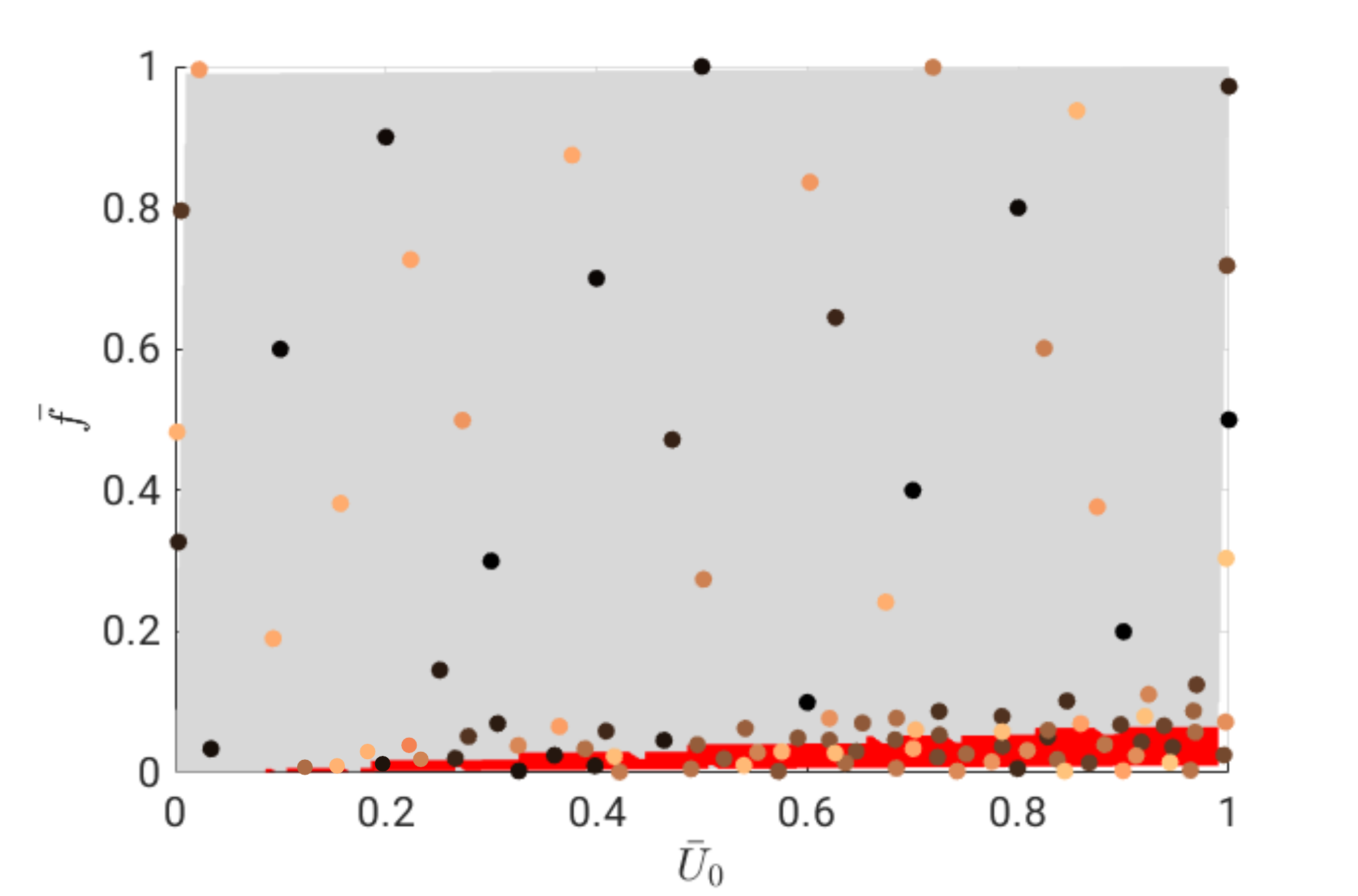}
\subcaption{Example set of 100 Samples}\label{fid::damage_samples}
\end{subfigure}%
\begin{subfigure}[t]{0.5\textwidth}
\includegraphics[width=\textwidth]{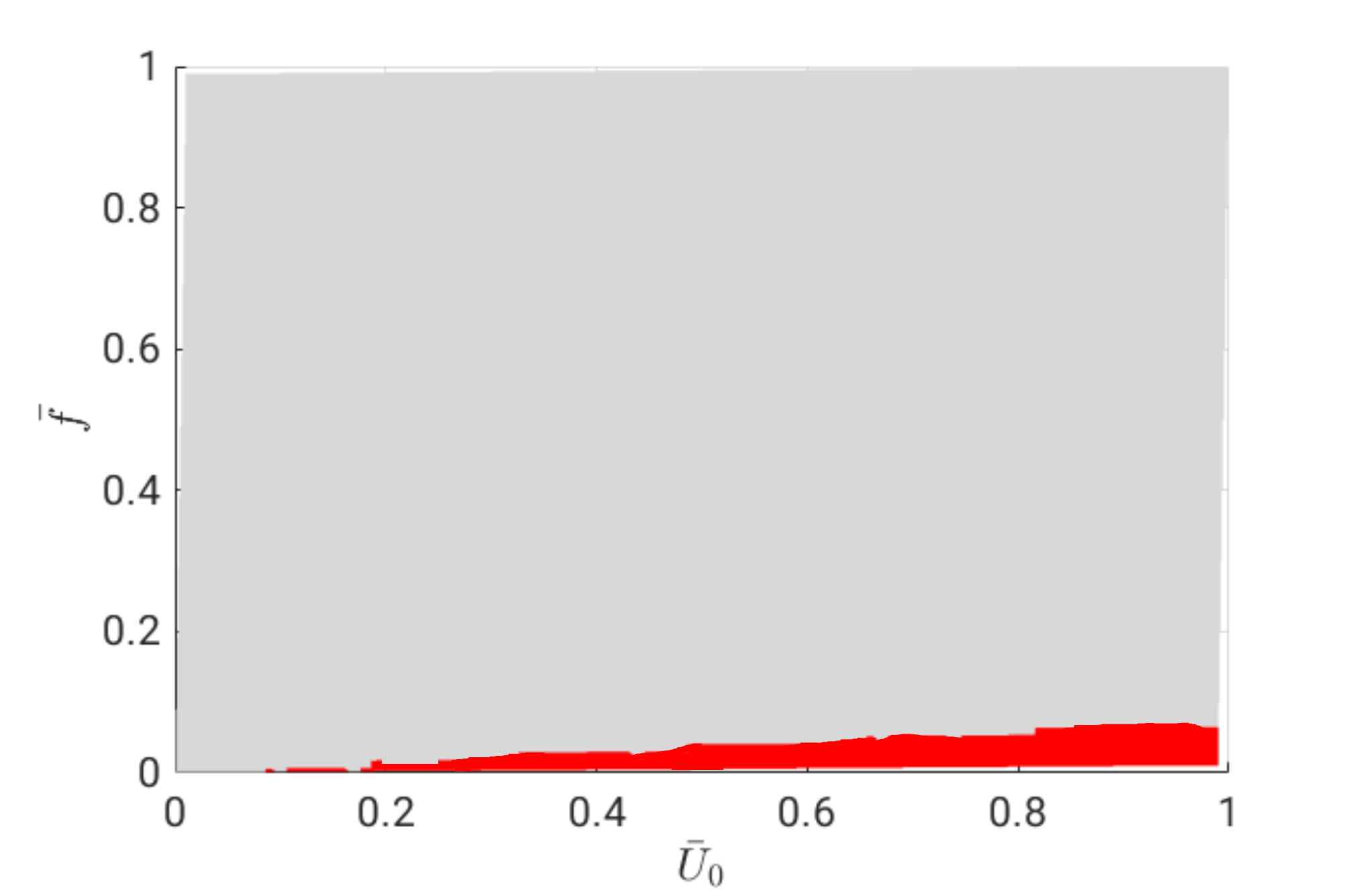}
\subcaption{\gls{mivor} surrogate classification}\label{fid::damage_meta}
\end{subfigure}
\caption{\gls{mivor} based on 100 samples for two-dimensional damage example.}\label{fig::class_damage_mivor}
\end{figure}

The reference response surface obtained from 10000 \gls{tplhd} points is plotted in Figure \ref{fid::damage_surf} over the normalized input domain. The resulting class labels over the normalized parametric space are displayed in Figure \ref{fid::damage_class}. The largest part of the parametric domain does not require maintenance after 100000 cycles, whereas the subdomain requiring maintenance represents a very small part of it. Initially $10$ samples are generated with \gls{tplhd} and $90$ supplementary samples are created adaptively. An example of \gls{mivor} dataset is shown in Figure \ref{fid::damage_samples}. It can be seen that the exploration component evenly spreads samples across the domain. The exploitation predominantly adds samples within and around the $\mathcal{C}_{1}$ region. The resulting metamodel created with these samples is shown in Figure \ref{fid::damage_meta}. It is in very good accordance with the reference classification (Figure \ref{fid::damage_class}).

\begin{figure}[ht!]
\centering
\begin{subfigure}[t]{0.5\textwidth}
\includegraphics[width=\textwidth]{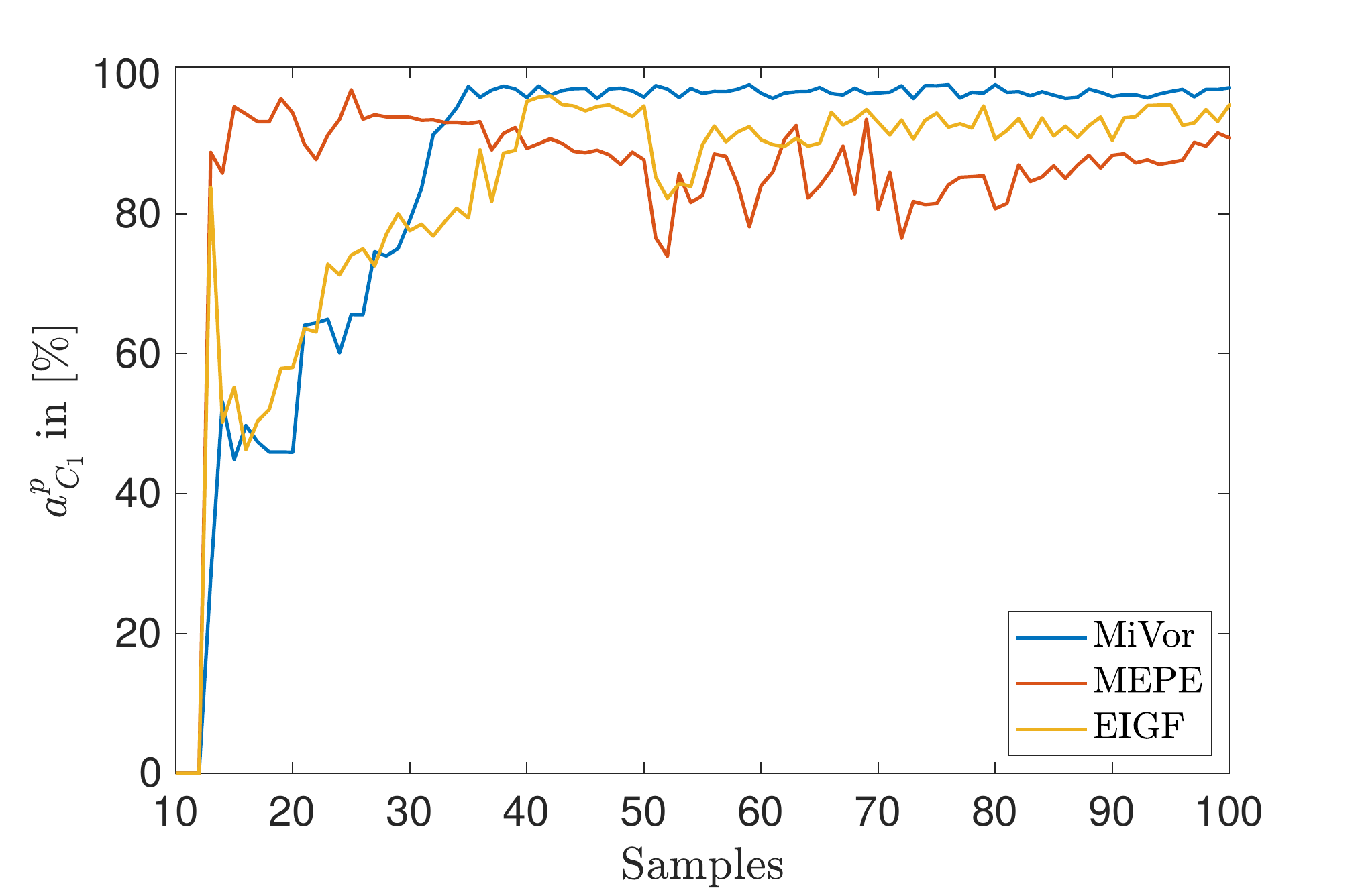}
\subcaption{$\mathcal{C}_{1}$}\label{fig::data_dataAbove}
\end{subfigure}
\begin{subfigure}[t]{0.5\textwidth}
\includegraphics[width=\textwidth]{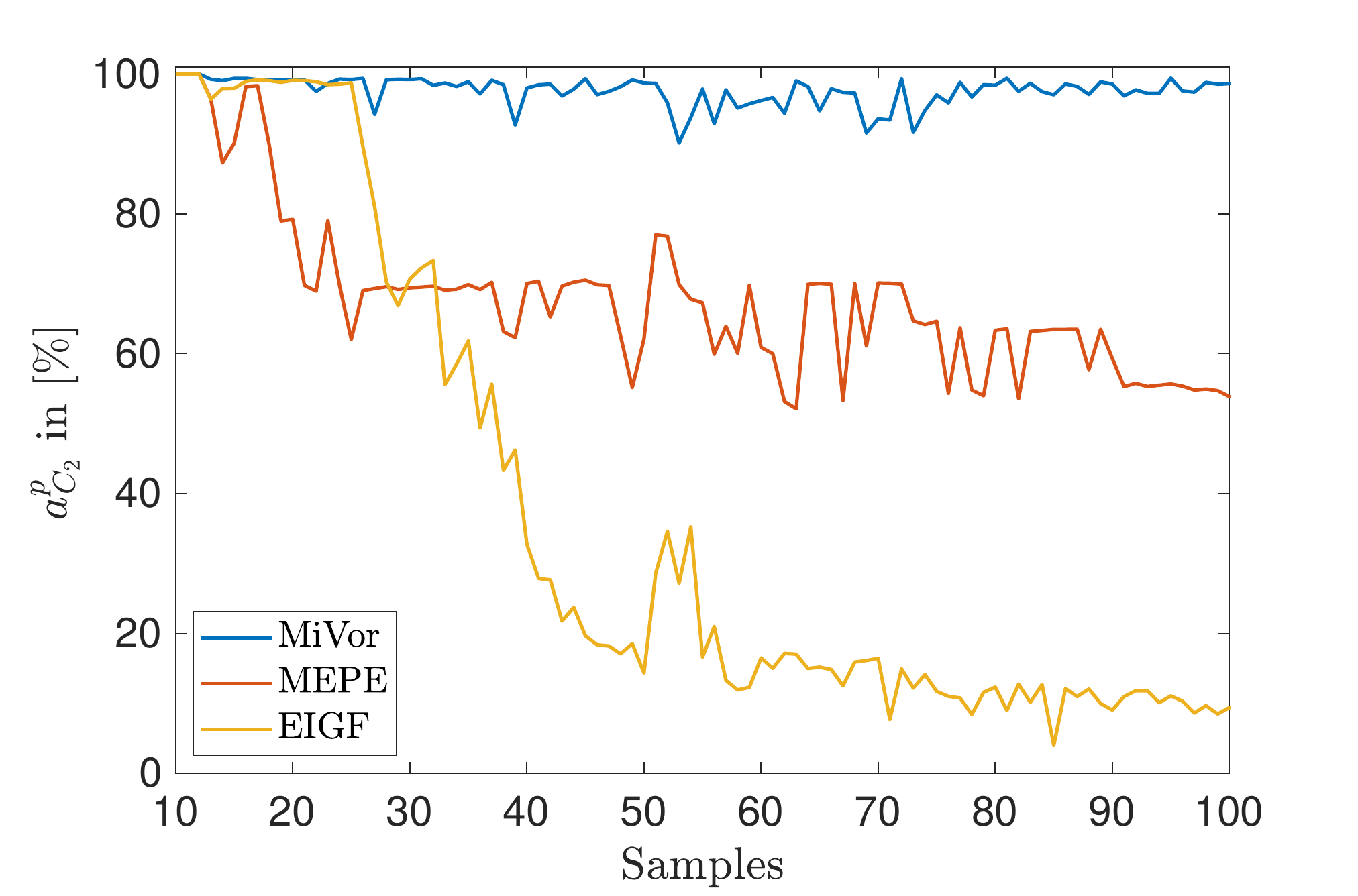}
\subcaption{$\mathcal{C}_{2}$}\label{fig::data_databelow}
\end{subfigure} 
\caption{Averaged error values for two-dimensional damage problem with binary output for different adaptive sampling techniques.}\label{fig::damage_date}
\end{figure}

The evolution of the error measures during the sampling processes for \gls{mivor}, \gls{mepe} and \gls{eigf} is plotted in Figure \ref{fig::damage_date}. The $a^{p}_{C_{1}}$ metric is displayed in Figure \ref{fig::data_dataAbove} over the number of samples in the dataset. For this measure, from an initial accuracy of $0\, \%$, all three adaptive techniques come close to acceptable prediction capability however the performance of \gls{mivor} is clearly superior. It can also be mentioned that the fluctuation of the curves is due to the particular shape of the response surface (see Figure \ref{fid::damage_surf}), which for the most part is nearly constant. \gls{mepe} and \gls{eigf} face here point clustering problems leading to numerical issues for the kriging method. In the proposed implementation of \gls{mivor} this problem is circumvented  by preventing sample point clustering as introduced in section \ref{sec::Clustering}.

\begin{figure}[ht!]
\centering
\begin{subfigure}[t]{0.5\textwidth}
\includegraphics[width=\textwidth]{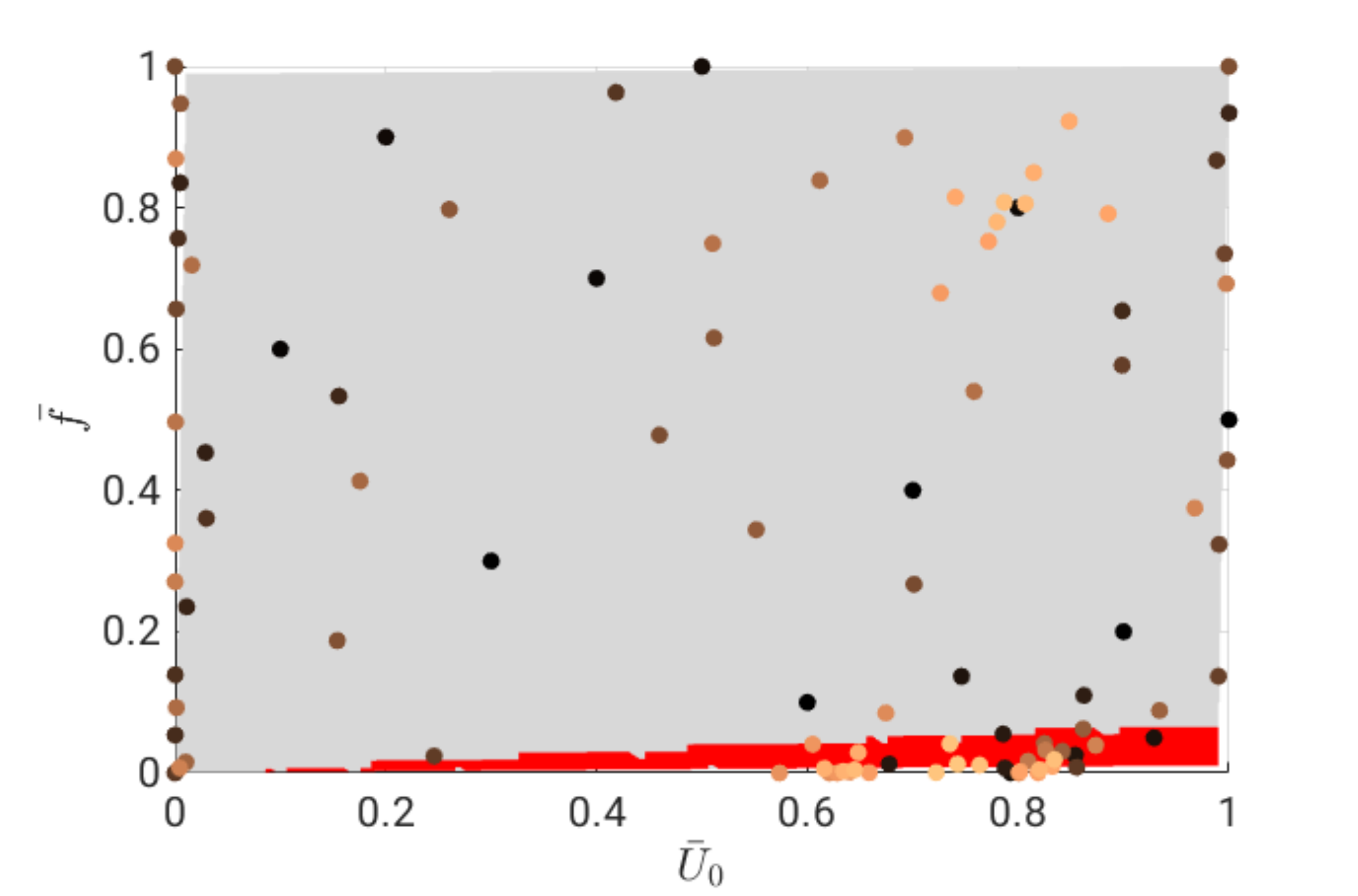}
\subcaption{\gls{mepe}}\label{fig:damage_MEPE_samples}
\end{subfigure}%
\begin{subfigure}[t]{0.5\textwidth}
\includegraphics[width=\textwidth]{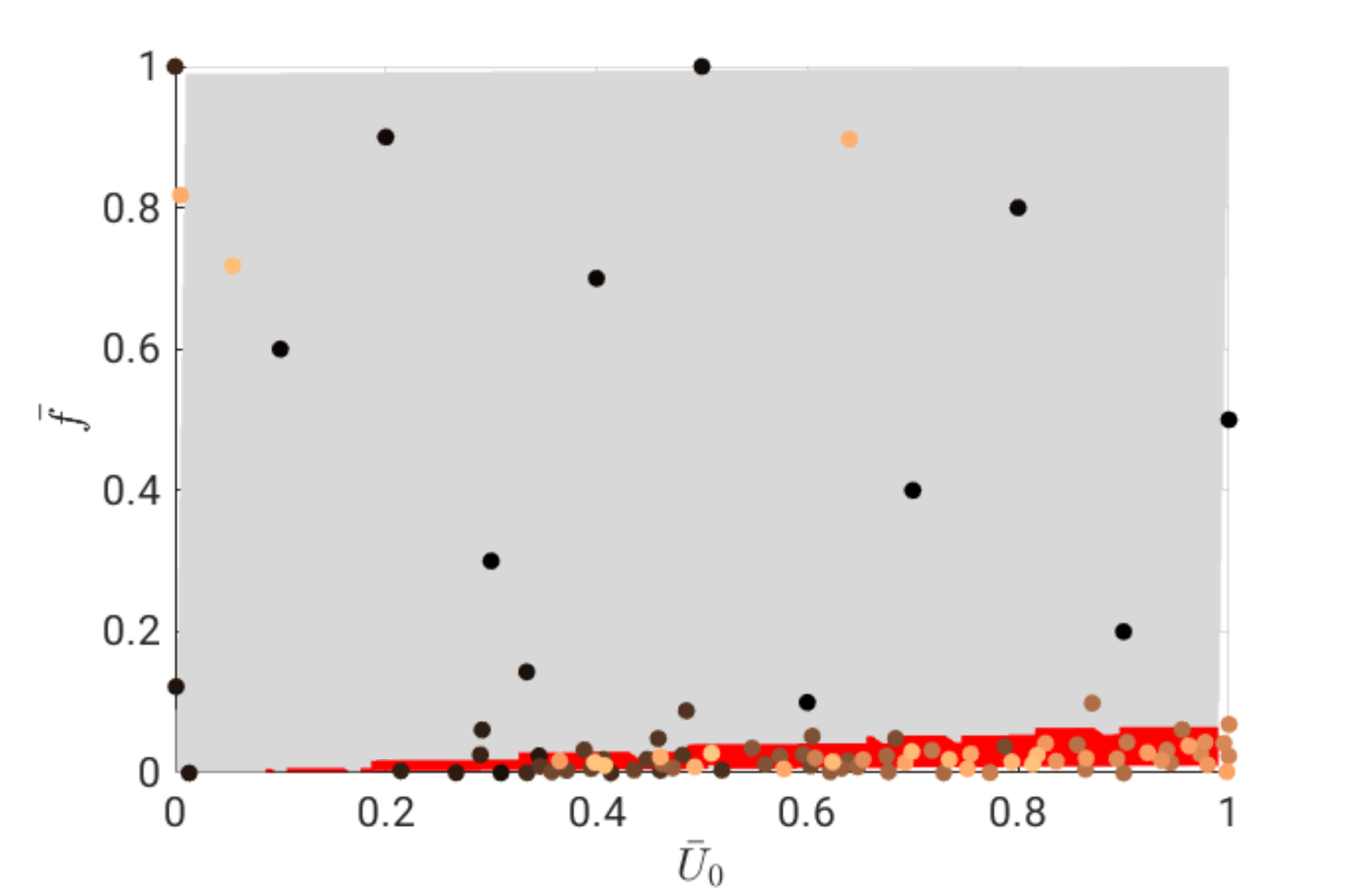}
\subcaption{\gls{eigf}}\label{fig:damage_EIGF_samples}
\end{subfigure}
\caption{Example \gls{mepe} and \gls{eigf} sets of 100 samples for damage classification in two-dimensional normalized parametric space.}\label{fig::class_damage}
\end{figure}

Concerning $a^{p}_{C_2}$ error shown in Figure \ref{fig::data_databelow}, \gls{mivor} is able to maintain a rather good prediction of this class while adding samples, whereas \gls{mepe} and \gls{eigf} show a drastic decrease of this metric over the adaptive process. This is due to point clustering issues as seen in Figure \ref{fig:damage_MEPE_samples} for \gls{mepe} and \ref{fig:damage_EIGF_samples} for \gls{eigf}.

This example appears rather simple in terms of classification as the quantity of interest on which the classification is based on a monotonic function in all the dimensions of the parametric domain. Let consider in a second case a quantity of interest which does not have a monotonic evolution.

\subsubsection{\gls{lle} classification}

The goal of the second application is to provide a classification surrogate model for a dynamic system, such that the parametric domain can be easily divided into regions of stable and unstable motion.

The Largest Lyapunov Exponent (LLE) \citep{oseledec1968multiplicative} is utilized as an indicator for chaotic motion of a dynamical system \citep{kocarev2006discrete}, i.e. positive \gls{lle} values indicate chaotic motion, whereas \gls{lle} values below zero point towards a stable behavior. Here again the precise value of the \gls{lle} on the whole response surface is not strictly required by the engineer, as the main concern is on the class label of the system.

\begin{figure}[htbp!]
\centering
  \includegraphics[width = 0.5\textwidth]{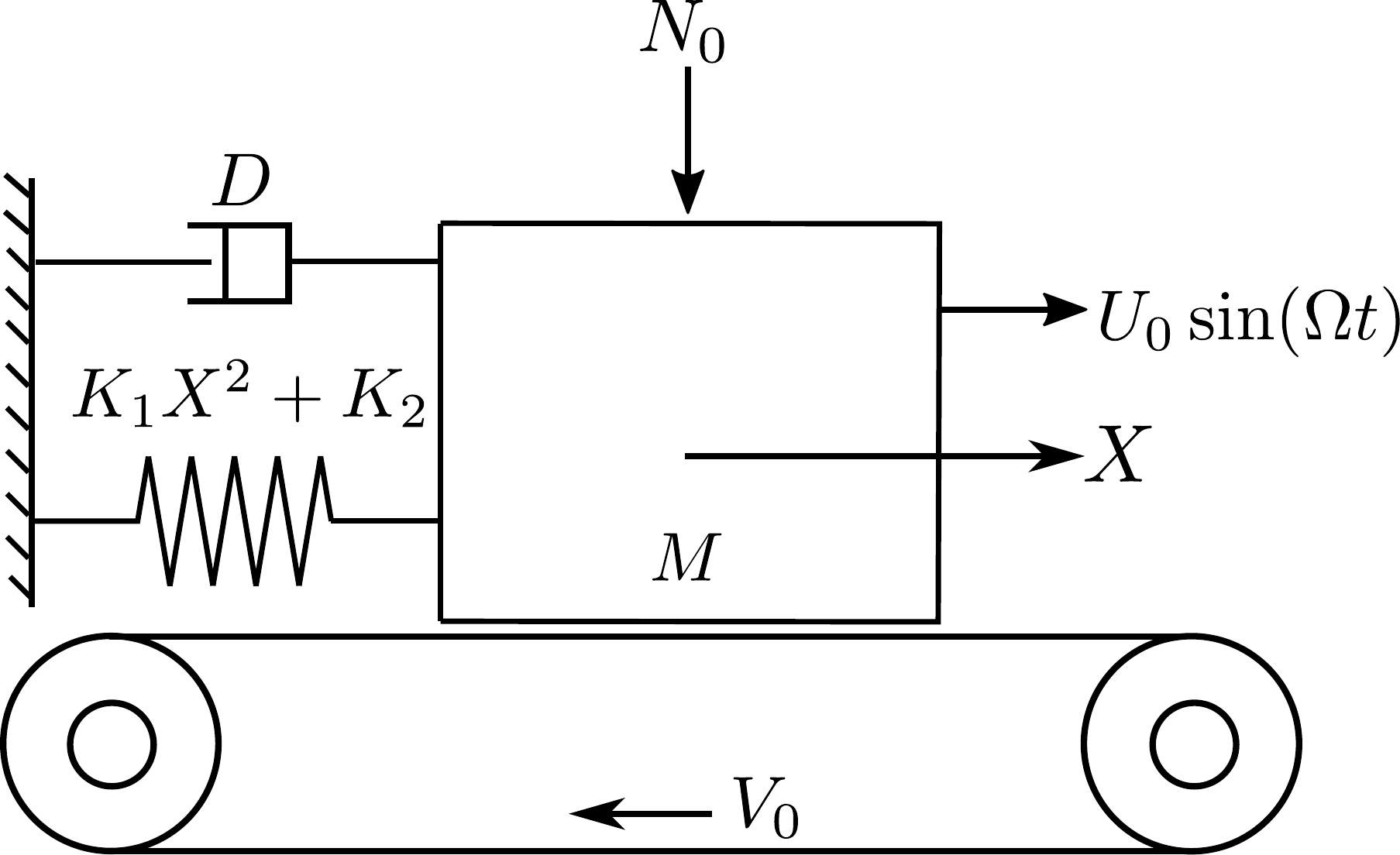}
\vspace{0.5cm}
\caption[Scheme of the analyzed nonlinear mass-on-belt system]{Scheme of the analyzed nonlinear mass-on-belt system.}
\label{fig:Application}       
\end{figure}

The mass-on-belt system of interest, a Duffing's type oscillator, is schematised in Figure \ref{fig:Application}. A rigid body with mass $M$ is placed on a belt moving with constant speed $V_{0}$. The motion of the mass described through its displacement $X$ is restricted by a dashpot with damping coefficient $D$ and a nonlinear spring with constants $K_{1}$ and $K_{2}$. A normal load $N_{0}$ as well as a time-dependent harmonic force with angular frequency $\Omega$ and amplitude $U_{0}$ is applied on the mass.
The mass is subject to friction on the moving belt governed by an elasto-plastic friction law as proposed by \cite{dupont2002single}. For more information see \cite[]{Jan_Master_thesis}. The LLE of the system is approximated by the algorithm of \cite{wolf1986quantifying}, in which the Jacobian matrix of the dynamic system is estimated from the numerical scheme proposed in \cite{balcerzak2018spectrum}.

\begin{figure}[htbp!]
\centering
\begin{subfigure}[t]{0.5\textwidth}
\includegraphics[width=\textwidth]{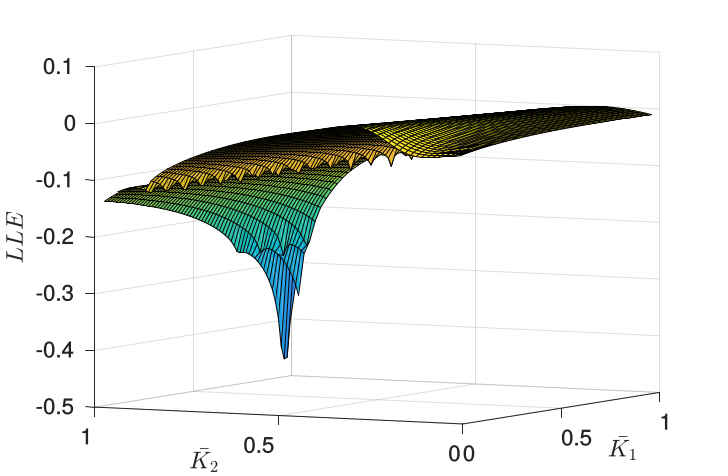}
\subcaption{Response surface}\label{fig::P1_surf}
\end{subfigure}%
\begin{subfigure}[t]{0.5\textwidth}
\includegraphics[width=\textwidth]{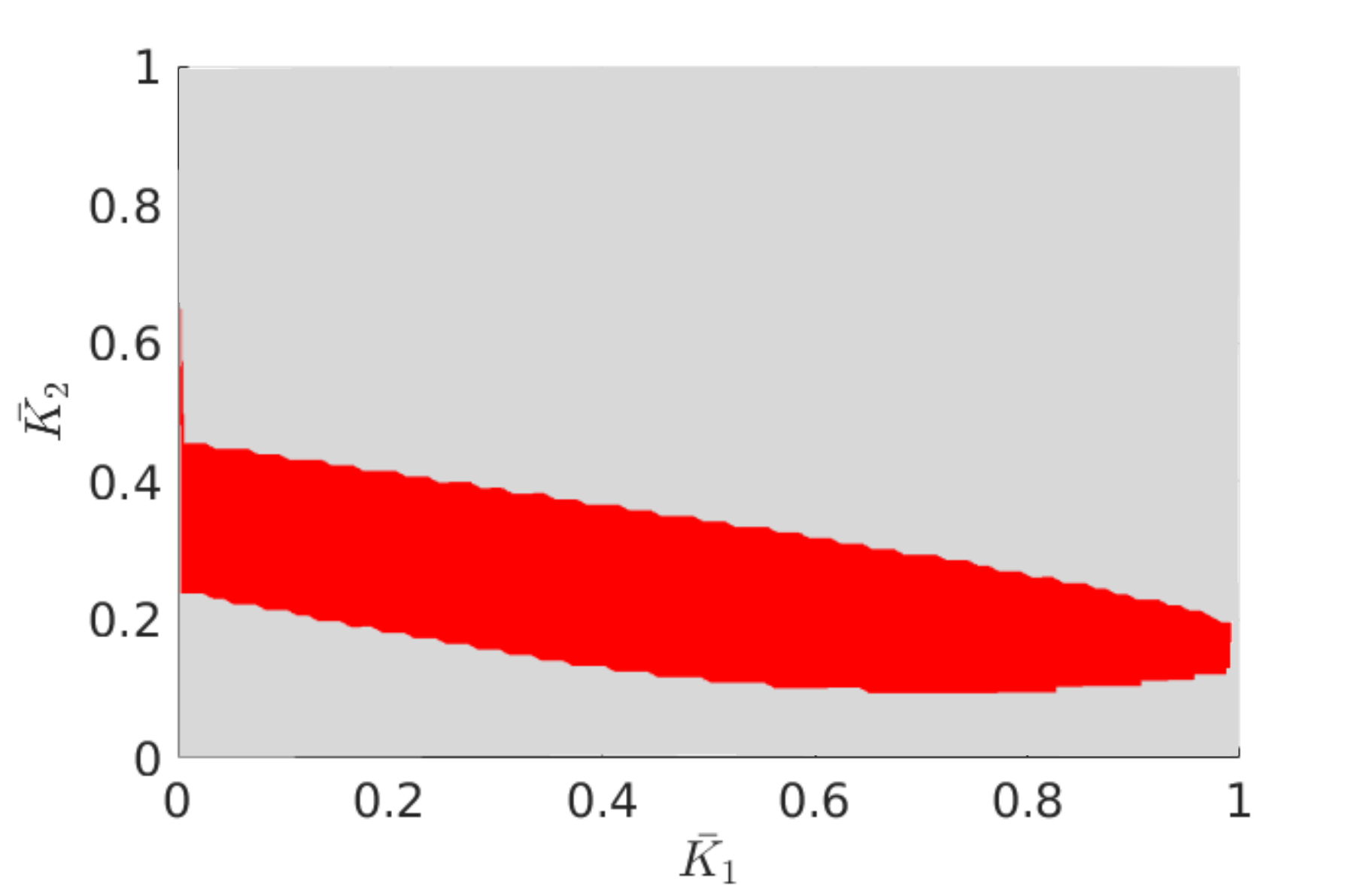}
\subcaption{Classification output}\label{fig::P1_class}
\end{subfigure} \\
\begin{subfigure}[t]{0.5\textwidth}
\includegraphics[width=\textwidth]{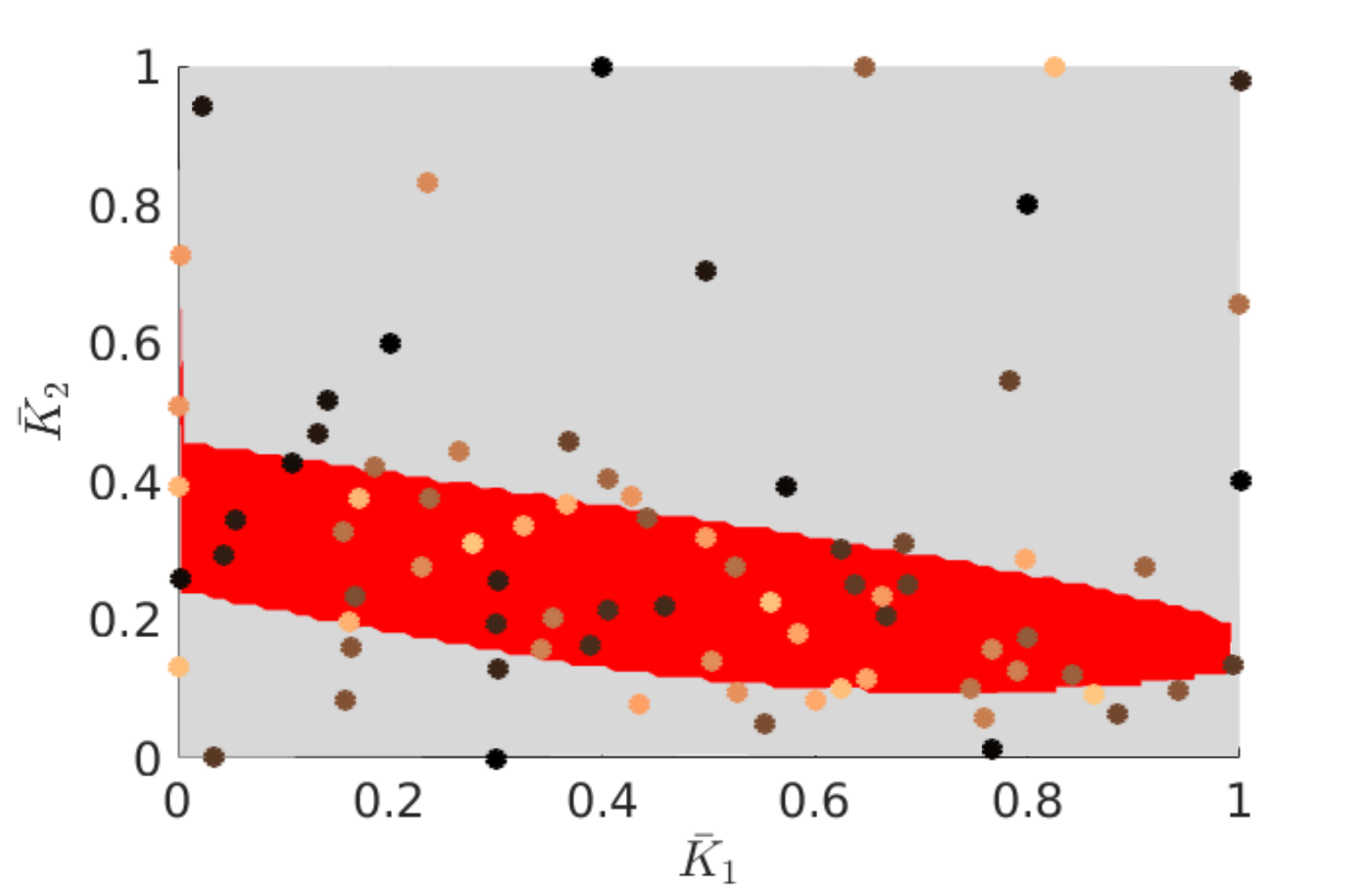}
\subcaption{Samples}\label{fig::P1_sample}
\end{subfigure}%
\begin{subfigure}[t]{0.5\textwidth}
\includegraphics[width=\textwidth]{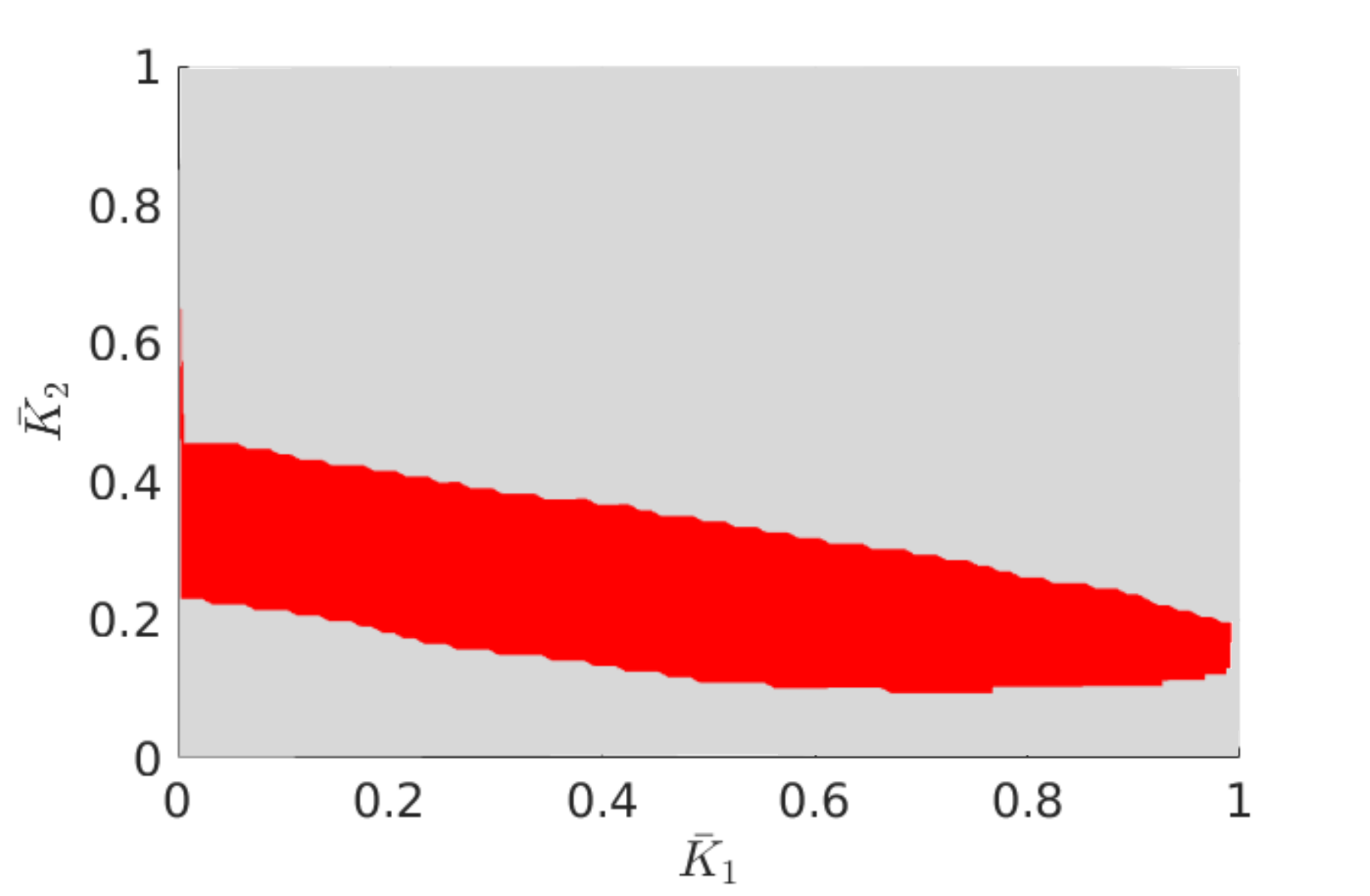}
\subcaption{Evaluated metamodel}\label{fig::P1_meta}
\end{subfigure}
\caption[Sample positions for different adaptive sampling techniques for the classification of chaotic motion]{\gls{mivor} for two-dimensional LLE problem after 60 samples. }\label{fig:SampleKSeta06Com}
\end{figure}

Consider the following set of deterministic parameters including the static friction component $\mu_{s} = 0.3$, the kinetic friction value $\mu_{k} = 0.15$ as well as
$M = 1 \, \text{kg}$,
$V_{0} = 0.1 \, \text{m}/\text{s}$,
$D = 0.0 \, \text{Ns}/\text{m}$,
$V_{0} = 0.1 \, \text{m}/\text{s}$,
$U_{0} = 0.1 \,\text{N}$,
$N_{0} = 1.0 \, \text{N}$ and $\Omega = 0.6 \, \text{rad}/\text{s}$.
The elasto-plastic friction force model consists of three additional parameters $\sigma_{0}$, $\sigma_{1}$ and $\sigma_{3}$, the values of which are chosen to be
$\sigma_{0} = 100.0 \, \text{N}/\text{m} $,
$\sigma_{1} = 10.0 \,  \text{Ns}/\text{m}$ and
$\sigma_{2} = 0.1 \, \text{Ns}/\text{m}$. Furthermore consider the input domain for the two spring stiffnesses given by \\$K_{1} \in  [0.5 , 1.0]\, \text{N}/\text{m}^{3}$
and $K_{2} \in [0.0,0.6] \, \text{N}/\text{m}$. The \gls{lle} response surface over this domain is plotted in Figure \ref{fig::P1_surf}. The function is not monotonic in both directions. The reference classification problem is displayed in Figure \ref{fig::P1_class}. It can be seen that the shape of the area belonging to \gls{lle} values above or equal $0$ is fairly simple, but embedded in the subdomain of class $\mathcal{C}_2$. Values with $LLE\geq 0$ lie predominantly in the middle of the given domain and are shaped like an ellipse. 
The problem is studied until 60 samples are reached starting from 5 samples created with \gls{tplhd}. An example set of \gls{mivor} sample points is shown in Figure \ref{fig::P1_sample}. It can be seen that a majority of the points are spent by exploitation to investigate in details the single $\mathcal{C}_{1}$ subdomain and so localize precisely the boundary between $\mathcal{C}_{1}$ and $\mathcal{C}_{2}$. The exploration component of the method spreads the rest of the points evenly in the parametric domain. It can be seen in Figure \ref{fig::P1_meta} that the output surrogate classification evaluated for the whole set of reference sample points successfully matches the reference solution of Figure \ref{fig::P1_class} from the knowledge of only 60 observations.

\begin{figure}[htbp!]
\centering
\begin{subfigure}[t]{0.49\textwidth}
\includegraphics[width=\textwidth]{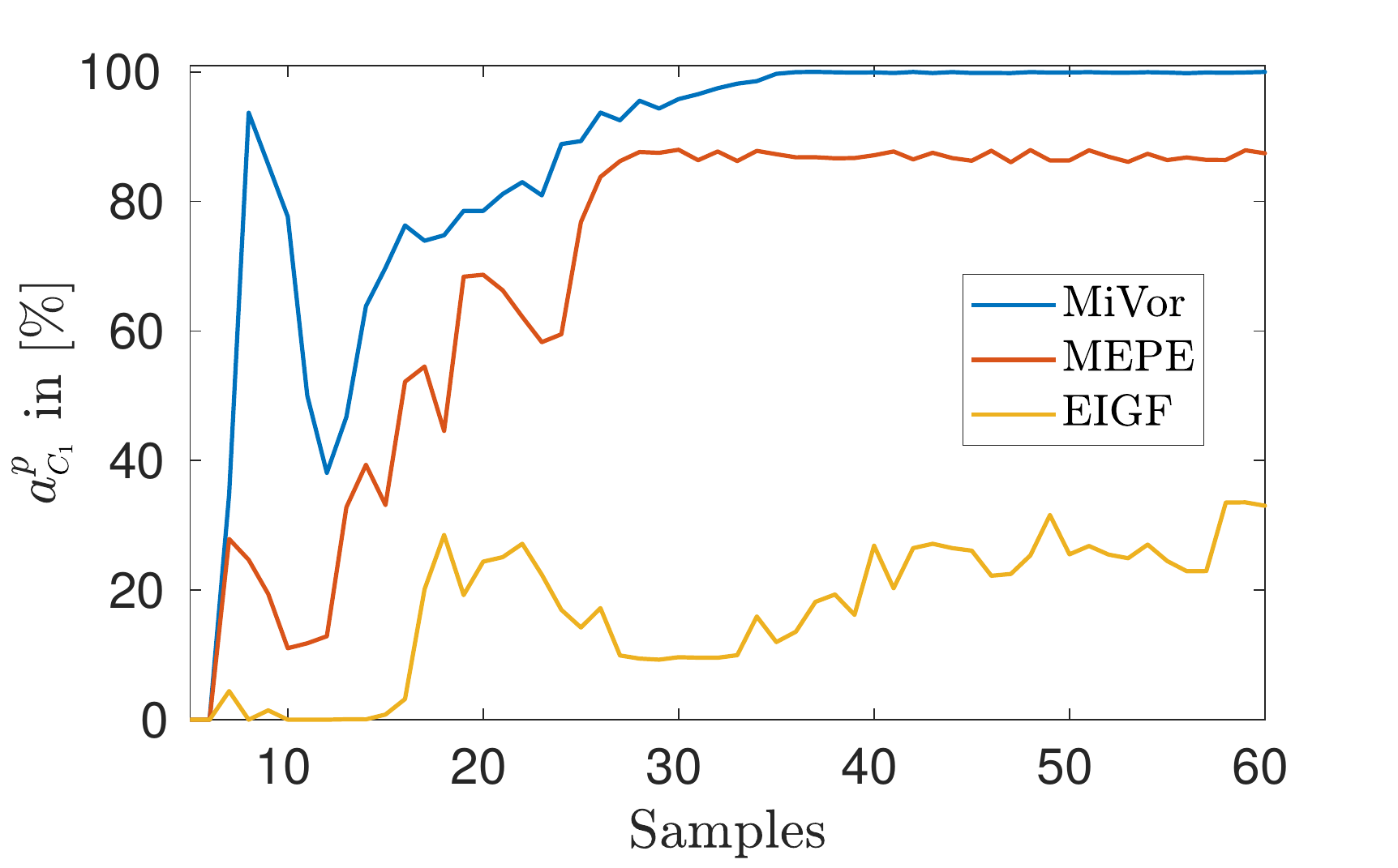}
\subcaption{$\mathcal{C}_{1}$}\label{fig::LLE_above}
\end{subfigure}
\begin{subfigure}[t]{0.49\textwidth}
\includegraphics[width=\textwidth]{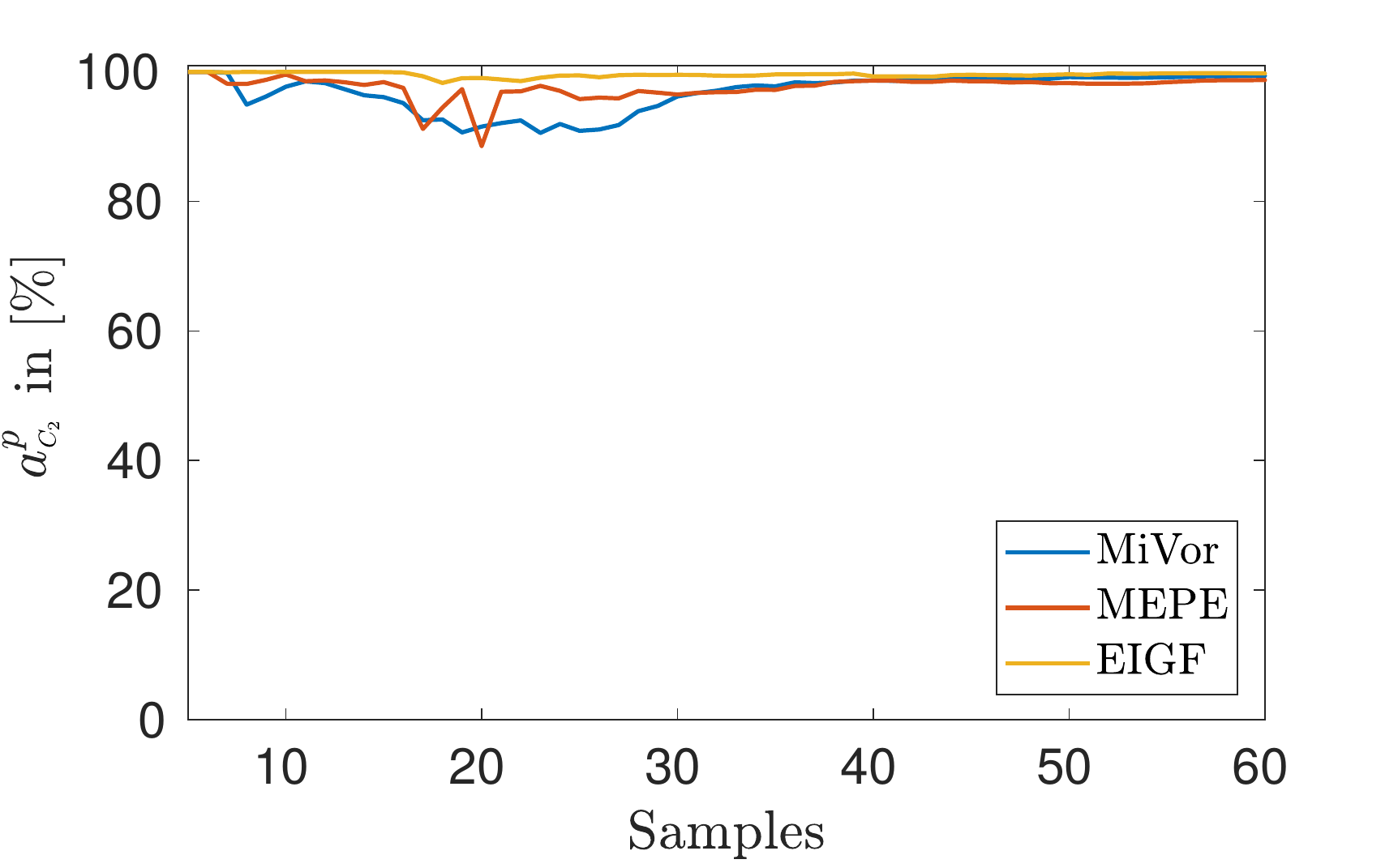}
\subcaption{$\mathcal{C}_{2}$}\label{fig::LLE_below}
\end{subfigure}%
\caption[]{Averaged error values for two-dimensional LLE problem with binary output for different adaptive sampling techniques.}\label{fig::LLE_data}
\end{figure}

The averaged error metrics evaluated on 20 independent realizations for \gls{mivor}, \gls{mepe} and \gls{eigf} are plotted over the sample size of the dataset in Figure \ref{fig::LLE_data}. It can be noticed that the value of the $a^{p}_{C_{2}}$ measure (Figure \ref{fig::LLE_below}) is around the optimum for all of the methods along the process. However there are crucial differences for $a^{p}_{C_{1}}$ as shown in Figure \ref{fig::LLE_above}, where \gls{mivor} largely outperforms the other two techniques by reaching an optimal metamodel able to correctly identify 100 $\%$ of the points belonging to class $\mathcal{C}_1$ after adding only around 35 samples.

\begin{figure}[htbp!]
\centering
\begin{subfigure}[t]{1.0\textwidth}
\includegraphics[scale=0.35]{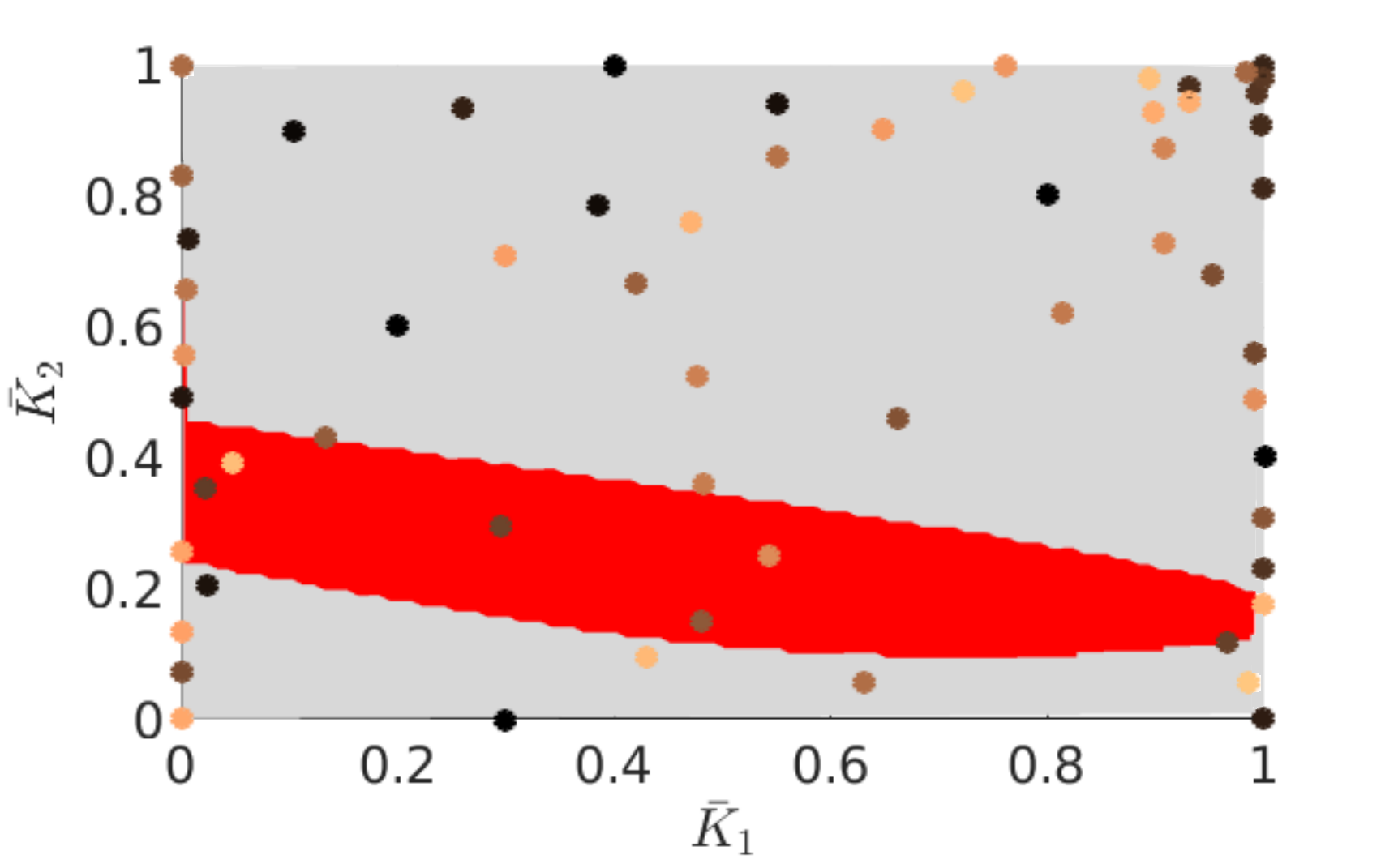}%
\includegraphics[scale=0.35]{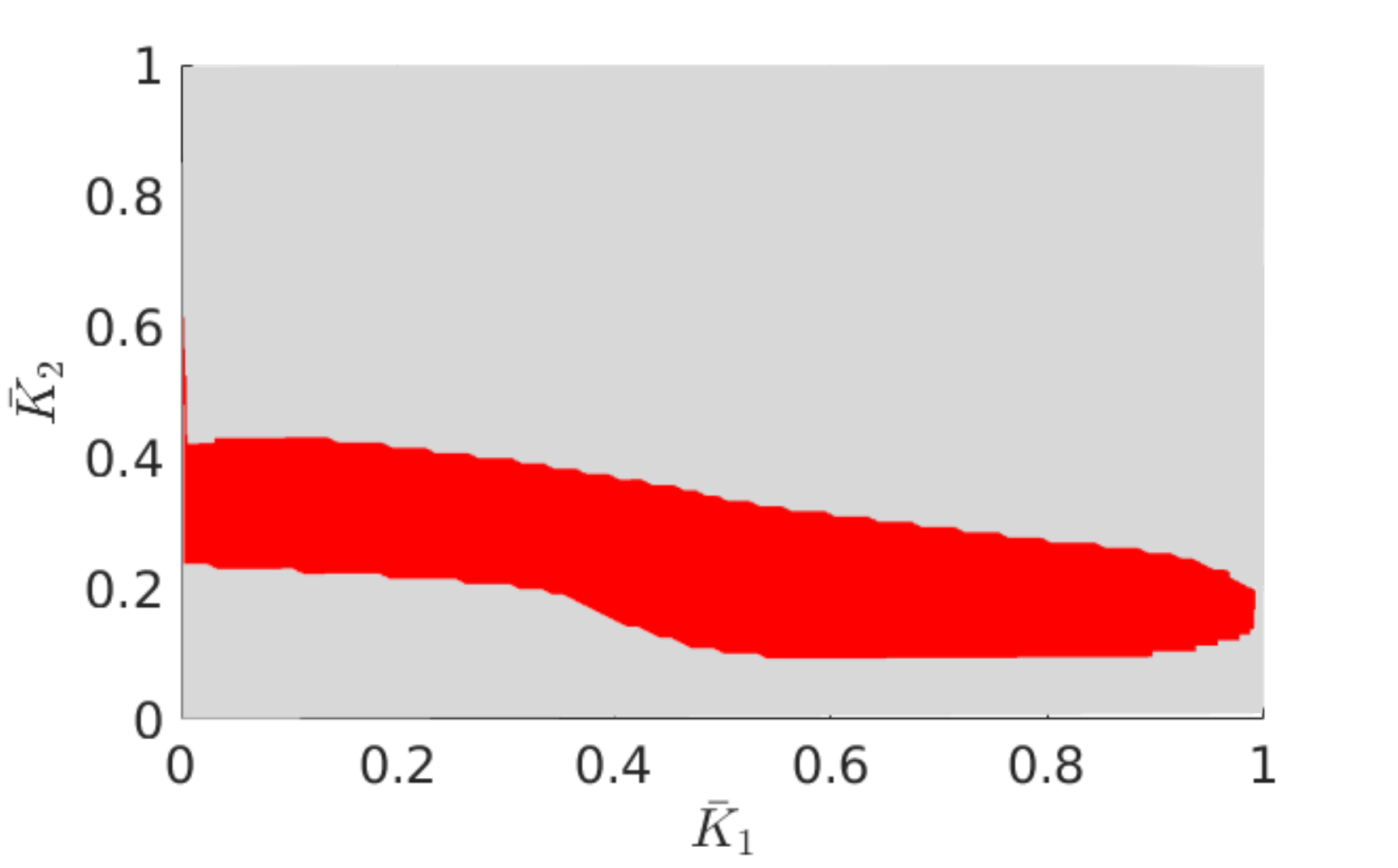}
\subcaption{\gls{mepe}}\label{fig::LLE_MEPE}
\end{subfigure} \\
\centering
\begin{subfigure}[t]{1.0\textwidth}
\includegraphics[scale=0.35]{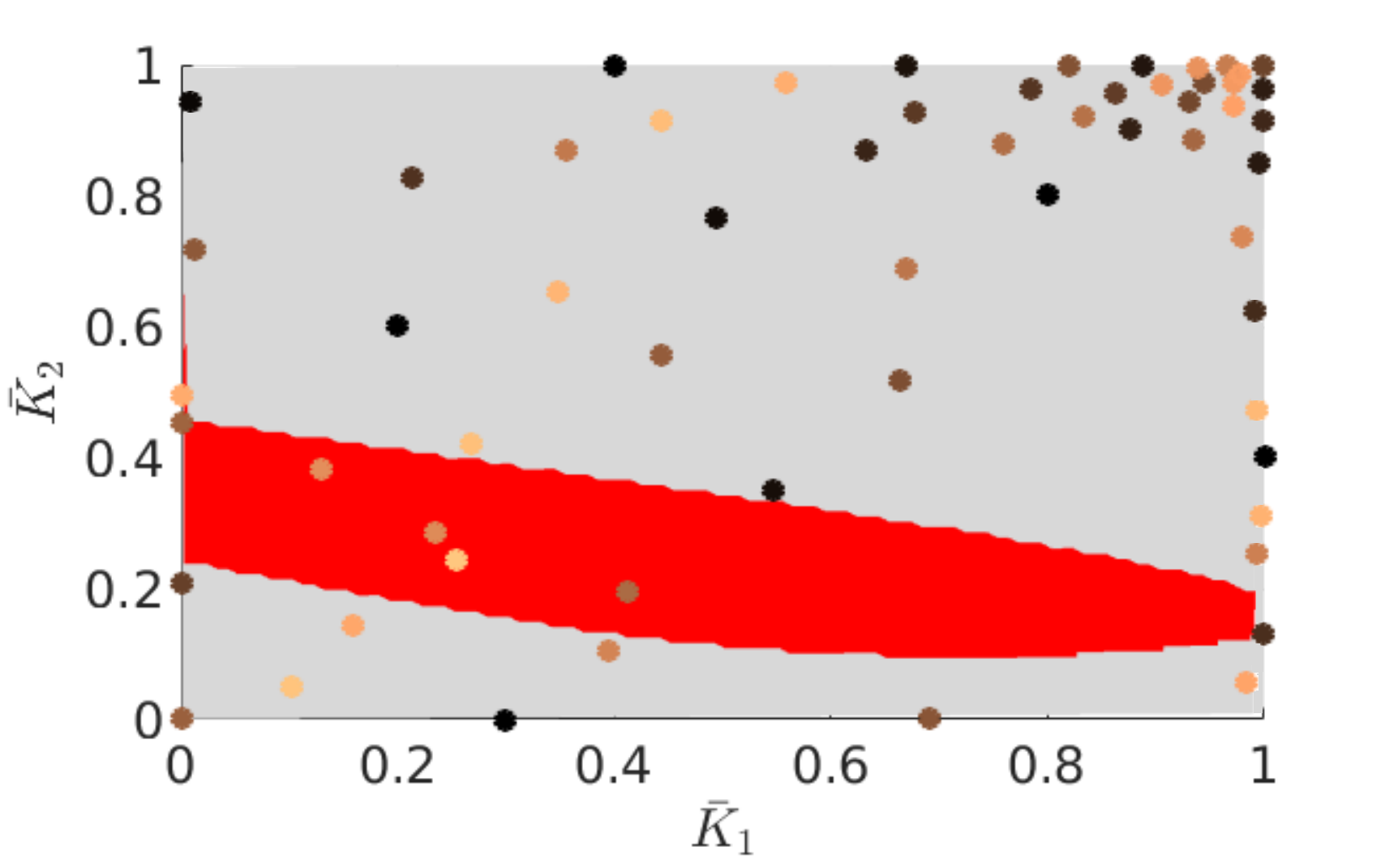}%
\includegraphics[scale=0.35]{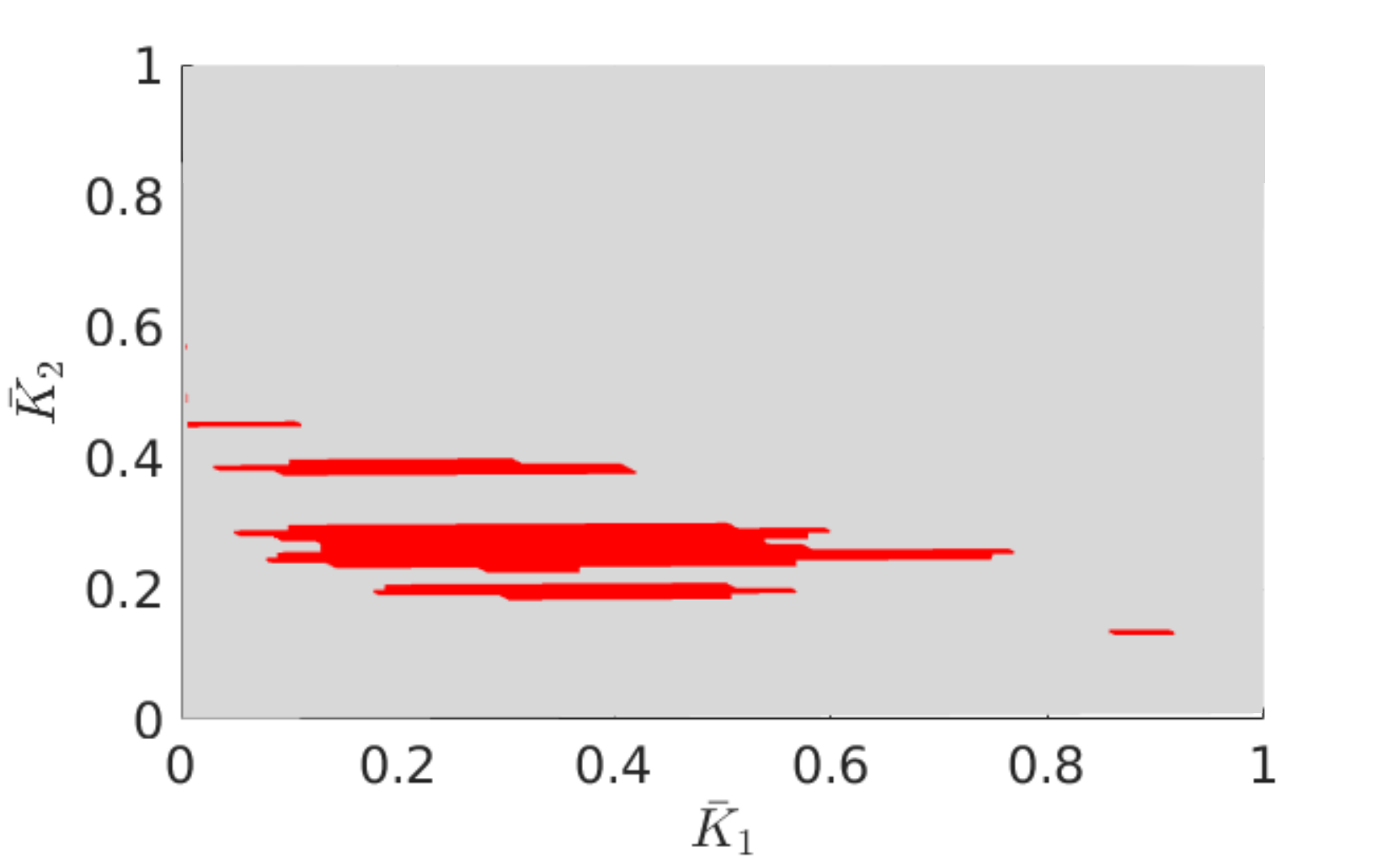}
\subcaption{\gls{eigf}}\label{fig::LLE_EIGF}
\end{subfigure}
\caption[Sample positions for different adaptive sampling techniques for the classification of chaotic motion]{\gls{mepe} and \gls{eigf} surrogate models based on 60 samples for two-dimensional LLE problem with example set of samples (left) and evaluated metamodel at reference points (right). }\label{fig::LLE_EIGF_MEPE}
\end{figure}

The problem of the two adaptive sampling techniques designed for regression purpose is evident when looking at the position of a sample set as displayed in Figure \ref{fig::LLE_EIGF_MEPE}. Here, the sample positions for \gls{mepe} and \gls{eigf} are shown on the left hand side of Figures \ref{fig::LLE_MEPE} and \ref{fig::LLE_EIGF} respectively. It can be seen that both methods focus on the zone where the response surface (see Figure \ref{fig::P1_surf}) shows drastic change.  The resulting metamodels are shown on the right-hand side. The \gls{mepe} surrogate classification appears much more proficient than \gls{eigf} for this case because \gls{mepe} has a more sophisticated exploration component. 

\section{Conclusion}

A dedicated classification-oriented kriging regression technique has been proposed to classify problems with highly fluctuating output in one- and two-dimensional parametric domains from only few observations. \gls{mivor} has shown promising results on a few classification problems, particularly for cases based on highly fluctuating and non-monotonic response surface. The presented adaptive technique is not limited to the proposed applications, it could be used for investigating any classification problem based on the knowledge of a continuous quantity of interest. Thus, an innovative and proficient adaptive sampling technique has been proposed for general classification using kriging with few observation points. In future work this method should be investigated with respect to its ability in tackling high-dimensional parametric problems. Besides, it would also be of interest to provide a robust error estimation framework to stop the adaptive scheme not only by reaching a maximally allowed number of observations but based on a desired accuracy level. 

\vspace{1cm}

\textbf{Acknowledgements} \\The authors acknowledge the financial support from the Deutsche Forschungsgemeinschaft under
Germany’s Excellence Strategy within the Cluster of Excellence
PhoenixD (EXC 2122, Project ID 390833453).

The results presented in this paper were partially carried out on the cluster system at the Leibniz University of Hannover, Germany.
\\ \\ \\
\textbf{Conflict of interest} \\ The authors declare that they have no conflict of interest.




%
%
%

\bibliographystyle{elsarticle-harv}   
\bibliography{bib}

\end{document}